\documentclass[conf]{new-aiaa} 
\usepackage[utf8]{inputenc}
\usepackage{soul}
\usepackage{graphicx}
\usepackage{amsmath}
\usepackage{cleveref}
\usepackage{subcaption}
\usepackage[version=4]{mhchem}
\usepackage{siunitx}
\usepackage{longtable,tabularx}
\usepackage{float}
\usepackage{bm}
\usepackage{booktabs}
\usepackage{multirow}
\usepackage{adjustbox}

\usepackage{titlesec}
\titleclass{\subsubsubsection}{straight}[
\subsubsection]
\newcounter{subsubsubsection}[subsubsection]
\renewcommand\thesubsubsubsection{\thesubsubsection.oman{subsubsubsection}}
\titleformat{\subsubsubsection}[runin]
  {\normalfont\normalsize\bfseries}{\thesubsubsubsection}{1em}{}
\titlespacing*{\subsubsubsection}{0pt}{3.25ex plus 1ex minus .2ex}{1em}

\title{BladeSDF : Unconditional and Conditional Generative Modeling of Representative Blade Geometries Using Signed Distance Functions}

\author[a,b]{Ashish S. Nair}
\author[a]{Sandipp Krishnan Ravi}
\author[a,c]{Itzel Salgado}
\author[a]{Changjie Sun}
\author[a]{Sayan Ghosh}
\author[a]{Liping Wang}

\affil[a]{GE Aerospace Research, Niskayuna, NY 12309, USA}
\affil[b]{University of Notre Dame, Holy Cross Dr, Notre Dame, IN 46556, USA}
\affil[c]{Northwestern University, 633 Clark St, Evanston, IL 60208, USA}

\begin{document}

\maketitle

\begin{abstract}

Generative AI has emerged as a transformative paradigm in engineering design, enabling automated synthesis and reconstruction of complex 3D geometries while preserving feasibility and performance relevance. This paper introduces a domain-specific implicit generative framework for turbine blade geometry using DeepSDF, addressing critical gaps in performance-aware modeling and manufacturable design generation. The proposed method leverages a continuous signed distance function (SDF) representation to reconstruct and generate smooth, watertight geometries with quantified accuracy. It establishes an interpretable, near-Gaussian latent space that aligns with blade-relevant parameters, such as taper and chord ratios, enabling controlled exploration and unconditional synthesis through interpolation and Gaussian sampling. In addition, a compact neural network maps engineering descriptors, such as maximum directional strains, to latent codes, facilitating the generation of performance-informed geometry. The framework achieves high reconstruction fidelity, with surface distance errors concentrated within $1\%$ of the maximum blade dimension, and demonstrates robust generalization to unseen designs. By integrating constraints, objectives, and performance metrics, this approach advances beyond traditional 2D-guided or unconstrained 3D pipelines, offering a practical and interpretable solution for data-driven turbine blade modeling and concept generation.

\end{abstract}


\section{Introduction}
Generative AI has emerged as a transformative paradigm in engineering design. It enables automated synthesis and reconstruction of complex 3D geometries, accelerating early-stage concept exploration while preserving feasibility and performance relevance. A key trend in this domain is the use of continuous implicit representations, particularly signed distance functions (SDFs). These representations allow the learning of compact, expressive manifolds of shapes that decode into smooth, watertight surfaces with well-behaved normals \cite{mescheder2019occupancy,chen2019implicit,gropp2020igrl,yariv2020multiview}. Such properties are essential for downstream analysis and manufacturability \cite{regenwetter2022deep}.

DeepSDF’s auto-decoder formulation has demonstrated high-quality shape representation, interpolation, and completion from partial inputs\cite{atzmon2020sal,takikawa2021nglod}. It reduces model size compared to prior approaches and supports inference from arbitrary SDF samples \cite{park2019deepsdf}. Recent advances have extended implicit learning to broader generative settings. Diffusion-based methods applied to implicit functions now achieve both unconditional and conditional generation for shape completion and single-view reconstruction. These methods expand the success of 2D diffusion into 3D implicit domains, demonstrating robustness to partial and noisy inputs through custom conditioning and modulation \cite{chou2023diffusion,ho2020ddpm,nichol2021improved,rombach2022ldm,vahdat2021scorelatent}. Beyond SDFs, the implicit-function literature includes techniques such as occupancy networks, NeRF-derived methods, and meta-learning SDF priors for generalization and semi/self-supervised regimes. These developments highlight the growing momentum toward flexible, differentiable 3D function-space representations \cite{chou2022gensdf,yariv2021volsdf,wang2021neus,sitzmann2020siren,chibane2020ifnets}. In parallel, text-to-3D and 3D-aware generative pipelines leverage image diffusion priors to synthesize or refine geometry from sparse supervision, extending 2D diffusion into 3D assets and offering complementary routes to implicit SDF-based modeling \cite{poole2022dreamfusion,lin2023magic3d,shue2022threeda}.

Despite this rapid progress, a specific gap remains in domain-focused, performance-aware generative modeling of turbine blade geometries. Most existing pipelines either rely on 2D renderings for performance guidance or operate on unconstrained 3D latent spaces. These approaches lack explicit enforcement of engineering specifications and fail to provide quantified reconstruction guarantees for blade-like forms. The engineering design review literature identifies persistent challenges in integrating constraints and objectives, modeling both form and functional performance simultaneously, and enabling interactive, design-in-the-loop workflows \cite{fuge2019ai4design,sun2021geomperf,maulik2020nncfd,yao2020surrogate,viquerat2021turbodl}.

Constrained generative pipelines in adjacent domains, such as automotive design, have demonstrated parameter-guided DeepSDF modeling and surrogate-based performance prediction for rapid iteration. However, these methods primarily target automotive geometries and image-based surrogates, rather than blade-centric implicit 3D reconstruction with clear geometric error quantification \cite{morita2024vehiclesdf}. While diffusion-SDF and related conditional implicit methods show promise for completion and reconstruction from partial inputs, they do not directly address turbine blade constraint enforcement. Additionally, they lack interpretable latent structures tied to blade parameters and fail to provide quantified reconstruction fidelity within engineering tolerances\cite{wang2023neuralparamcad}.

This paper addresses these gaps by introducing a DeepSDF-based generative framework tailored specifically to turbine blades. The framework makes three key contributions aligned with engineering requirements:

\begin{enumerate}
    \item \textbf{SDF-Based Blade Representation}: The framework formalizes an SDF-based blade representation and training regime. It focuses supervision near the surface using truncation and banded sampling. This approach ensures smooth, watertight geometry with well-behaved normals, which are suitable for learning and interpolation.
    \item \textbf{Interpretable Latent Manifold}: The framework establishes an interpretable, approximately Gaussian latent manifold. This latent space supports unconditional generation through interpolation and sampling. It also reveals principal axes aligned with blade-relevant geometric attributes, enabling controllable and explainable generative behavior within the blade domain.
    \item \textbf{Conditional Pathway for Performance-Informed Synthesis}: The framework introduces a conditional pathway that maps engineering-relevant descriptors, such as maximum directional strain triplets, to latent codes. This enables performance-informed synthesis while maintaining reconstruction feasibility objectives.
\end{enumerate}

The proposed method operationalizes a domain-specific implicit generative workflow for turbine blades using DeepSDF. It learns a continuous SDF representation to reconstruct and generate smooth, watertight geometries with quantified accuracy. The framework provides an interpretable, near-Gaussian latent space for controlled exploration and enables conditioning on engineering descriptors, such as maximum directional strains, to link geometry with performance context.

Overall, this approach advances beyond traditional 2D-guided or unconstrained 3D pipelines. It offers manufacturable, smooth outputs and performance-informed concept generation. The proposed framework demonstrates robust generalization to unseen designs and provides a practical, interpretable solution for data-driven turbine blade modeling and design generation.

The paper is organized as follows: Section 1 introduces the motivation and background for generative AI in engineering design, emphasizing signed distance functions (SDFs) for 3D geometry reconstruction and synthesis. Section 2 outlines the methodology, including the SDF-based blade representation, training regime, and latent manifold development. Section 3 describes the synthetic blade dataset and ground-truth SDF generation. Section 4 analyzes the learned latent space and its alignment with blade-relevant attributes. Section 5 presents results on reconstruction fidelity, latent space analysis, and performance-informed generative modeling. Finally, Section 6 concludes with a summary of contributions and implications for turbine blade modeling and concept generation.
\section{Methodology}

\subsection{Signed Distance Function (SDF)}
\label{subsec: SDF}
We represent each blade by a signed distance function (SDF), an implicit, differentiable field whose zero level set defines the surface, with the objective of obtaining smooth, watertight geometry with well-behaved normals suitable for learning and interpolation. The signed distance function (SDF) equals the Euclidean distance to the surface, with a sign indicating whether a point is outside or inside the solid \cite{jones2006sdf,baerentzen2005fastmarch}.

Let $\mathcal{S}\subset\mathbb{R}^{3}$ be a closed, orientable surface that bounds a compact region $\Omega$ (interior). For any $\bm{x}\in\mathbb{R}^{3}$, given the unsigned point-to-surface distance
\begin{equation}
\mathrm{dist}(\bm{x},\mathcal{S}) \;=\; \inf_{\bm{y}\in\mathcal{S}}\;\|\bm{x}-\bm{y}\|_{2}.
\end{equation}
The SDF $s^\ast:\mathbb{R}^{3}\rightarrow\mathbb{R}$ is
\begin{equation}
s^\ast(\bm{x}) \;=\;
\begin{cases}
\phantom{-}\mathrm{dist}(\bm{x},\mathcal{S}), & \bm{x}\notin \Omega \quad (\text{outside}),\\[3pt]
-\mathrm{dist}(\bm{x},\mathcal{S}), & \bm{x}\in \Omega \quad (\text{inside}),
\end{cases}
\qquad
\mathcal{S} \;=\; \{\bm{x}\in\mathbb{R}^{3}\;:\; s^\ast(\bm{x}) = 0\}.
\end{equation}

\subsubsection{Truncation and Sampling for Training}

To focus supervision near $\mathcal{S}$ and avoid large, uninformative residuals\cite{gropp2020igrl}, we use a \emph{clamped} (truncated) SDF
\begin{equation}
s_{\delta}(\bm{x}) \;=\; \mathrm{clip } (s^\ast(\bm{x}), -\delta, \delta\big), \qquad \delta>0,
\end{equation}
and sample query points $\{\bm{x}_j\}$ densely in a narrow band $\{\,\bm{x}:\; |s^\ast(\bm{x})|\le\delta\,\}$ plus a sparse set in the bounding box. The training pairs are $\big\{(\bm{x}_j,\, s_{\delta}(\bm{x}_j))\big\}$. For this work, the clamp distance was set to $\delta=0.1$.


\subsubsection{From Point Clouds to SDF Ground Truth}
\label{subsubsec:pc_to_sdf}

In our setting, each blade design is available as a point cloud with both near-surface samples and interior samples, so we use a convex-hull proxy to obtain the inside/outside signs and a KD-tree for distance magnitudes\cite{barill2018fastwinding}. Let $\mathcal{P}=\{\mathbf{p}_i\}_{i=1}^{N}\subset\mathbb{R}^3$ be the point cloud and its convex hull is defined as,
\begin{equation}
  \mathcal{H}=\mathrm{conv}(\mathcal{P})
  =\Big\{\sum_{i=1}^{N}\lambda_i \mathbf{p}_i \;:\; \lambda_i\ge 0,\ \sum_{i=1}^{N}\lambda_i=1\Big\}.
\end{equation}
The boundary $\partial\mathcal{H}$ is a triangular mesh with faces $f=1,\dots,F$, each lying on a plane
$\Pi_f=\{\mathbf{x}:\ \mathbf{n}_f^\top \mathbf{x}=b_f\}$ with unit normal $\mathbf{n}_f$ and offset
$b_f=\mathbf{n}_f^\top \mathbf{v}_f$ for any vertex $\mathbf{v}_f$ of face $f$.
All face normals are oriented outward using the hull-vertex centroid. With outward normals, a point is inside the hull iff it satisfies all supporting half-spaces:
\begin{equation}
  \mathbf{x}\in\mathcal{H}
  \quad\Longleftrightarrow\quad
  \mathbf{n}_f^\top \mathbf{x}\le b_f \quad \forall f .
\end{equation}
For a numerically robust sign, the violation margin is defined as
\begin{equation}
  m(\mathbf{x})=\max_{f}\big(\mathbf{n}_f^\top \mathbf{x}-b_f\big),
\end{equation}
and, with a tolerance $\tau\ge 0$, the sign is computed as
\begin{equation}
  \sigma(\mathbf{x})=
  \begin{cases}
    -1, & m(\mathbf{x})\le \tau \quad \text{(inside)},\\[2pt]
    \ \,1, & m(\mathbf{x})> \tau \quad \text{(outside)}.
  \end{cases}
\end{equation}

As the blades designs are near-convex and the cloud includes interior points, the hull encloses the measured volume and only mildly over-approximates nonconvex features; this yields stable signs even under noisy or non-uniform sampling.

\begin{figure}[h!]
    \centering
    \includegraphics[width=0.5\linewidth]{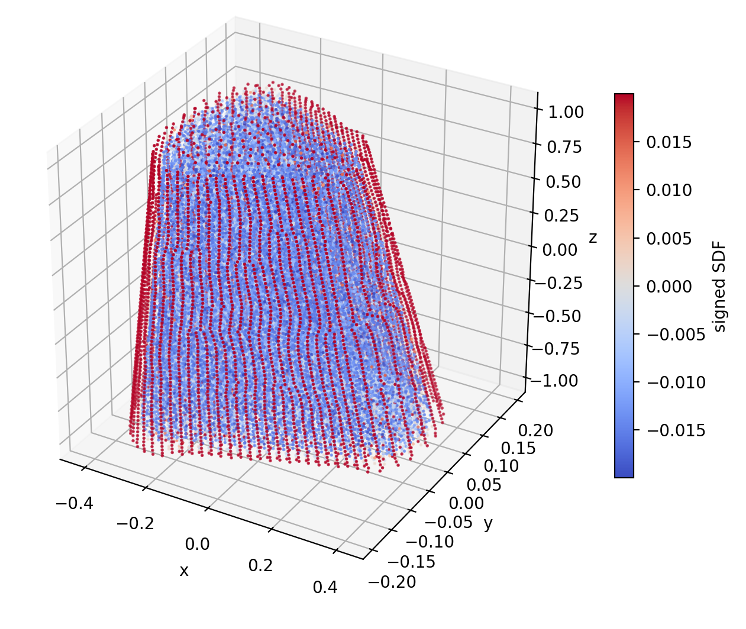}
    \caption{Ground-truth SDF near the surface : Labeled samples for one blade obtained from the point cloud (convex-hull sign + KD-tree magnitude); visualization restricted to $|s|\le 0.1$ ($\delta=0.1$) to highlight the training band.}
    \label{fig:gt_sdf}
\end{figure}

For the unsigned distance, we decouple magnitude from sign and query nearest neighbors on a surface-focused subset. Let $\mathcal{P}_{\text{surf}}=\{\mathbf{q}\in\mathcal{P}: |m(\mathbf{q})|\le \tau_s\}$ collect points near the hull boundary. With a KD-tree over $\mathcal{P}_{\text{surf}}$, the unsigned distance is computed as,
\begin{equation}
    d(\mathbf{x})=\min_{\mathbf{q}\in \mathcal{P}_{\text{surf}}}\|\mathbf{x}-\mathbf{q}\|_2,
\end{equation}
which approximates $\mathrm{dist}(\mathbf{x},\mathcal{S})$ per query. The resulting approximate SDF is
\begin{equation}
    \tilde{s}(\mathbf{x})=\sigma(\mathbf{x})\,d(\mathbf{x}),\qquad
\tilde{s}_\delta(\mathbf{x})=\mathrm{clip}\big(\tilde{s}(\mathbf{x}),-\delta,\delta\big),
\end{equation}
with $\delta>0$ to emphasize near-surface samples during training. With this labeling pipeline in place, we generate $N_i=20{,}000$ ground-truth SDF pairs per blade. Prior to sampling and labeling, all designs are globally normalized to the unit cube $[-1,1]^3$. We draw $50\%$ of query points in a near-surface band (e.g., $\{\,\mathbf{x}:|\tilde{s}(\mathbf{x})|\le \delta\,\}$ with $\delta=0.1$) and the remaining $50\%$ uniformly in the bounding box, and we clamp targets with $\delta=0.1$ when forming $\tilde{s}_\delta(\mathbf{x})$. Figure~\ref{fig:gt_sdf} visualizes the ground-truth SDF samples for a representative blade, generated from its point cloud via the convex-hull sign and KD-tree distance pipeline, only points within the truncation band $|s|\le \delta=0.1$ are shown to emphasize near-surface supervision.

\subsection{DeepSDF}

We adopt a decoder-only (auto-decoder) formulation: a shared MLP decoder
$f_{\theta}:\mathbb{R}^{k}\times\mathbb{R}^{3}\!\to\!\mathbb{R}$ maps a
per-design latent code $\mathbf{z}_{i}\in\mathbb{R}^{k}$ and a 3D query
$\mathbf{x}\in\mathbb{R}^{3}$ to a signed distance value,
\begin{equation}
    s_{\theta}(\mathbf{z}_{i},\mathbf{x}) \;=\; f_{\theta}(\mathbf{z}_{i},\mathbf{x}).
\end{equation}

In our experiments the latent dimension is set to $k=256$ and $f_{\theta}$ is an 8-layer MLP of width 512 with ReLU activations, batch normalization, and dropout with probability $p=0.2$ after each hidden layer\cite{sitzmann2020siren,takikawa2021nglod}.

\subsubsection{Training (Joint Optimization of Decoder and Latents)}
For each training design $i=1,\dots,n_{\text{train}}$ we form an SDF supervision set $\mathcal{D}_{i}=\{(\mathbf{x}_{ij},\,\tilde{s}_{\delta}(\mathbf{x}_{ij}))\}_{j=1}^{N_i}$ as described in Section~\ref{subsubsec:pc_to_sdf} (clamp $\delta=0.1$, near-surface biased sampling). We minimize a clamped reconstruction loss with a quadratic latent prior,
\begin{equation}
    \mathcal{L}(\theta,\{\mathbf{z}_{i}\})
=\frac{1}{\sum_i N_i}\sum_{i}\sum_{j}
\Bigl|\,
\mathrm{clip }(f_{\theta}(\mathbf{z}_{i},\mathbf{x}_{ij}),-\delta,\delta\big)
-\tilde{s}_{\delta}(\mathbf{x}_{ij})
\,\Bigr|
\;+\;\lambda_{z}\,\frac{1}{n_{\text{train}}}\sum_{i}\|\mathbf{z}_{i}\|_{2}^{2}.
\end{equation}

The latent prior keeps codes near a zero-mean Gaussian and improves generalization\cite{kingma2014vae,rezende2015flows}. We optimize $\theta$ and $\{\mathbf{z}_{i}\}$ jointly using Adam with initial step size $10^{-3}$ and a step scheduler that halves the rate every 500 iterations. Dropout and batch normalization are used in training mode only. In our setup $n_{\text{train}}=222$ designs and each design contributes $N_i=20{,}000$ labeled SDF pairs.

\subsubsection{Inference on Testing Set (Latent-only Optimization)}
At test time the decoder is frozen and only a new code $\mathbf{z}$ is optimized to fit SDF observations of the unseen design, 
\begin{equation}
    \mathbf{z}^{\star}
=\arg\min_{\mathbf{z}}\;
\frac{1}{N}\sum_{j=1}^{N}
\Bigl|\,
\mathrm{clip }(f_{\theta}(\mathbf{z},\mathbf{x}_{j}),-\delta,\delta\big)
-\tilde{s}_{\delta}(\mathbf{x}_{j})
\,\Bigr|
\;+\;\lambda_{z}\|\mathbf{z}\|_{2}^{2},
\end{equation}
initialized at $\mathbf{z}=\mathbf{0}$ (or the latent mean). This isolates the representation learned by $f_{\theta}$ from test-time fitting of the code. To extract a mesh, we evaluate the scalar field $F(\mathbf{x})=f_{\theta}(\mathbf{z}^{\star},\mathbf{x})$ on a uniform grid $\mathbf{x}\in[-1,1]^{3}$ and apply marching cubes to the zero iso-surface $\{\mathbf{x}:F(\mathbf{x})=0\}$, producing a watertight triangle mesh \cite{lorensen1987marchingcubes,kazhdan2006poisson}.

\subsection{Blade Designs Dataset}

\begin{figure}[h!]
    \centering
    \includegraphics[width=0.5\linewidth]{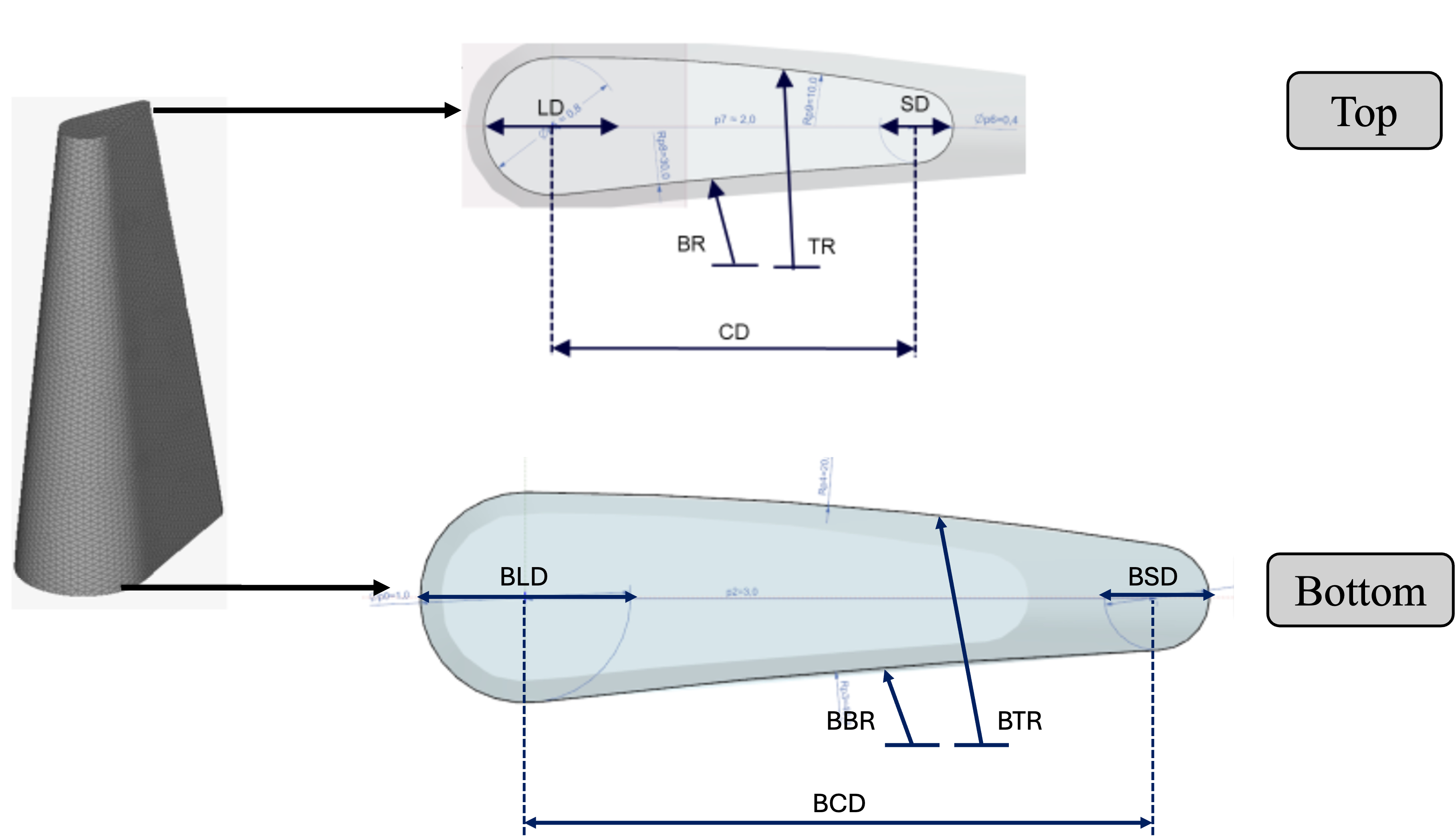}
    \caption{Blade parameterization for automated dataset generation }
    \label{fig:blade_schematic}
\end{figure}

The dataset comprises synthetic blade geometries parameterized by two coaxial surfaces, “bottom” and “top”, whose parameters are depicted in Figure.~\ref{fig:blade_schematic} and their numerical bounds are specified in Table~\ref{tab:blade_params}. Bottom-surface parameters include the large and small diameters (BLD, BSD), the inter-surface center distance (BCD), and blend radii (BBR, BTR). The top surface is derived from the bottom via three scale ratios $K_1,K_2,K_3\in[0.2,0.8]$ applied to BLD, BSD, and BCD, respectively, plus independent top blend radii (BR, TR), producing blades that span tapered to prismatic forms and shorter to longer top chords. We use a dataset with  $n_{\text{train}}=222$ training designs and $n_{\text{test}}=300$ test designs, normalize each geometry globally to the unit cube $[-1,1]^3$, and then compute ground-truth SDF supervision using a convex-hull sign test and KD-tree distance magnitude\cite{berger2017reconSurvey}.  

\begin{table}[h!]
  \centering
  \caption{Blade geometry parameter ranges}
  \label{tab:blade_params}
  \begin{tabular}{llcc}
    \toprule
    \textbf{Geom} & \textbf{Parameter} & \textbf{Low} & \textbf{High} \\
    \midrule
    \multirow{5}{*}{Bottom}
      & Large Diameter (BLD)        & 0.5 & 2   \\
      & Small Diameter (BSD)        & 0.2 & 1   \\
      & Center Dist (BCD)           & 2   & 4   \\
      & Bot Large Radius (BBR)      & 5   & 70  \\
      & Top Large Radius (BTR)      & 5   & 70  \\
    \midrule
    \multirow{5}{*}{Top}
      & Large Diameter (LD) $=K_{1}\!\cdot\!\mathrm{BLD}$ & 0.2 & 0.8 \\
      & Small Diameter (SD) $=K_{2}\!\cdot\!\mathrm{BSD}$ & 0.2 & 0.8 \\
      & Center Dist (CD) $=K_{3}\!\cdot\!\mathrm{BCD}$    & 0.2 & 0.8 \\
      & Bot Large Radius (LR)                           & 2   & 50  \\
      & Top Large Radius (TR)                          & 2   & 50  \\
    \bottomrule
  \end{tabular}
\end{table}

\begin{figure}[!ht]
    \centering
    \includegraphics[height=0.65\linewidth, width=1.0\linewidth]{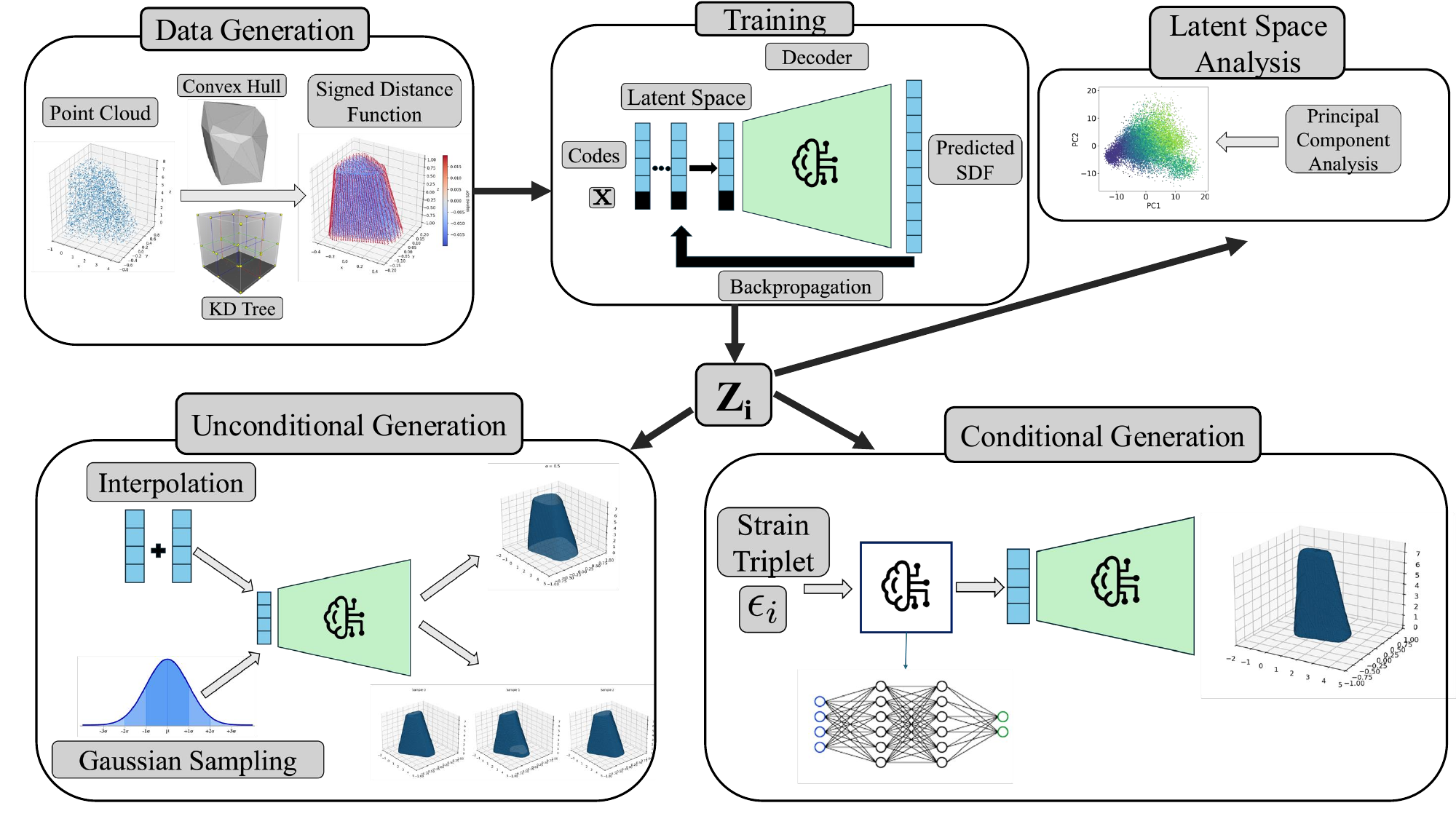}
    \caption{End-to-end workflow for DeepSDF-based blade modeling. Data generation: point clouds to ground-truth SDF via convex-hull sign and KD-tree distance; normalized to ($[-1,1]^3$). Training (auto-decoder): jointly learn decoder weights and 256-D per-design latents from truncated SDF pairs with an ($L_2$) latent prior. Latent-space analysis: PCA and marginals reveal interpretable axes and an approximately Gaussian, well-centered code distribution. unconditional generation: synthesize shapes by latent interpolation and diagonal-Gaussian sampling. Conditional generation: NN-map ($g_\phi$) from target maximum strains (($\varepsilon_x$,$\varepsilon_y$,$\varepsilon_z$)) to latents for goal-directed designs. Reconstruction and evaluation: extract meshes with marching cubes; assess accuracy via a surface distance metric and NRMSE.}
    \label{fig: Workflow_schematic}
\end{figure}

Figure~\ref{fig: Workflow_schematic} summarizes our end-to-end workflow: beginning with data Generation, we convert blade point clouds into ground-truth SDF labels using the convex-hull sign and KD-tree distance pipeline, with designs globally normalized and near-surface sampling emphasized. In training, a decoder-only DeepSDF model jointly learns per-design latent codes and network weights from these truncated SDF pairs, establishing a smooth implicit representation for each blade. The learned latent Space is then probed via PCA and code statistics to reveal interpretable axes aligned with original parameters and to justify simple, Gaussian-like priors. Under unconditional generation, we synthesize new geometries either by linear interpolation between codes or by sampling a diagonal Gaussian fit to the training codes; conditional generation augments this with an neural network (NN)-map from target maximum strains to latent codes for goal-directed synthesis. Finally, meshes are extracted from the implicit field via marching cubes for visualization and downstream analysis. Together, these stages form a coherent pipeline from raw geometric data to reconstruction, analysis, and design generation.

\section{Results}

This section reports results on (i) reconstructing blade geometries with using the DeepSDF framework, (ii) analyzing the learned latent space and its correspondence to the blades’ original parametric design variables, and (iii) generating novel designs. We evaluate linear interpolations between latent codes and sampling by fitting a Gaussian to the training-code distribution. We also learn a conditional mapping from maximum directional strain components $(\varepsilon_x,\varepsilon_y,\varepsilon_z)$ to latent codes, enabling synthesis of designs conditioned on target mechanical responses. We summarize the quantitative and qualitative outcomes for each route and highlight key trends.

\subsection{Training}

\begin{figure}[h!]
    \centering
    \includegraphics[width=0.5\linewidth]{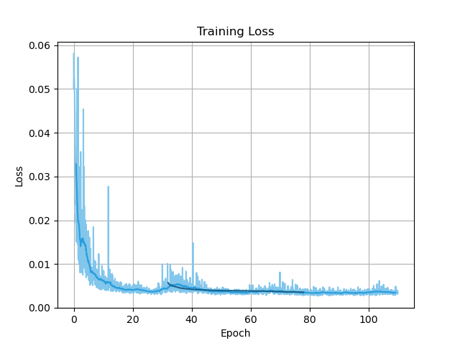}
    \caption{Training loss convergence for DeepSDF auto-decoder (8×512, $z=256$) : Clamped SDF L1 loss ($\delta=0.1$) drops >90\% by \~100 epochs, then plateaus.}
    \label{fig:Training_loss}
\end{figure}

 The standard DeepSDF auto-decoder formulation was used for this work, jointly optimizing per-design latent codes $\mathbf{z}_i$ and decoder weights $\theta$. The decoder was an 8-layer MLP (width 512) with batch normalization and dropout ($p=0.2$) after each hidden layer. Latent codes had dimension 256. Training used $n_{\text{train}}=222$ blade designs; evaluation used a held-out set of $n_{\text{test}}=300$ designs. The loss used clamped SDF residuals with truncation $\delta=0.1$. We optimized with Adam, the initial learning rate $\eta_0=10^{-3}$, and a step schedule that halves $\eta$ every 500 iterations. Figure~\ref{fig:Training_loss} shows training loss versus epoch: the loss decreased by >90\% within $\sim$120 epochs and then plateaued, indicating convergence.

\begin{figure}[h!]
    \centering
    \includegraphics[width=0.5\linewidth]{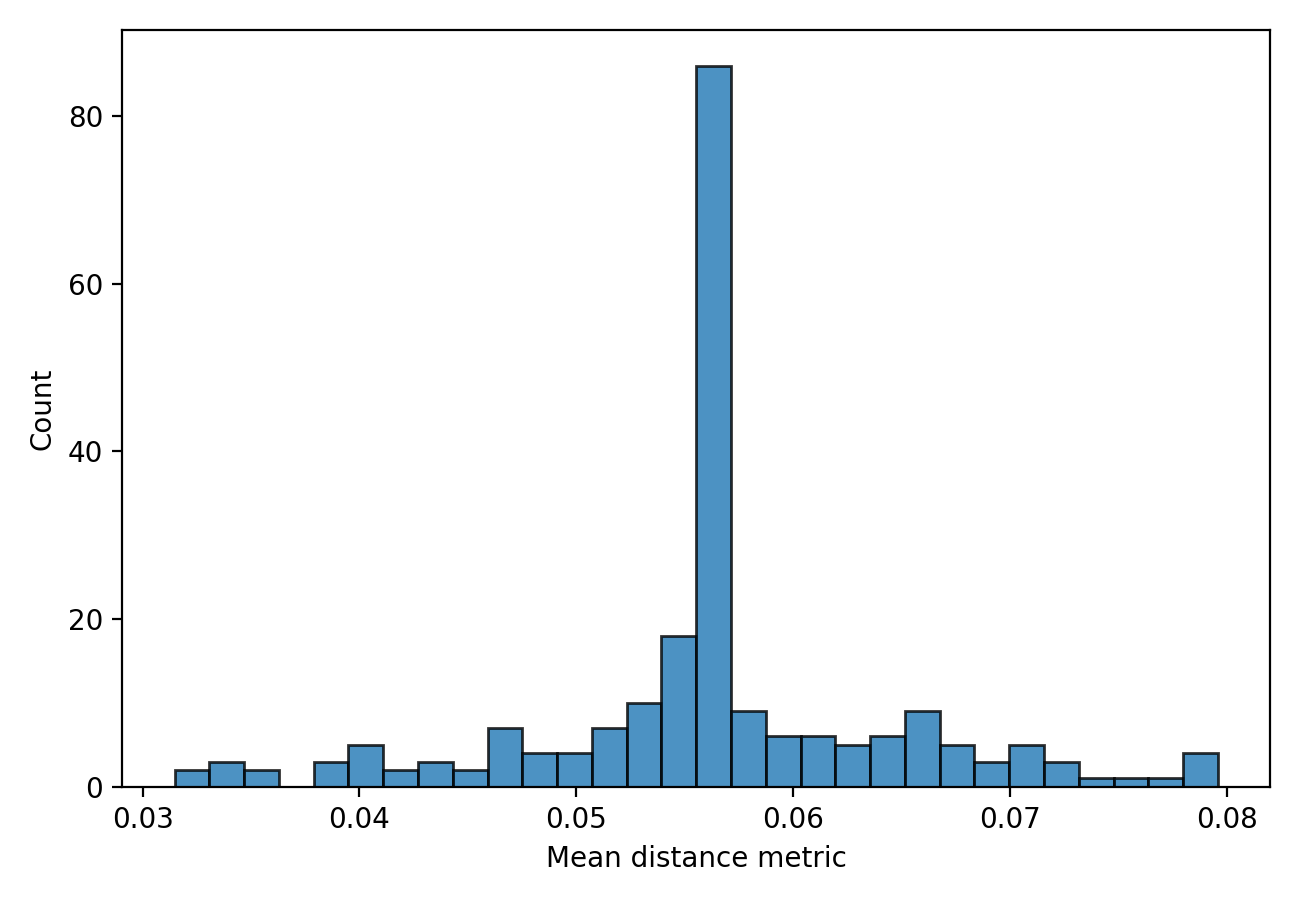}
    \caption{Training-set distribution of surface distance (Eq.~\ref{eq:Distance Metric}) : Histogram over $n_{\text{train}}=222$ reconstructions; mass concentrates near $5.5\times10^{-2}$ ($<1\%$ of $D_{\max}$), with a shallow tail to $8\times10^{-2}$.}
    \label{fig:distance_metric_train}
\end{figure}

To quantitatively assess reconstruction accuracy, we define a directed surface distance from the reference blade surface $S_{\mathrm{ref}}$ to the predicted surface $S_{\mathrm{pred}}$,

\begin{equation}
    \mathrm{Distance\ Metric}\!\left(S_{\mathrm{ref}}, S_{\mathrm{pred}}\right)
=\frac{1}{N}\sum_{i=1}^{N}\;\min_{y\in S_{\mathrm{pred}}}\left\|x_i - y\right\|_2,
\qquad \{x_i\}_{i=1}^{N}\subset S_{\mathrm{ref}},
\label{eq:Distance Metric}
\end{equation}
which measures the average point-to-surface nearest-neighbor distance. Figure~\ref{fig:distance_metric_train} shows the distribution of this metric between the original blade point clouds and the DeepSDF reconstructions over the training set. From the figure, for $\sim 80$ designs (36\% of the 222 designs in training), the distance metric concentrates around $5.5\times 10^{-2}$, which is <1\% of the blade’s maximum extent ($D_{\max}\approx 7$ in the $z$ direction). The maximum observed value is $8\times 10^{-2}$, which is roughly 1.1\% of $D_{\max}$, and remains close to the 1\% target. Considering this distribution, we conclude that DeepSDF achieves $\approx$1\% absolute surface deviation (relative to $D_{\max}$) across the training designs, with the majority of cases comfortably below that threshold.

\subsection{Reconstructions on the Testing Set}

This section presents a qualitative and quantitative assessment of DeepSDF reconstructions on the held-out test set ($\sim$300 blade designs). During test-time reconstruction, the decoder weights are frozen and only the per-design latent code is optimized, independently for each blade, to reduce the discrepancy between the predicted SDF and the ground-truth SDF. The reconstructed surface is then predicted from the SDF values using the marching cubes algorithm which determines the zero iso-surface of the predicted SDF function. This isolates the representation capacity of the decoder learned during training from test-time fitting, attributing any improvements solely to the latent code refinement.

\begin{figure}[h!]
    \centering
    \begin{subfigure}[b]{0.49\textwidth}
        \centering
        \includegraphics[width=\textwidth]{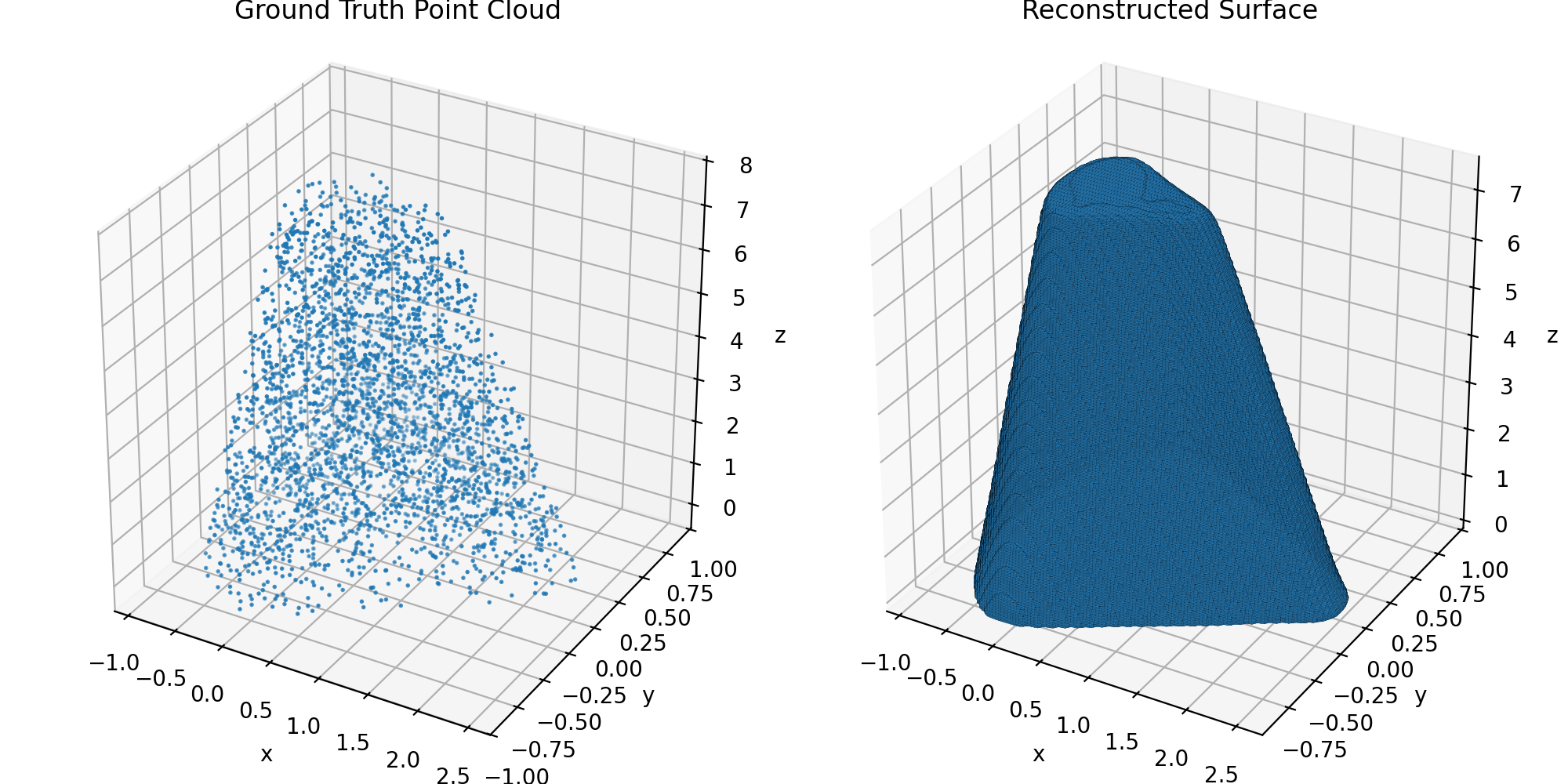}
        \caption{}
    \end{subfigure}
    \hfill
    \begin{subfigure}[b]{0.49\textwidth}
        \centering
        \includegraphics[width=\textwidth]{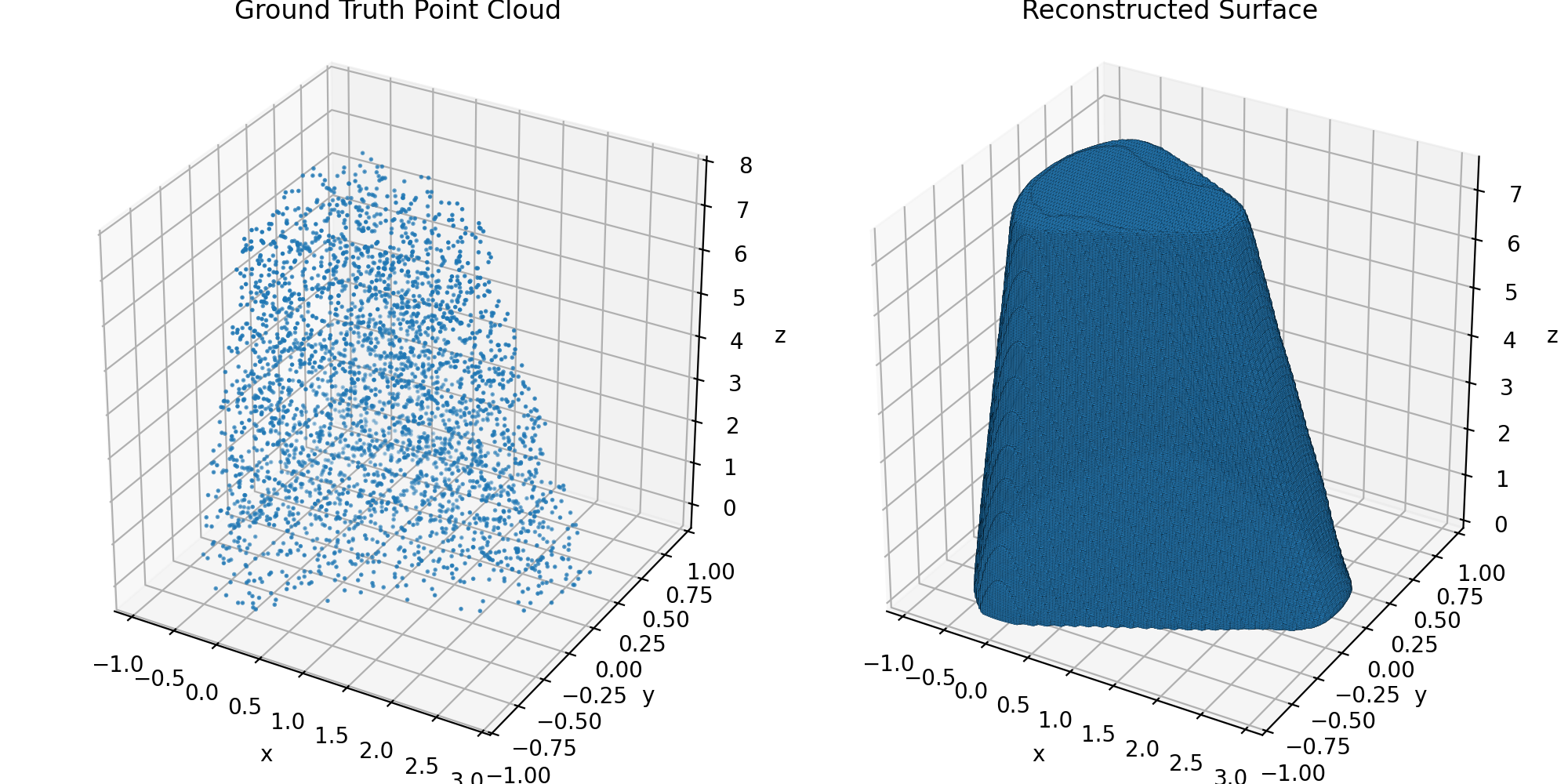}
        \caption{}
    \end{subfigure}

    \begin{subfigure}[b]{0.49\textwidth}
        \centering
        \includegraphics[width=\textwidth]{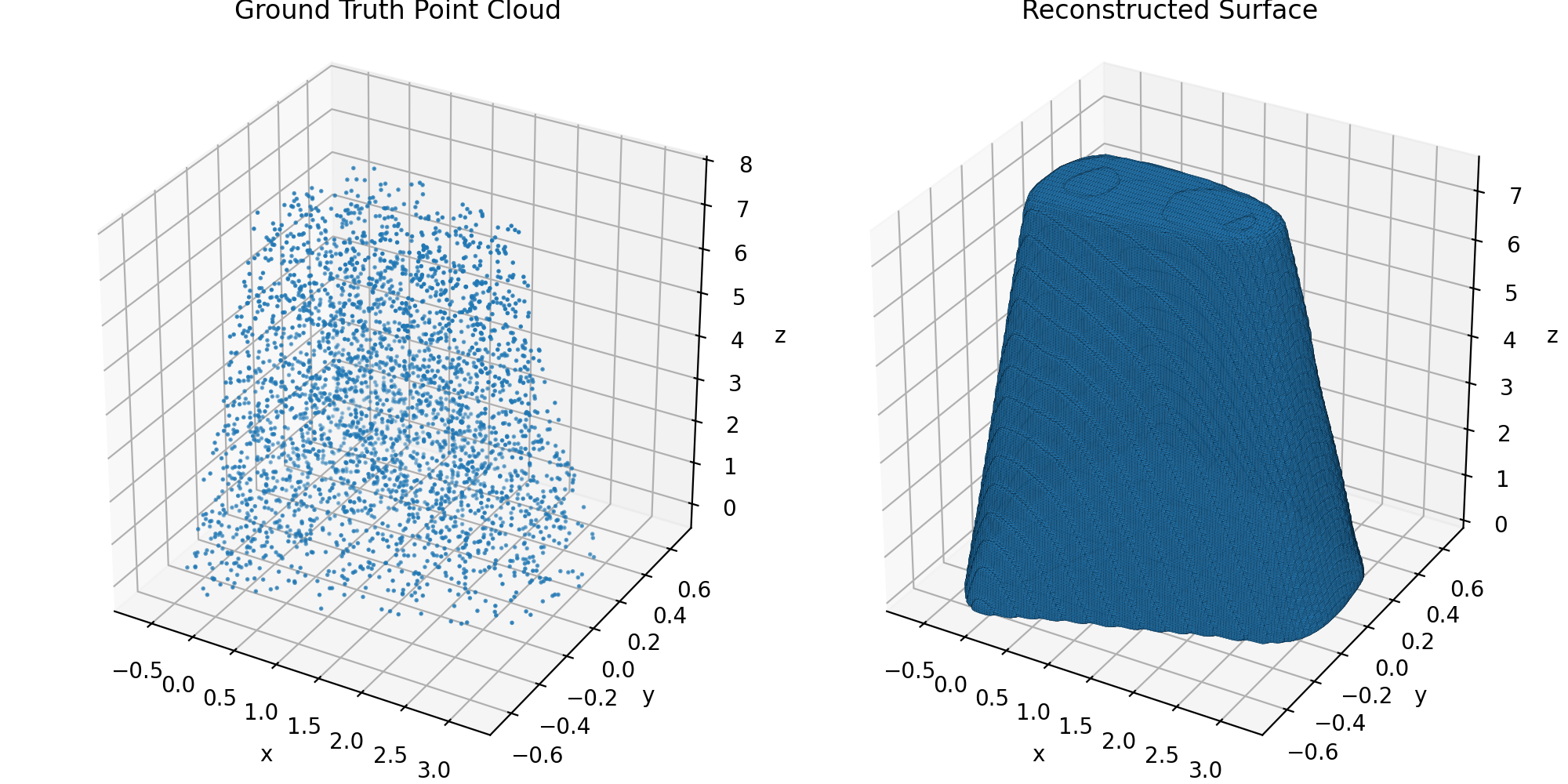}
        \caption{}
    \end{subfigure}
    \hfill
    \begin{subfigure}[b]{0.49\textwidth}
        \centering
        \includegraphics[width=\textwidth]{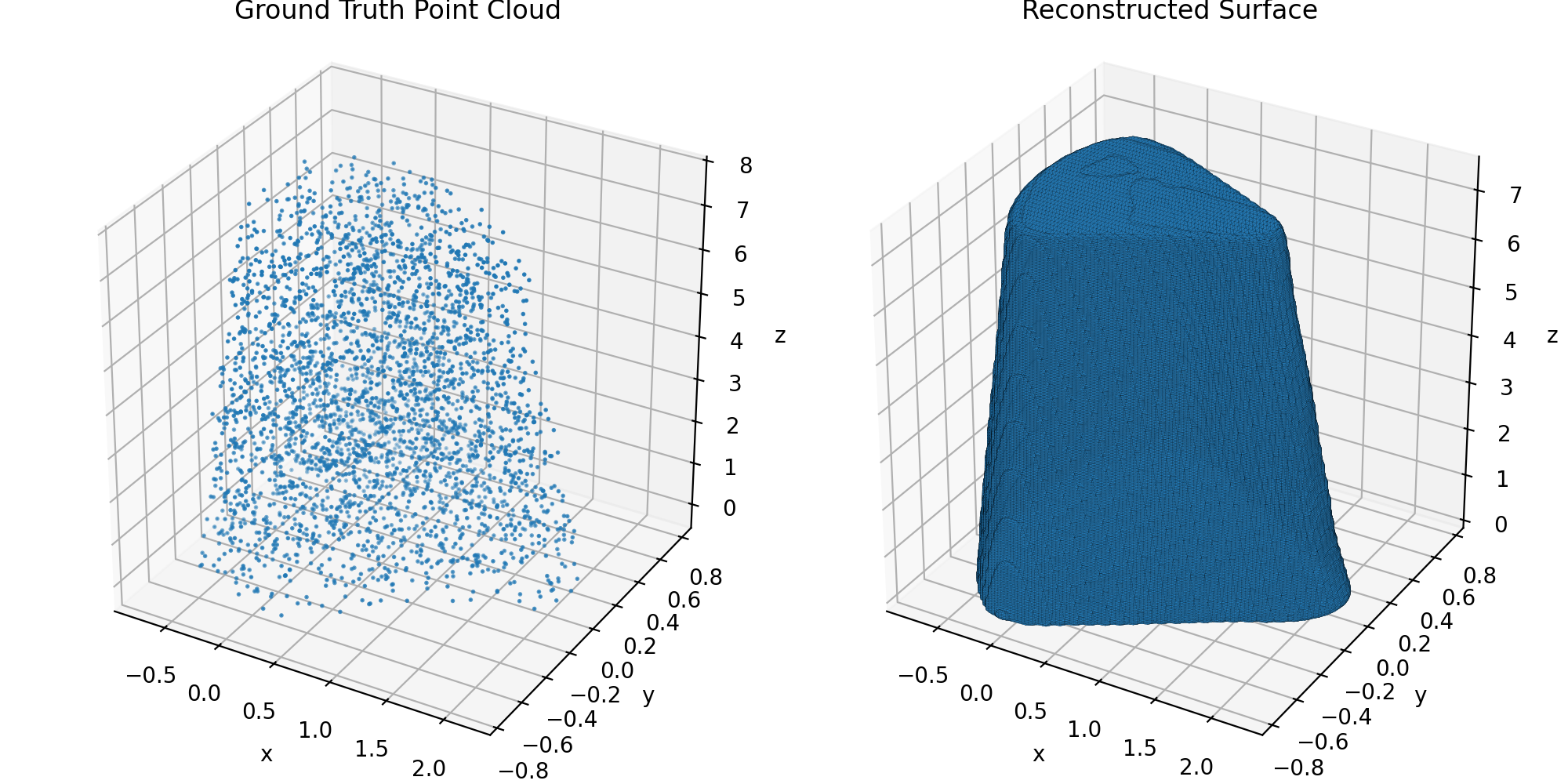}
        \caption{}
    \end{subfigure}
    
    \caption{Reconstructions on the held-out test set for (a) Design 302, (b) Design 306, (c) Design 387, (d) Design 498. Each subfigure shows a ground-truth point cloud (left) and its DeepSDF reconstruction (right), obtained by optimizing only the per-design latent code with the decoder frozen. The decoded surfaces appear smooth and free of spurious artifacts, closely reproducing the gross shape features.}
    \label{fig:Reconstructions_test}
\end{figure}

Figure~\ref{fig:Reconstructions_test} shows side-by-side comparisons of the original blade point clouds (left) and the reconstructed surfaces (right) across a range of designs (spanning different underlying parametric settings). Visual inspection indicates that the reconstructions exhibit smooth, well-formed surfaces with no obvious artifacts or spurious oscillations, closely following the gross shape features of the references. One of the primary test-time objectives was to verify whether DeepSDF yields smooth, physically plausible blade geometries which is satisfied. Quantitative error statistics for these reconstructions are reported below and corroborate the visual findings.

\begin{figure}[h!]
    \centering
    \includegraphics[width=0.5\linewidth]{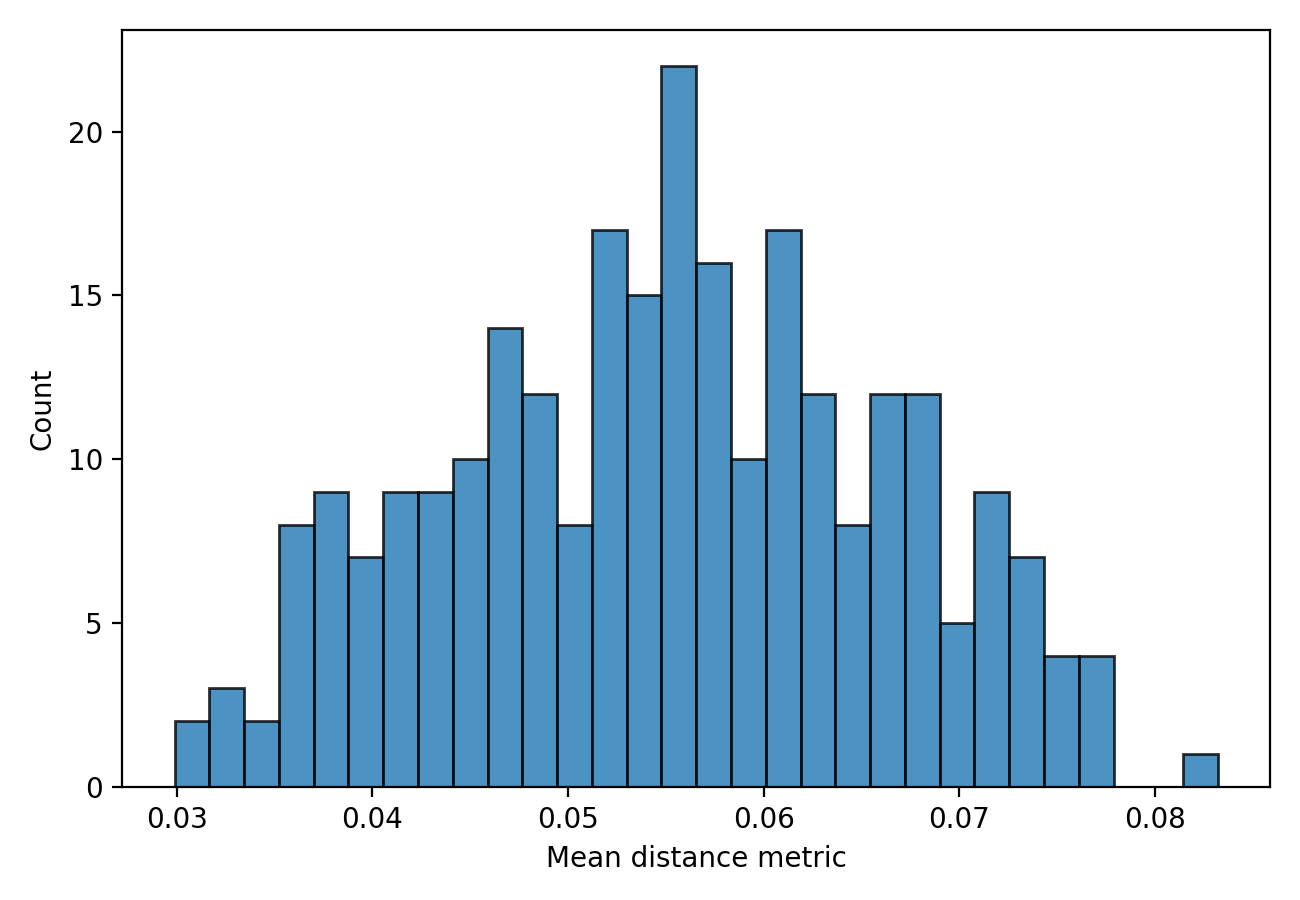}
    \caption{Test-set distribution of surface distance (Eq.~\ref{eq:Distance Metric}) : Histogram over $n_{\text{test}}=300$ reconstructions; overall range is comparable to training, but the distribution is flatter with increased mass in $5.5\times10^{-2}$–$8\times10^{-2}$; values remain $\lesssim 1\%$ of $D_{\max}$.}
    \label{fig:distance_metric_test}
\end{figure}

Figure~\ref{fig:distance_metric_test} reports the distribution of the distance metric (Eq.~\ref{eq:Distance Metric}) between the surface represented by the original point cloud and the reconstructed surface predicted by the DeepSDF framework on the held-out test set. The observed minimum and maximum values closely match those of the training set, indicating no large generalization gap in absolute scale. The key difference lies in the shape of the distribution. While the training histogram is sharply concentrated near $5.5\times 10^{-2}$ (i.e., $<\!1\%$ of the blade’s maximum extent, $D_{\max}\approx 7$ in the $z$ direction), the test-set histogram is more evenly spread, with noticeably greater mass in the range $[\,5.5\times 10^{-2},\,8\times 10^{-2}\,]$. This shift is consistent with the evaluation methodology: during training, both decoder weights and per-design latent codes are jointly optimized to match SDF values, whereas at test time the decoder weights are frozen and only the latent code is optimized for each design. Consequently, test reconstructions inherit a fixed representational capacity from the trained decoder, leading to a broader error distribution even though the overall range remains comparable to training. In summary, DeepSDF maintains $\sim\!1\%$ absolute deviation relative to $D_{\max}$ across the test set, but with a flatter distribution than in training due to the frozen-decoder constraint.

\subsection{Latent Space Analysis}

In this section, we examine what the learned latent space captures with respect to the blades’ original parametric design variables and whether its empirical distribution can support generative sampling of new designs. Each design is encoded by a 256-dimensional latent code. To probe its structure, we perform principal component analysis (PCA) on the test-set latent codes and project all test designs onto the plane spanned by the first two principal components (PC1–PC2), yielding a 2D embedding for visualization and traversal analyses.

\begin{figure}[H]
  \centering

  \includegraphics[width=0.60\linewidth]{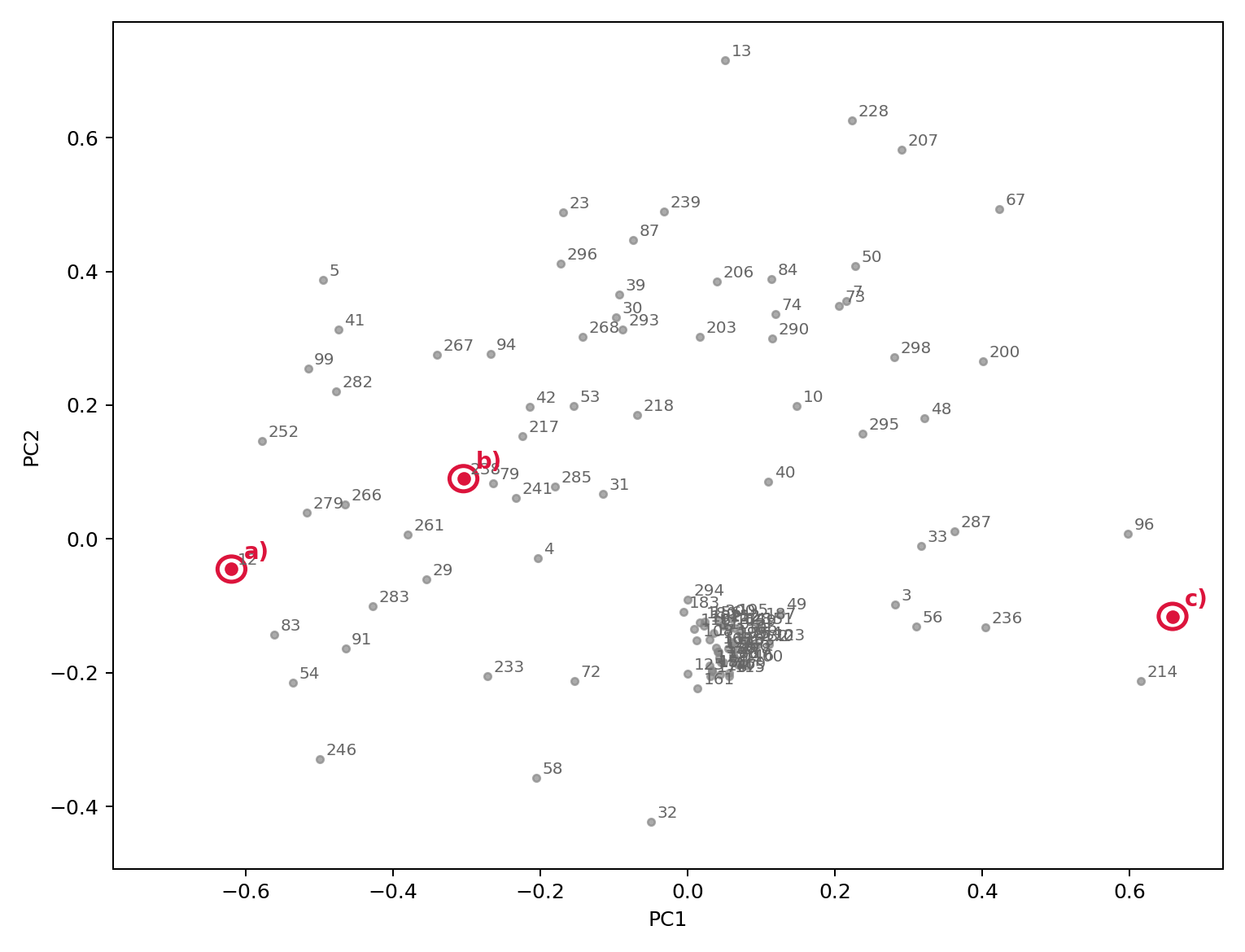}

  \vspace{0.75em}

  \begin{subfigure}[t]{0.45\linewidth}
    \centering
    \includegraphics[width=\linewidth]{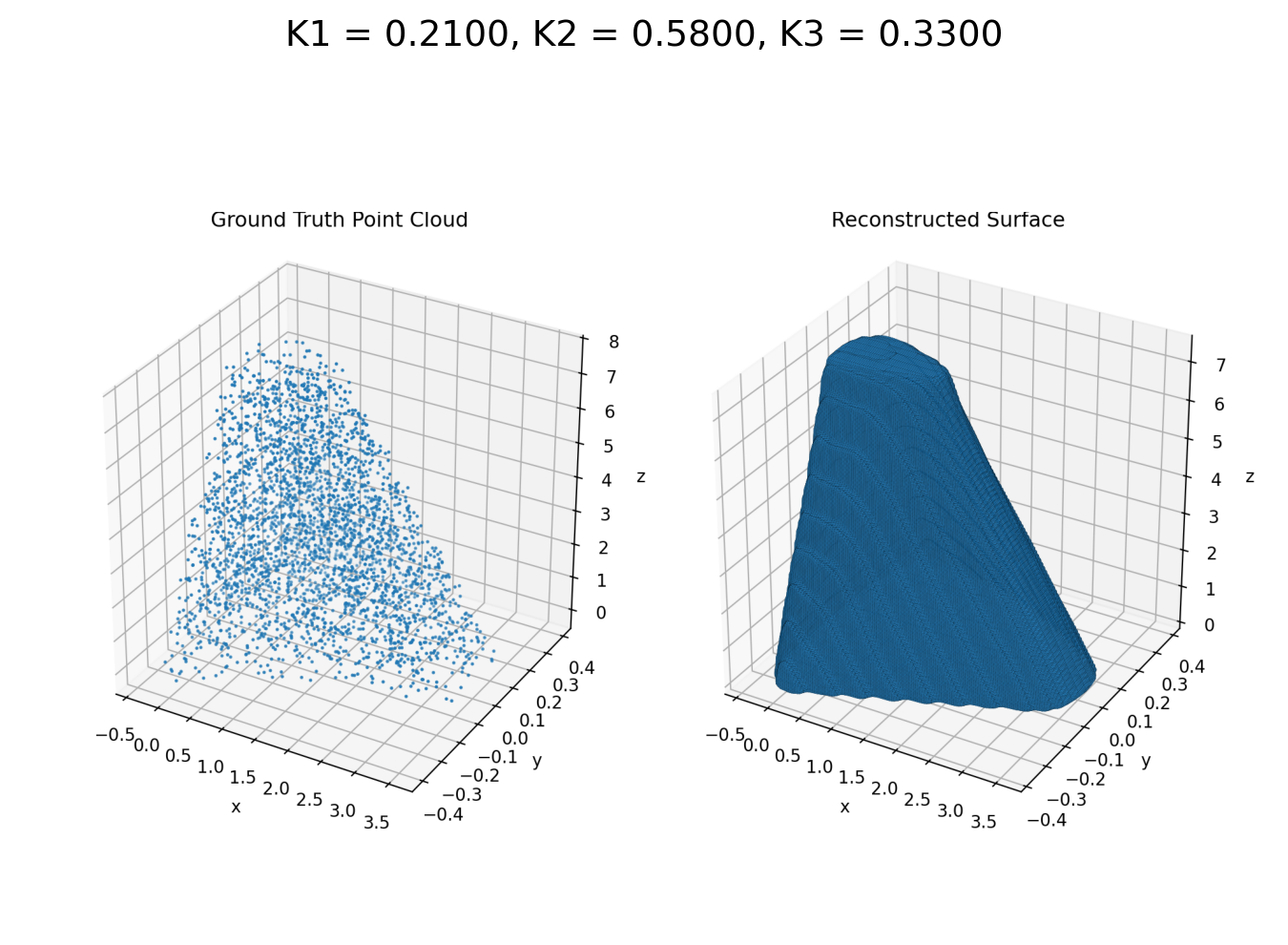}
    \caption{}
  \end{subfigure}
  \hfill
  \begin{subfigure}[t]{0.45\linewidth}
    \centering
    \includegraphics[width=\linewidth]{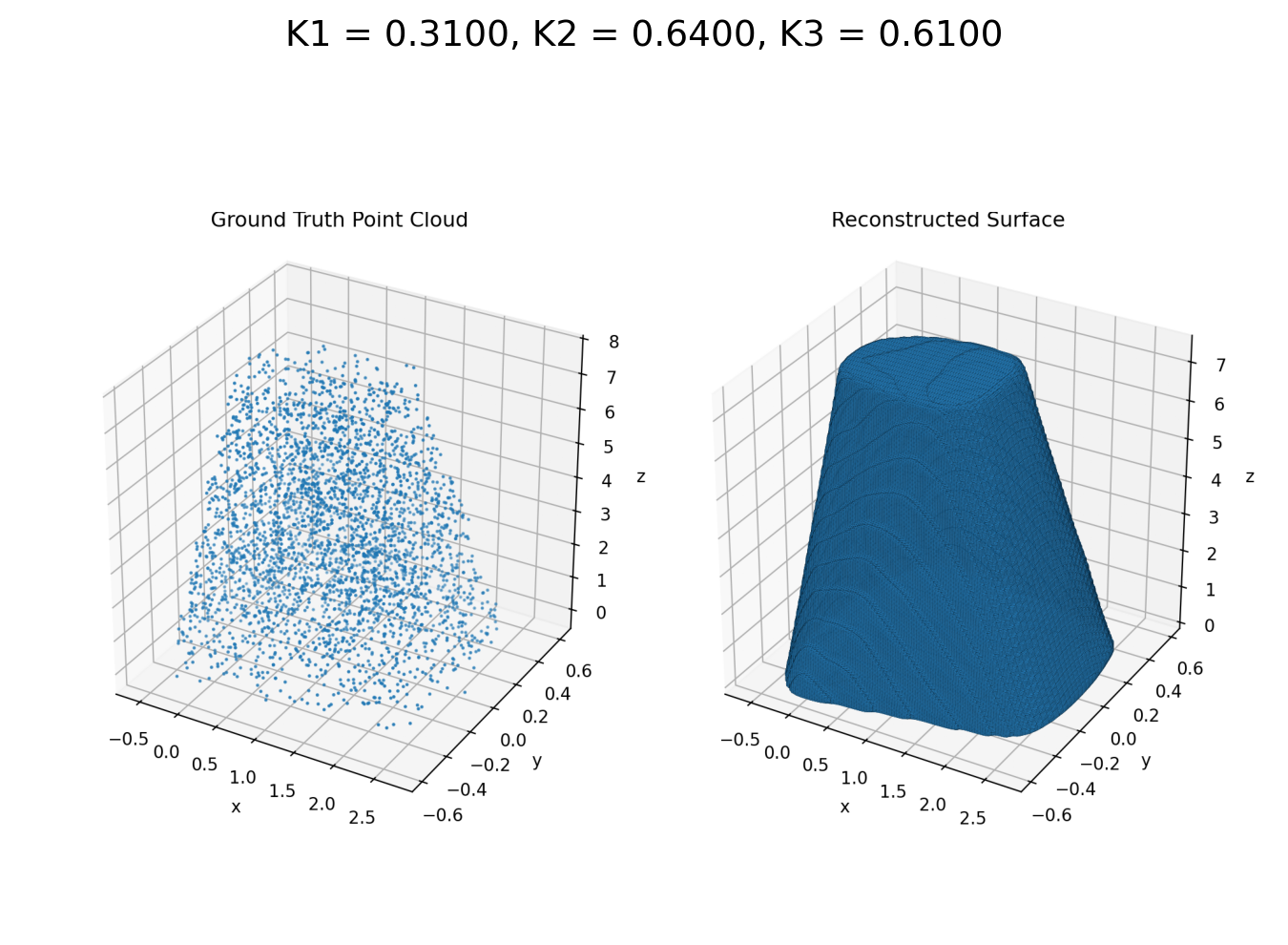}
    \caption{}
  \end{subfigure}
  \hfill
  \begin{subfigure}[t]{0.45\linewidth}
    \centering
    \includegraphics[width=\linewidth]{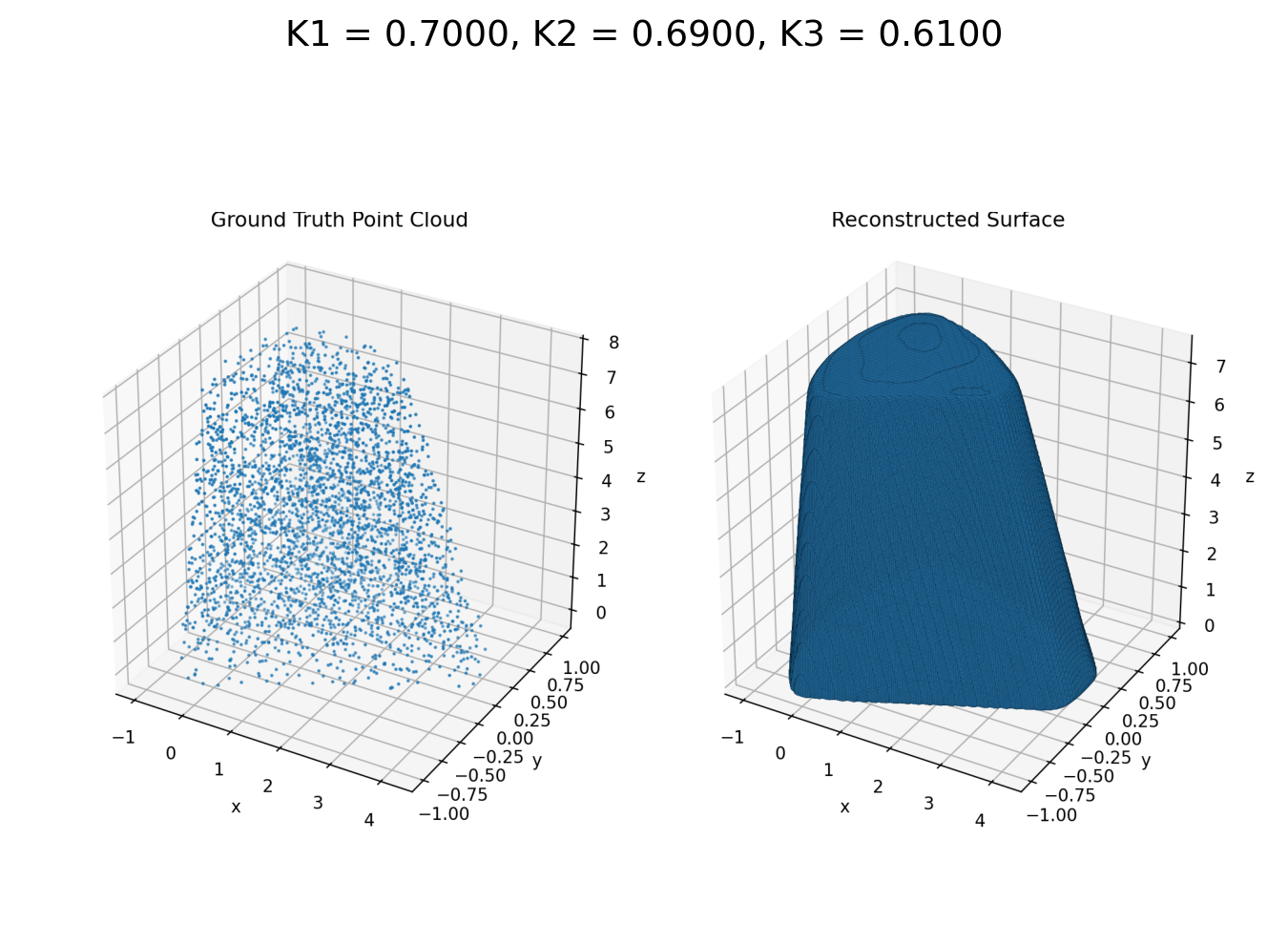}
    \caption{}
  \end{subfigure}

  \caption{PC1 traversal and correlation with $K_1$ for (a) Design 12, (b) Design 238, (c) Design 96. Three decoded samples ordered by increasing PC1 coordinate; $K_1$ ($\frac{SD}{BSD}$) increases monotonically from 0.2 to 0.8, producing a transition from tapered/conical to prismatic blade shapes and indicating that PC1 is strongly correlated with $K_1$.}
  \label{fig:Traverse_PC1}
\end{figure}

Figure~\ref{fig:Traverse_PC1} overlays this 2D PCA embedding with representative reconstructions obtained by traversing the PC1 direction. Along the traversal, we compare the ground-truth point clouds (left) to the corresponding DeepSDF reconstructions (right). Our goal is to assess whether PC1 aligns with any original design parameter. We focus on $K_1$, defined as the ratio of the small-diameter {($\frac{SD}{BSD}$)} of the top surface to that of the bottom surface. As we move along PC1, $K_1$ increases monotonically from $K_1=0.2$ to $K_1=0.8$ and the corresponding blade shapes transition from more conical/tapered geometries (low ratio) to more prismatic/rectangular profiles (high ratio). This consistent morphological trend indicates that PC1 is strongly correlated with the physically meaningful taper ratio $K_1$, providing an interpretable axis within the learned latent manifold.

\begin{figure}[H]
  \centering

  \includegraphics[width=0.65\linewidth]{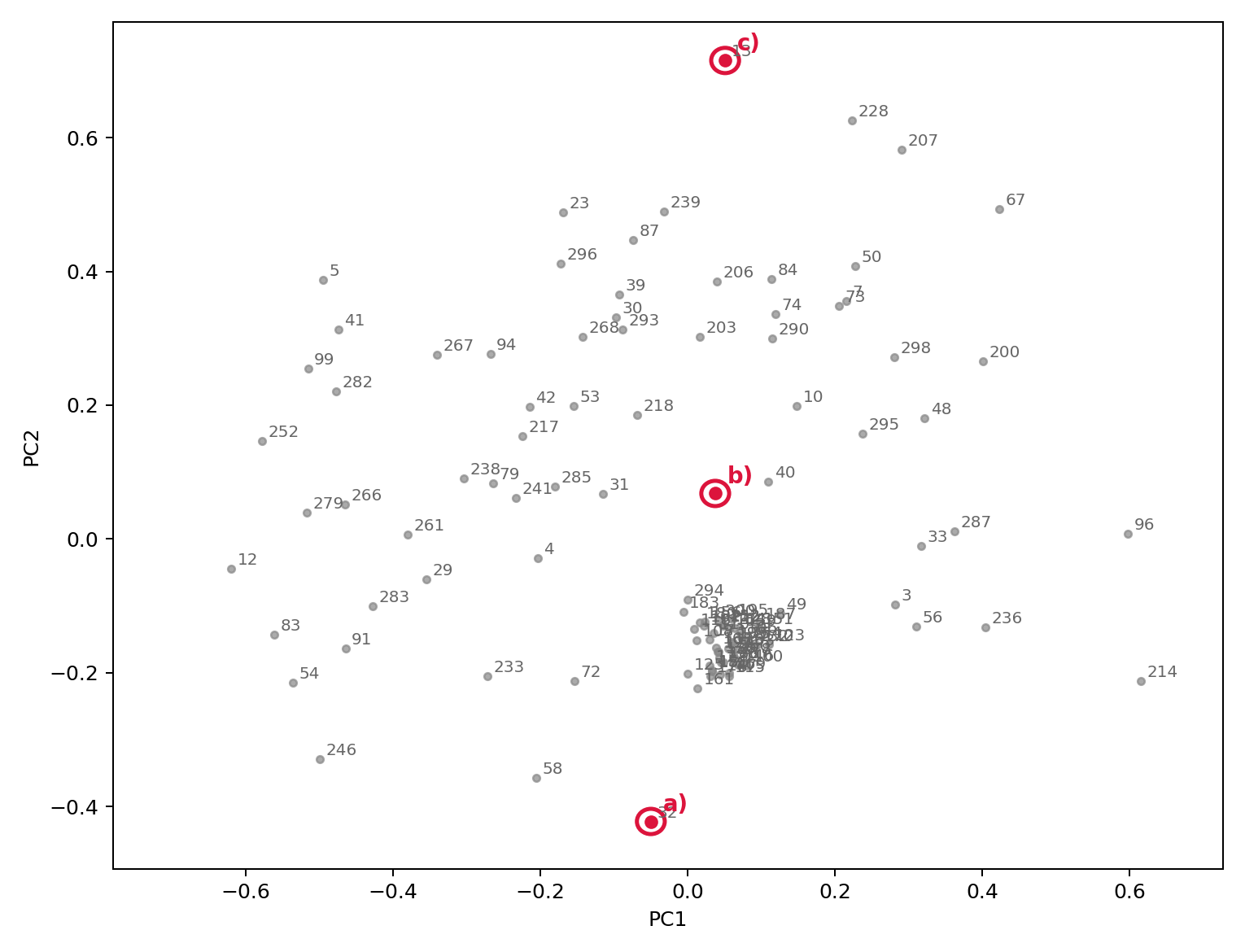}

  \vspace{0.75em}

  \begin{subfigure}[t]{0.45\linewidth}
    \centering
    \includegraphics[width=\linewidth]{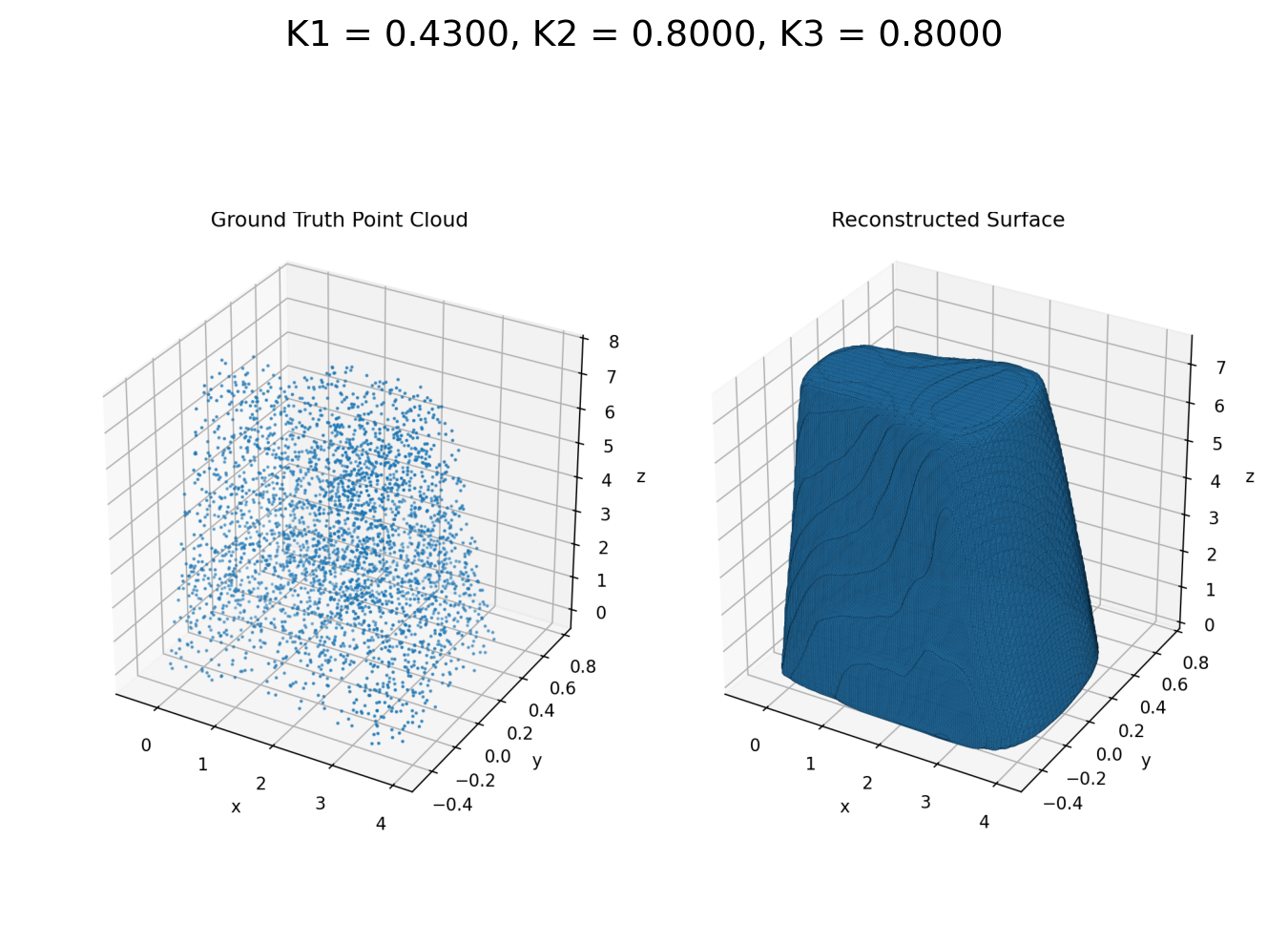}
    \caption{}
  \end{subfigure}
  \hfill
  \begin{subfigure}[t]{0.45\linewidth}
    \centering
    \includegraphics[width=\linewidth]{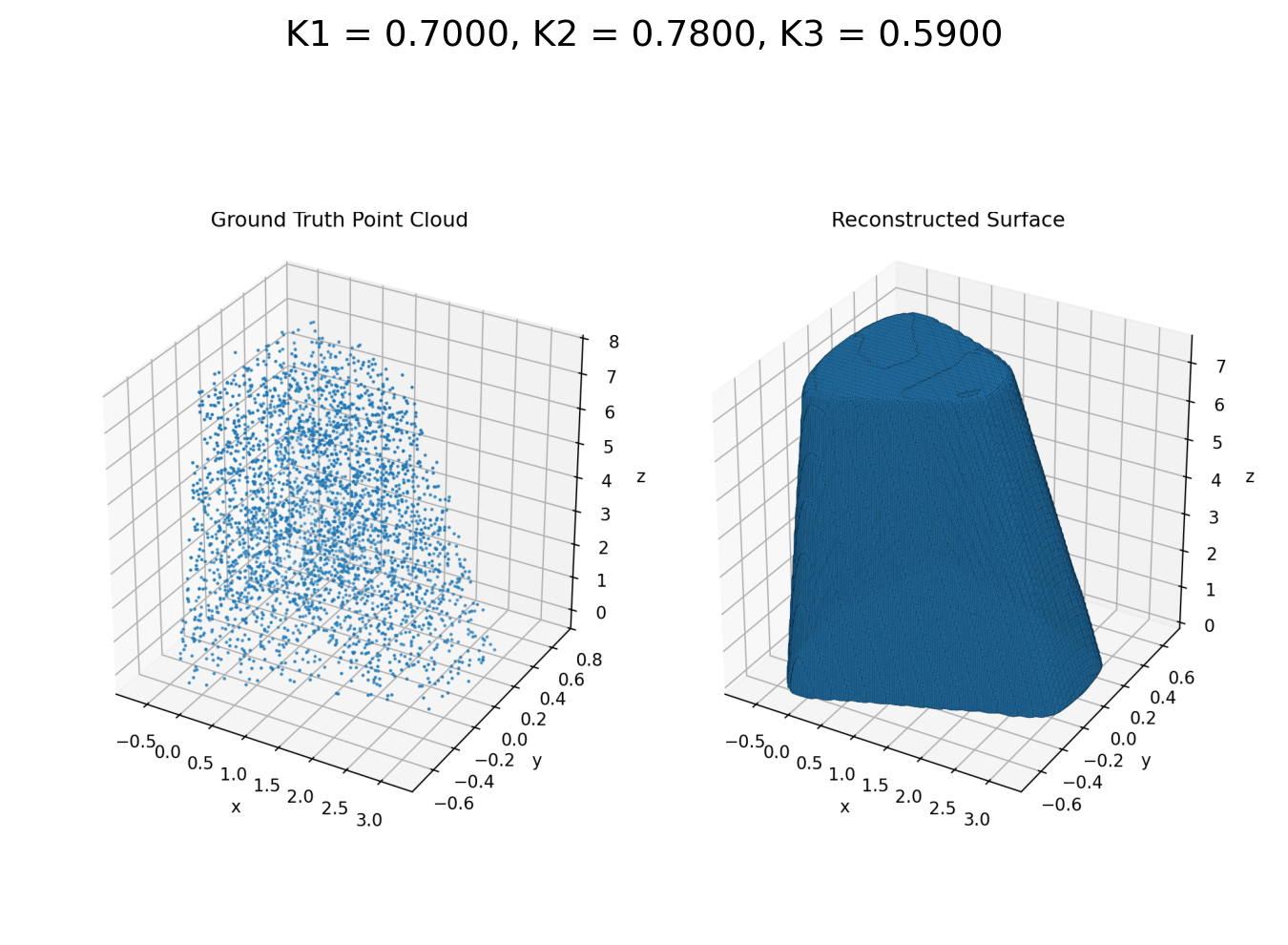}
    \caption{}
  \end{subfigure}
  \hfill
  \begin{subfigure}[t]{0.45\linewidth}
    \centering
    \includegraphics[width=\linewidth]{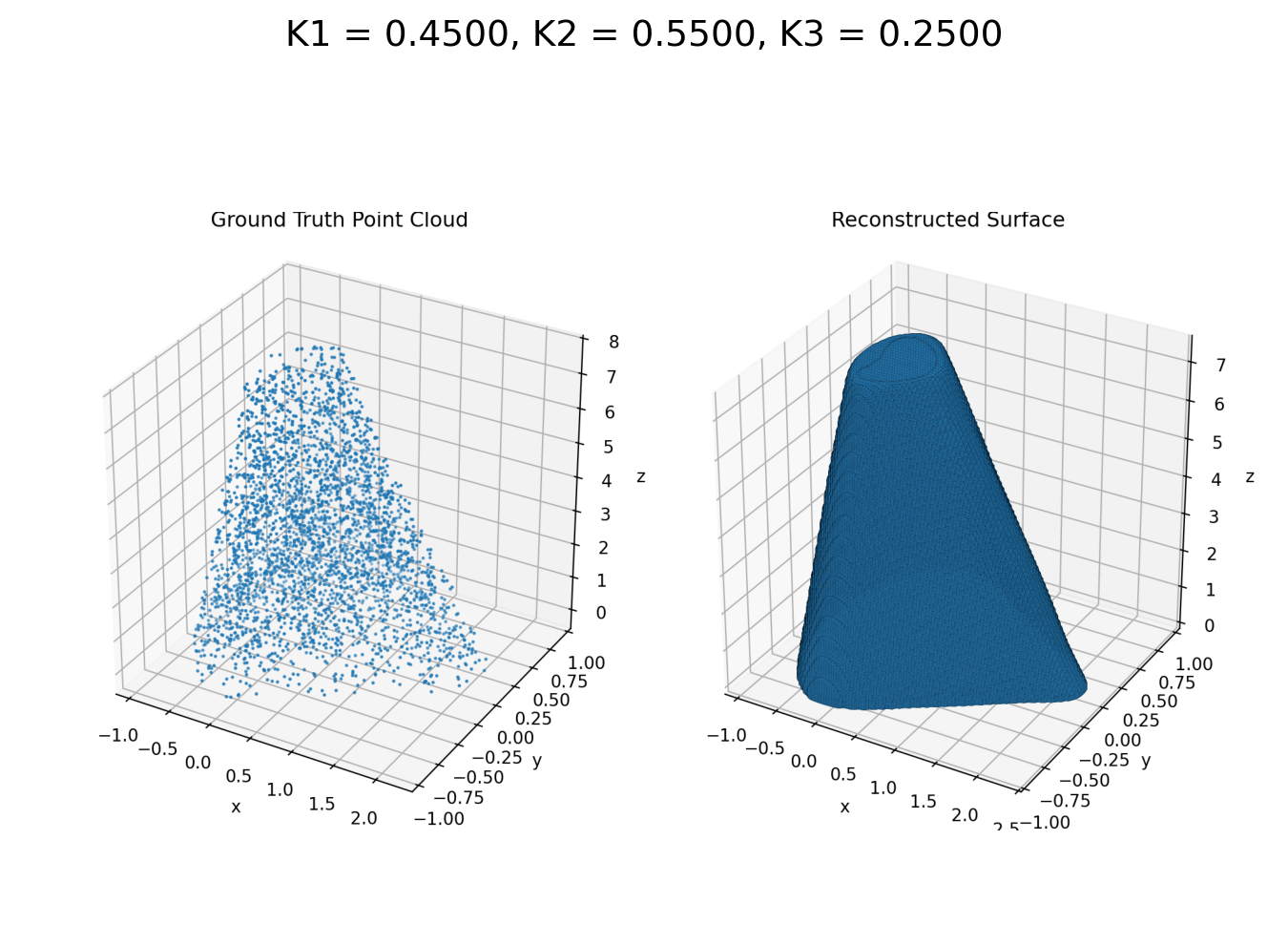}
    \caption{}
  \end{subfigure}

  \caption{PC2 traversal and correlation with $K_3$ for (a) Design 13, (b) Design 50, (c) Design 32. Three decoded samples ordered by increasing PC2 coordinate; $K_3$ ($\frac{CL}{BCL}$) decreases from 0.8 to 0.2, shifting from rectangular/prismatic to trapezoidal geometries and indicating strong alignment of PC2 with $K_3$.}
  \label{fig:Traverse_PC2}
\end{figure}

Figure~\ref{fig:Traverse_PC2} repeats the traversal analysis along the PC2 axis. Moving from bottom to top (increasing PC2 coordinate), the design parameter $K_3$, defined as the ratio of chord length on the top surface to that on the bottom surface {($\frac{CD}{BCD}$)} decreases monotonically from $K_3=0.8$ to $K_3=0.2$. Visually, the corresponding reconstructions transition from more rectangular/prismatic blades (high top-to-bottom chord ratio) to more trapezoidal blades, where the top-surface chord is markedly shorter than the bottom-surface chord. Other gross features remain comparatively stable along this path, isolating the chord-ratio effect. This monotonic morphing indicates that PC2 is strongly aligned with $K_3$, complementing the PC1–$K_1$ alignment described above. Taken together, the first two principal directions provide interpretable coordinates within the learned latent manifold that map to salient parameters from the original design parameterization.

\begin{figure}[h!]
    \centering
    \begin{subfigure}[b]{0.49\textwidth}
        \centering
        \includegraphics[width=\textwidth]{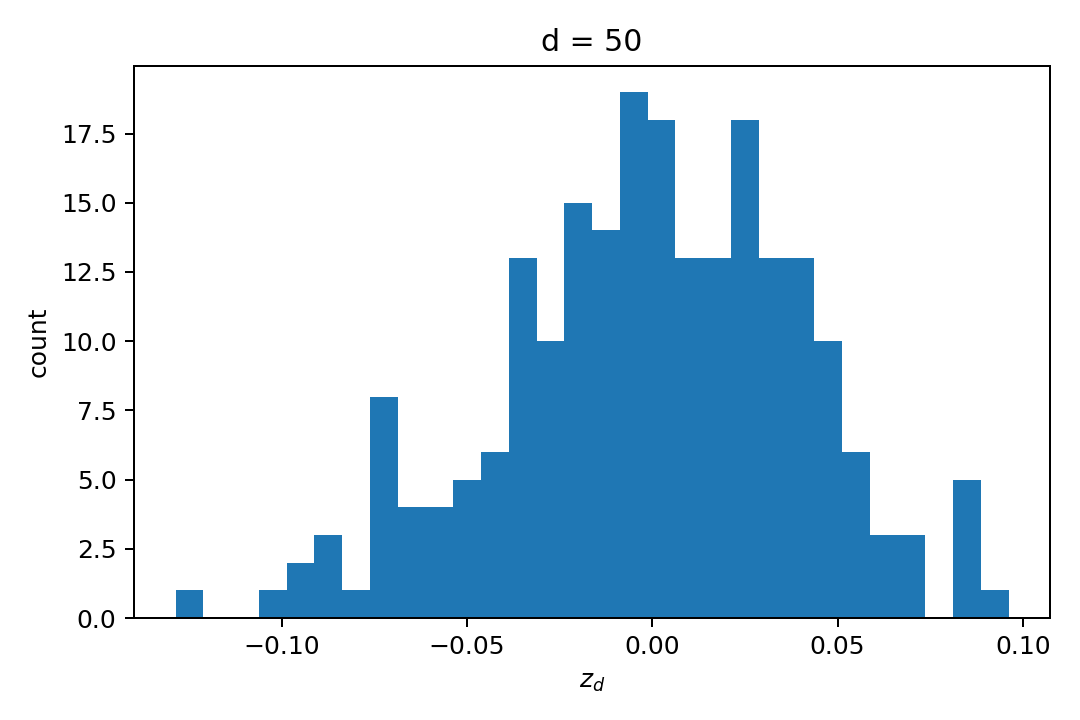}
        \caption{}
    \end{subfigure}
    \hfill
    \begin{subfigure}[b]{0.49\textwidth}
        \centering
        \includegraphics[width=\textwidth]{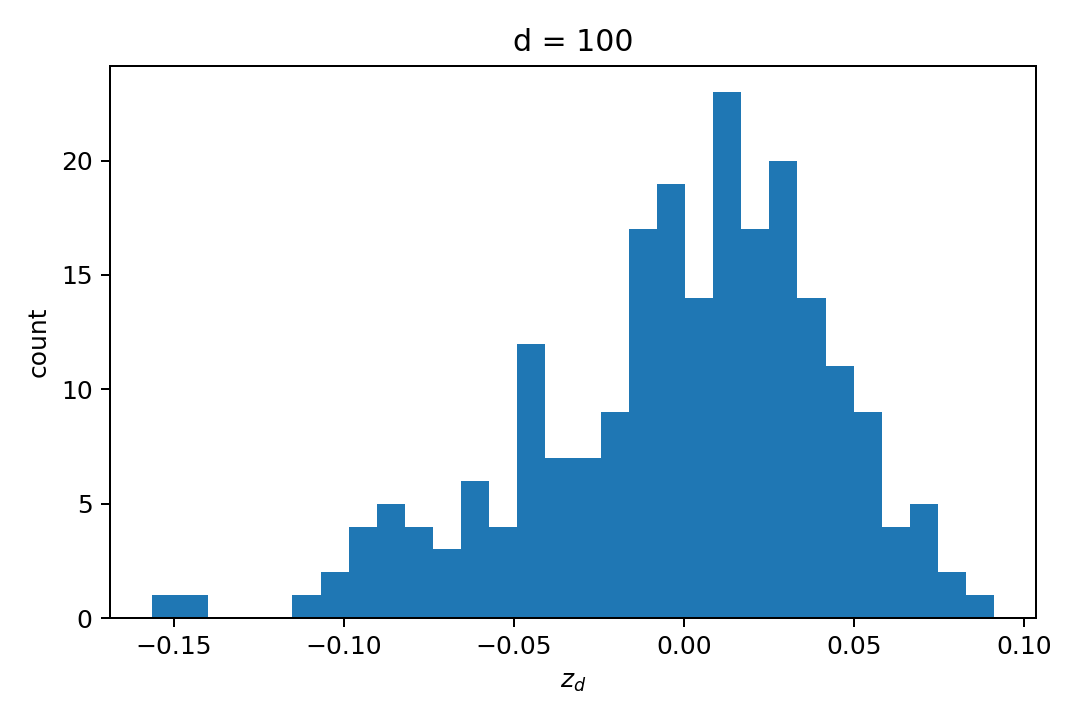}
        \caption{}
    \end{subfigure}

    \begin{subfigure}[b]{0.49\textwidth}
        \centering
        \includegraphics[width=\textwidth]{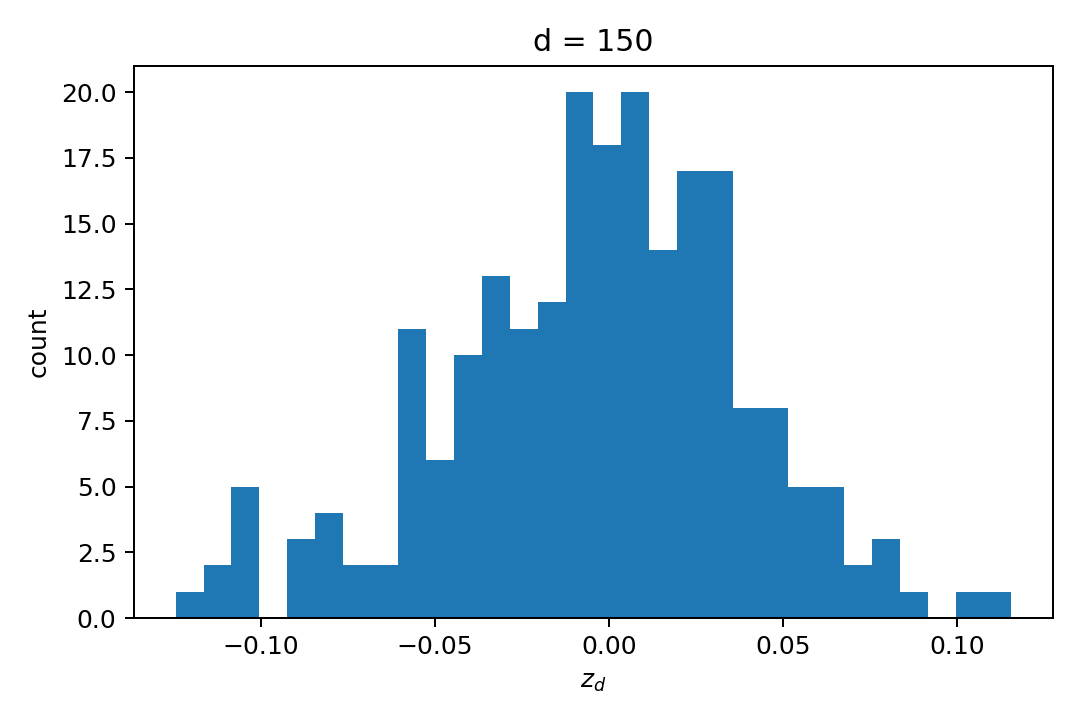}
        \caption{}
    \end{subfigure}
    \hfill
    \begin{subfigure}[b]{0.49\textwidth}
        \centering
        \includegraphics[width=\textwidth]{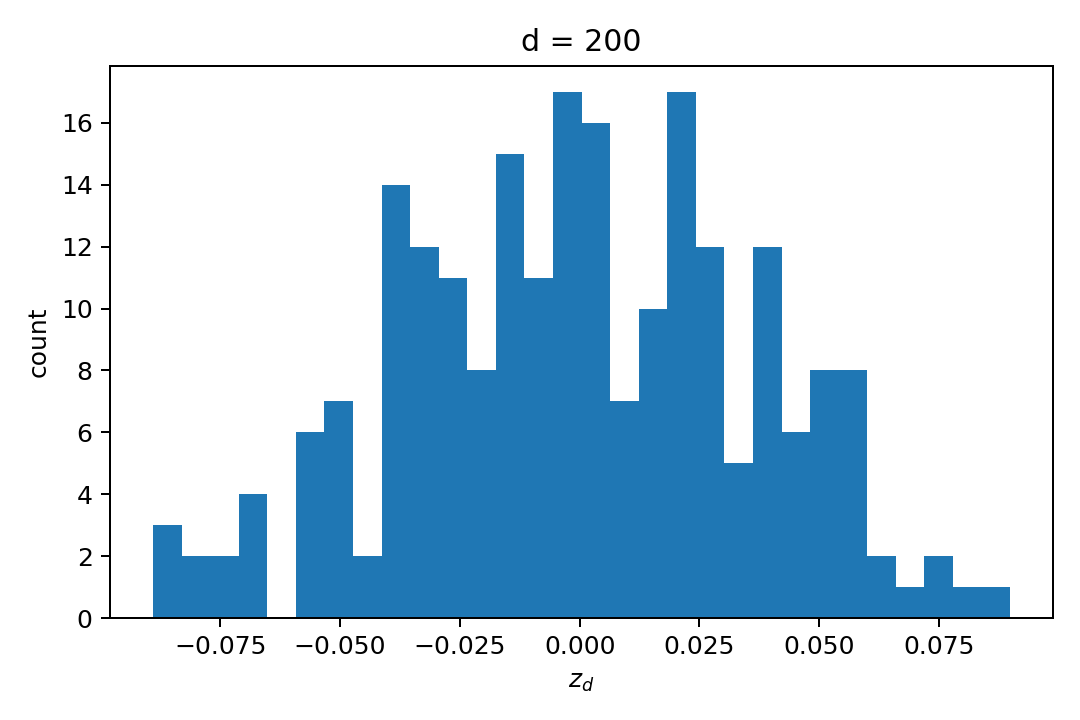}
        \caption{}
    \end{subfigure}
    
    \caption{Training-set latent code distribution at selected dimensions (a) d=50, (b) d=100, (c) d=150, (d) d=200. Histograms of latent coordinates ($z_d$) over ($n_{\text{train}}=222$) designs; each marginal is well-approximated by ($\mathcal N(0,0.1)$), consistent with the Gaussian prior used for code initialization/regularization.}
    \label{fig:Latent_distribution_train}
\end{figure}

We also examine the empirical distribution of individual latent coordinates across designs for both the training and test sets. Figure~\ref{fig:Latent_distribution_train} shows histograms of the coordinates $z_d$ at indices $d\in\{50,100,150,200\}$ for the training set. Across all plotted coordinates, the latent codes are well-approximated by Gaussian distributions with mean $\approx 0$ and variance $\approx 0.1$, matching the Gaussian prior used to initialize and regularize the latent codes during DeepSDF training. This alignment indicates that, given joint optimization of decoder weights and codes, the decoder expressivity is sufficient to represent the training designs while keeping codes near the prior; only small, near-Gaussian variations in $z$ are needed to span the observed design variability.

\begin{figure}[h!]
    \centering
    \begin{subfigure}[b]{0.49\textwidth}
        \centering
        \includegraphics[width=\textwidth]{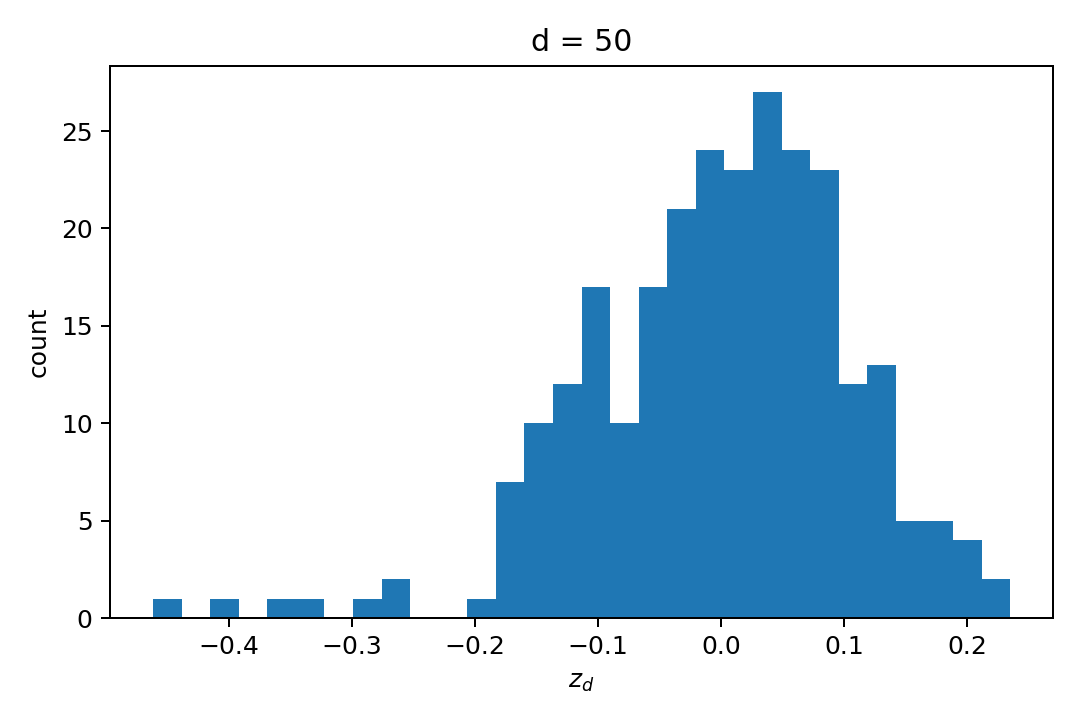}
        \caption{}
    \end{subfigure}
    \hfill
    \begin{subfigure}[b]{0.49\textwidth}
        \centering
        \includegraphics[width=\textwidth]{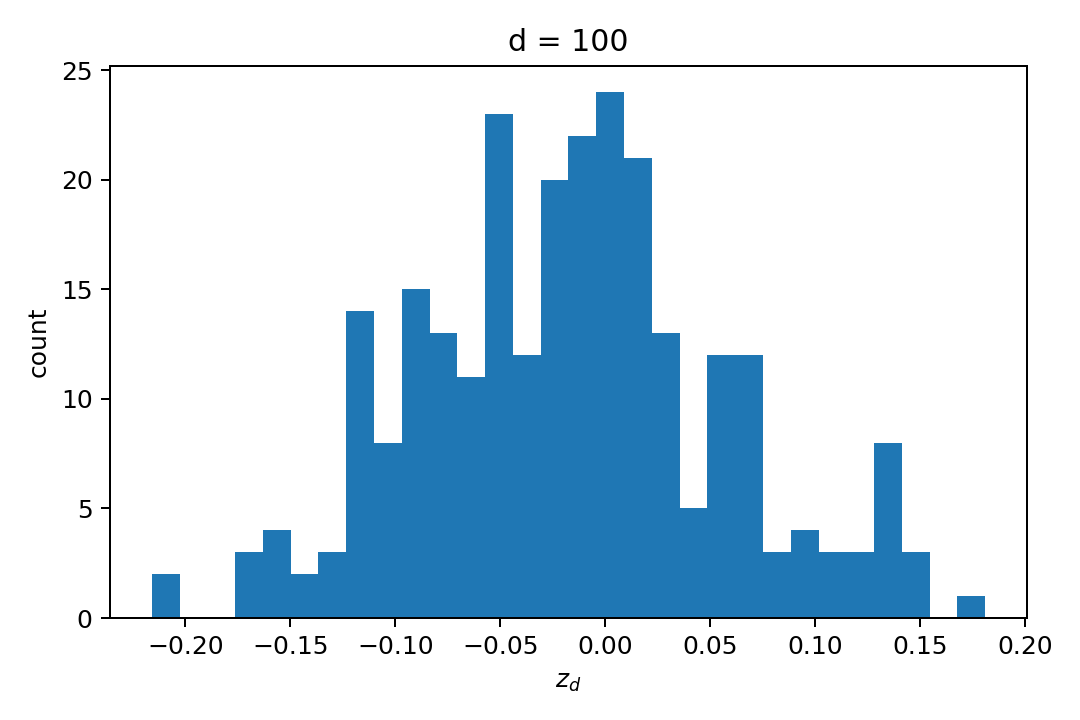}
        \caption{}
    \end{subfigure}
    
    \begin{subfigure}[b]{0.49\textwidth}
        \centering
        \includegraphics[width=\textwidth]{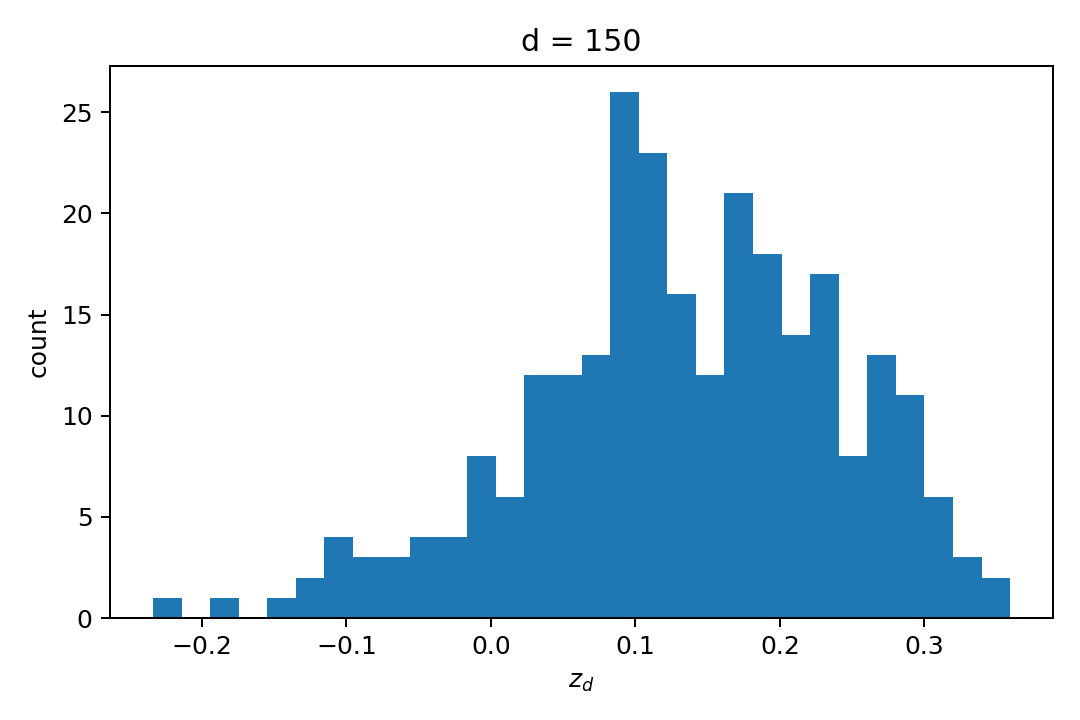}
        \caption{}
    \end{subfigure}
    \hfill
    \begin{subfigure}[b]{0.49\textwidth}
        \centering
        \includegraphics[width=\textwidth]{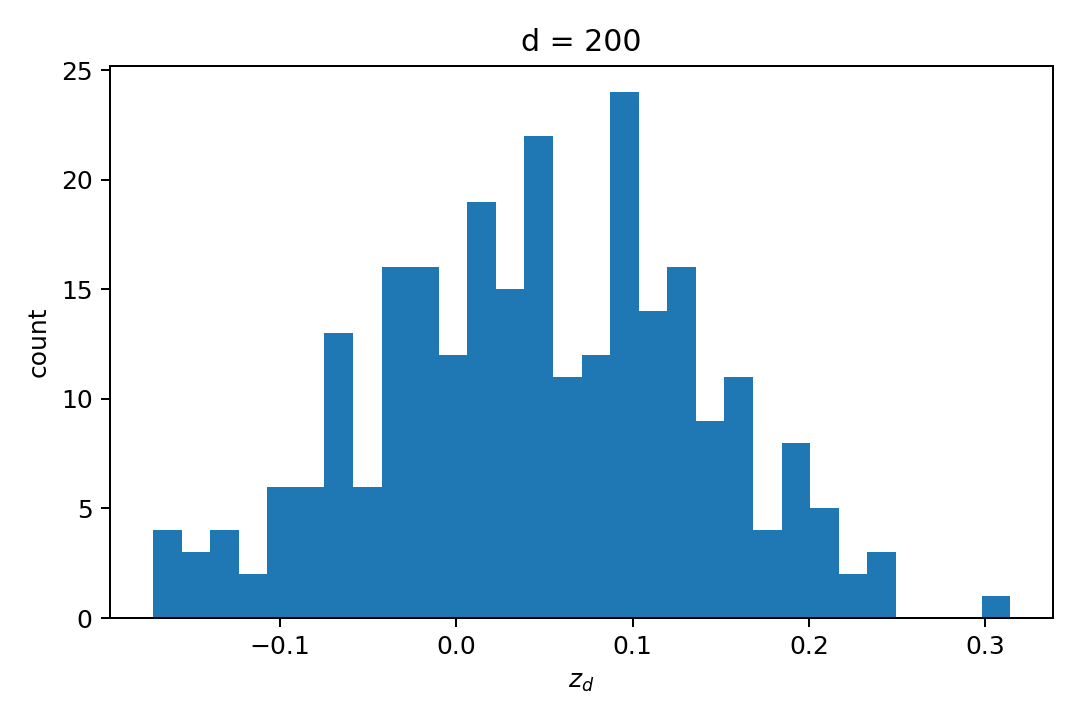}
        \caption{}
    \end{subfigure}
    
    \caption{Test-set latent code distribution at selected latent dimensions (a) d=50, (b) d=100, (c) d=150, (d) d=200. Histograms of latent coordinates ($z_d$) over ($n_{\text{test}}=300$) designs; means remain near zero while variances broaden relative to training, reflecting larger latent-code excursions with a frozen decoder.}
    \label{fig:Latent_distribution_test}
\end{figure}

Figure~\ref{fig:Latent_distribution_test} reports the same distributions for the test set. The means remain near zero, but the variances broaden. This variance increase is explained by the fact that at test time the decoder weights are frozen, so fitting previously unseen shapes relies solely on adjusting the latent code, which can require larger deviations away from the prior to compensate for limited decoder flexibility on out-of-sample geometries. Any mild domain shift (or higher geometric complexity/noise) in the test shapes will further push codes outward, even when an $L_2$ prior is retained during test-time optimization. The broader test-set distributions are therefore consistent with the flatter distance metric distribution observed earlier (Figure.~\ref{fig:distance_metric_test}).

In summary, the learned latent space is approximately Gaussian and well-centered, with training-set codes closely following the prior and test-set codes showing modest variance increase due to the frozen-decoder constraint. Consequently, fitting a multivariate Gaussian to the training codes is a reasonable choice for generative sampling\cite{kingma2014vae}.

\subsection{Generative Designs}

In this section, novel blade geometries are synthesized from the learned, approximately Gaussian, well-centered latent space characterized in the previous section. We consider (i) unconditional generation via: (a) linear interpolation between training codes $\mathbf z_a,\mathbf z_b$ using $\mathbf z(\alpha)=(1-\alpha)\mathbf z_a+\alpha\,\mathbf z_b$, and (b) independent Gaussian sampling by fitting a diagonal normal $\mathcal N(\mu,\mathrm{diag}(\sigma^2))$ to the training codes  and decoding samples $\mathbf z\sim\mathcal N(\mu,\mathrm{diag}(\sigma^2))$. We also examine (ii) conditional generation, learning a neural map $g_\phi:\mathbb R^3\!\to\!\mathbb R^{256}$ from maximum directional strains $(\varepsilon_x,\varepsilon_y,\varepsilon_z)$ to latent codes. For a target $\varepsilon^{*}$, we decode $\mathbf z^{*}=g_\phi(\varepsilon^{*})$ with the fixed DeepSDF decoder. For unconditional generation, we assess geometric plausibility qualitatively via side-by-side visualizations and for conditional generation, we quantify accuracy using the surface distance metric (Eq.~\ref{eq:Distance Metric}) between target shapes and decoded outputs.

\subsubsection{Unconditional Generation Through Linear Interpolation Between Latents}

 Unconditioned designs are generated by linear interpolation between two training codes $\mathbf z_a,\mathbf z_b\in\mathbb R^{256}$. For $\alpha\in[0,1]$, the latent path is
\begin{equation}
    \mathbf z(\alpha)=(1-\alpha)\,\mathbf z_a+\alpha\,\mathbf z_b,
\end{equation}
and each blended code is decoded by the fixed DeepSDF decoder $f_\theta:\mathbb R^{256}\times\mathbb R^3\!\to\!\mathbb R$ to a surface given by the zero level set
\begin{equation}
    \mathcal{S}(\alpha)=\big\{\,\mathbf x\in\mathbb R^3:\; f_\theta\big(\mathbf z(\alpha),\mathbf x\big)=0\,\big\}.
\end{equation}

\begin{figure}[H]
    \centering
    \begin{subfigure}[b]{0.4\textwidth}
        \centering
        \includegraphics[width=\textwidth]{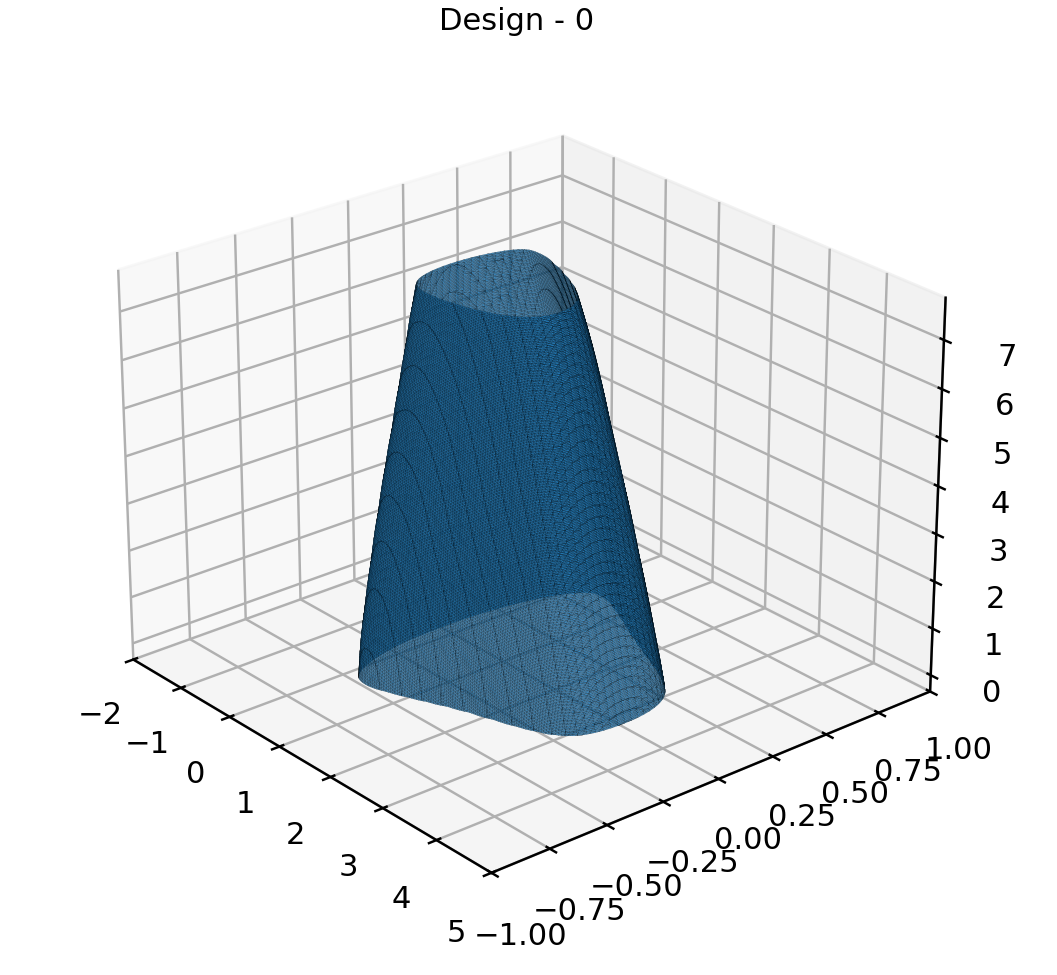}
        \caption{}
    \end{subfigure}
    \begin{subfigure}[b]{0.4\textwidth}
        \centering
        \includegraphics[width=\textwidth]{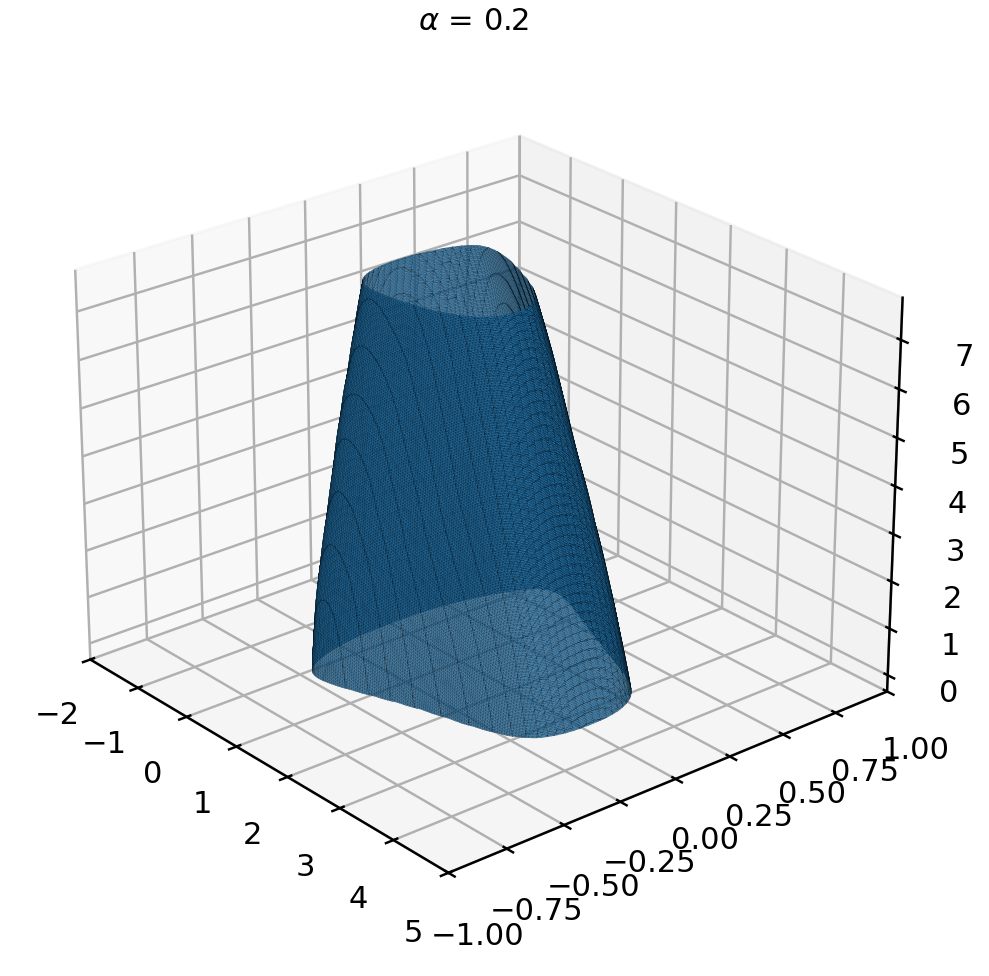}
        \caption{}
    \end{subfigure}

    \begin{subfigure}[b]{0.4\textwidth}
        \centering
        \includegraphics[width=\textwidth]{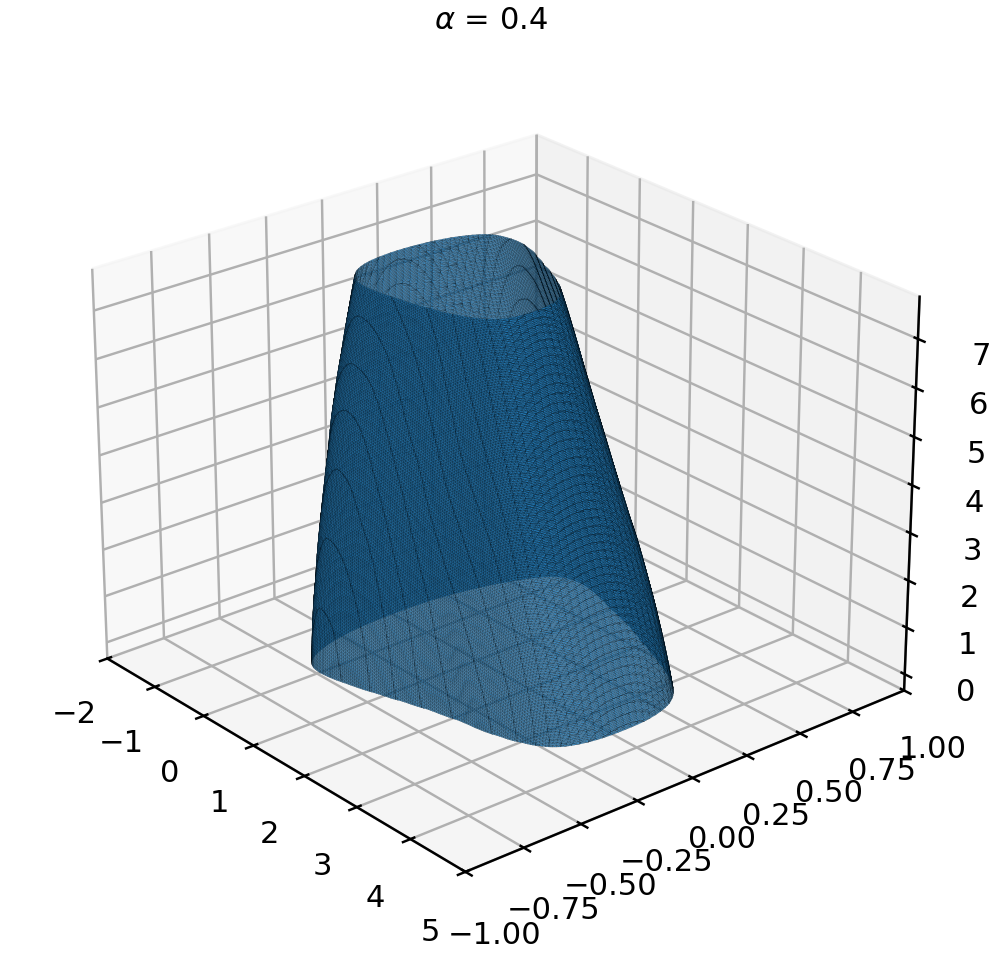}
        \caption{}
    \end{subfigure}
    \begin{subfigure}[b]{0.4\textwidth}
        \centering
        \includegraphics[width=\textwidth]{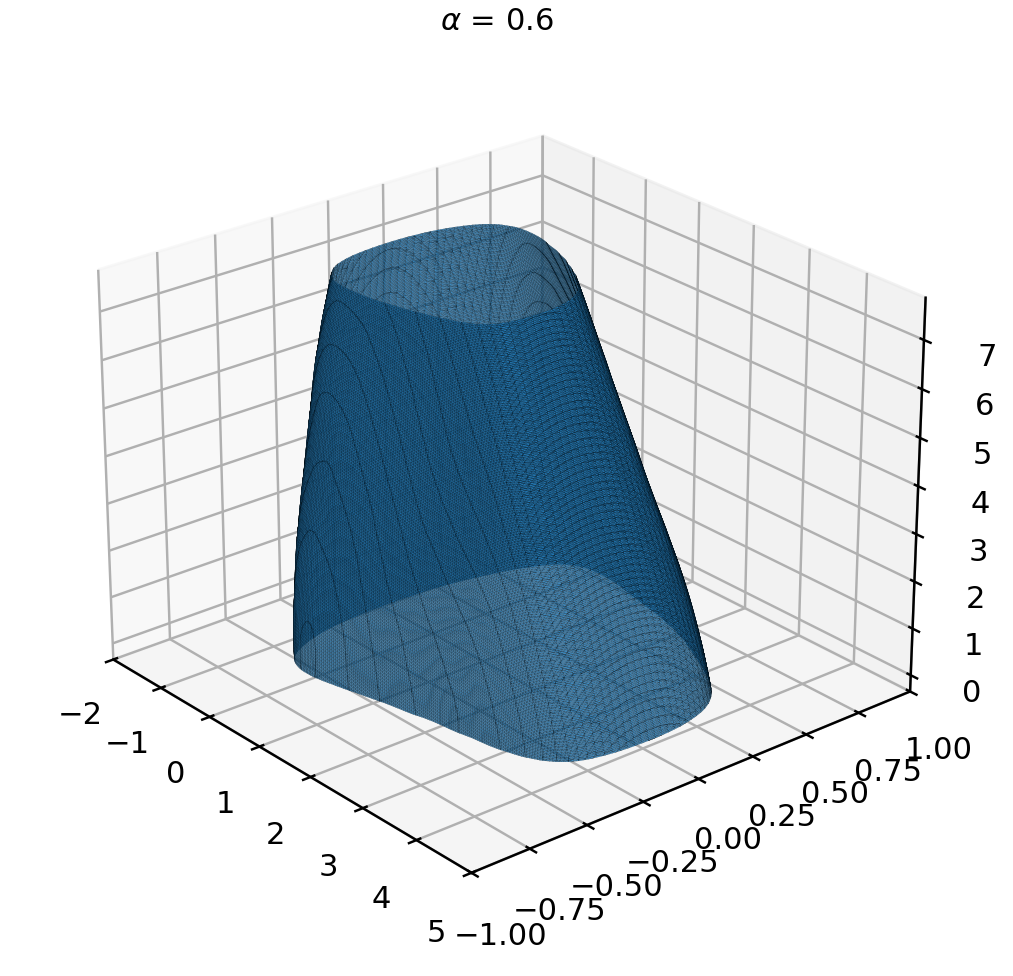}
        \caption{}
    \end{subfigure}

    \begin{subfigure}[b]{0.4\textwidth}
        \centering
        \includegraphics[width=\textwidth]{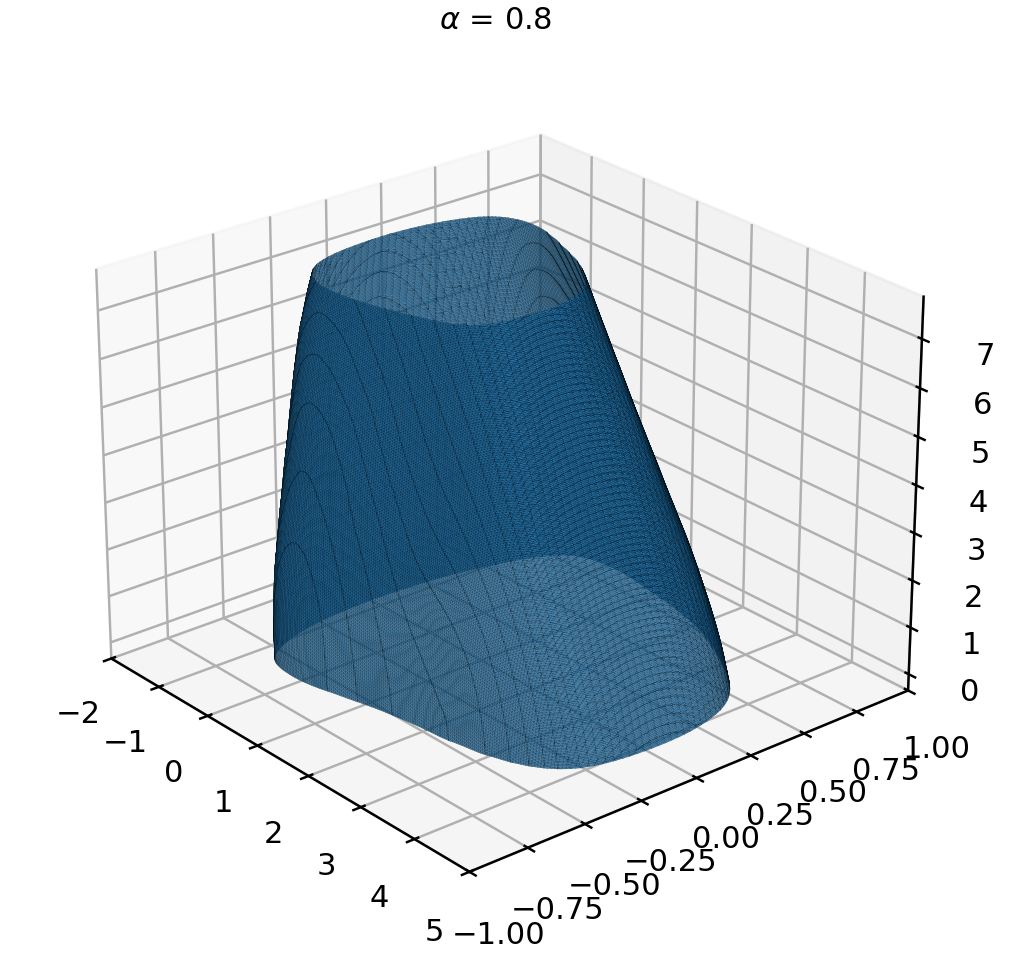}
        \caption{}
    \end{subfigure}
    \begin{subfigure}[b]{0.4\textwidth}
        \centering
        \includegraphics[width=\textwidth]{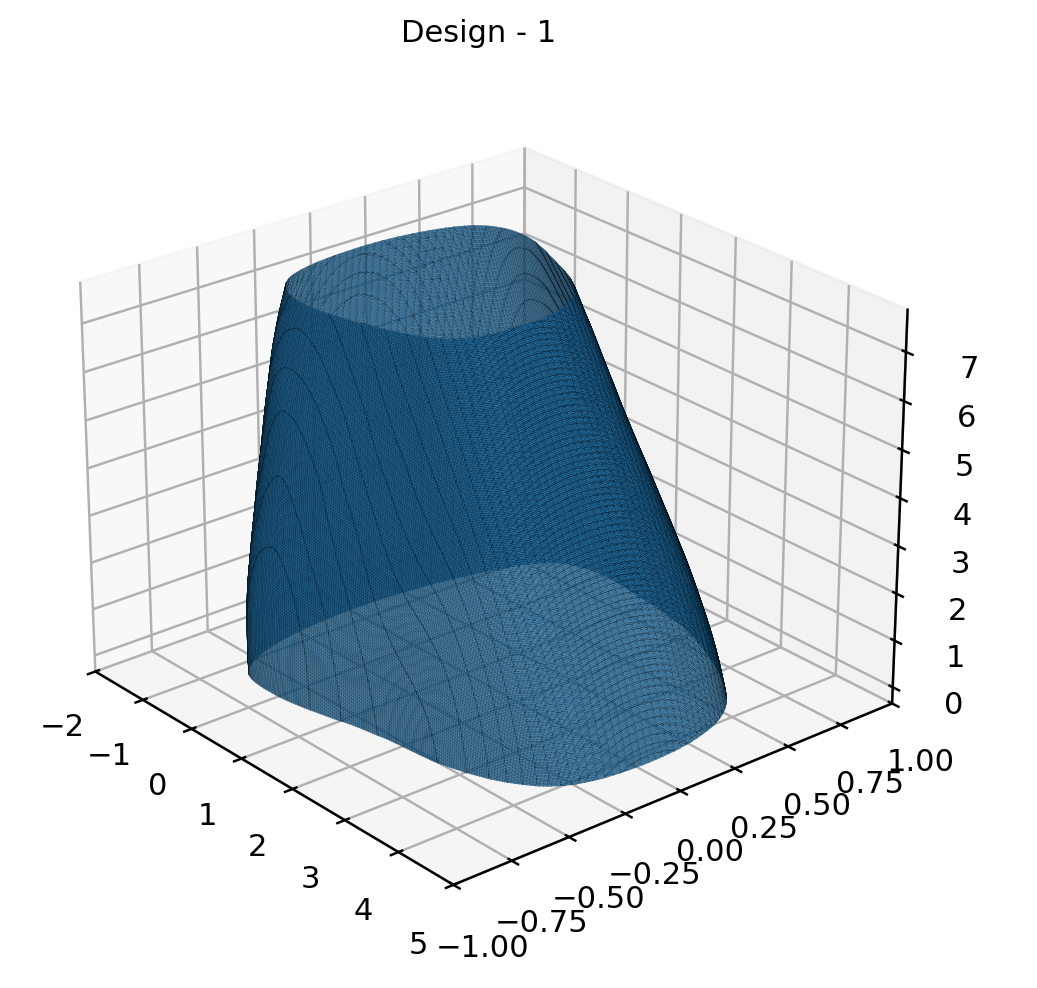}
        \caption{}
    \end{subfigure}
    \caption{unconditional generation via latent interpolation : (a) Design-0; (b–e) decoded blades for $\alpha\in\{0.2,0.4,0.6,0.8\}$ along $\mathbf z(\alpha)=(1-\alpha)\mathbf z_0+\alpha\,\mathbf z_1$; (f) Design-1. Intermediates transition smoothly from Design-0 toward Design-1.}
    \label{fig:Generation_linear_interpolation}
\end{figure}

Figure~\ref{fig:Generation_linear_interpolation} shows designs generated by interpolating between Design-0 and Design-1 with $\alpha\in\{0.2,0.4,0.6,0.8\}$. Relative to Design-0, Design-1 has larger volume and longer chord lengths on both the top and bottom surfaces. As $\alpha$ increases from 0.2 to 0.8, the decoded blades transition monotonically from a geometry close to Design-0 toward Design-1, with smooth, visually plausible changes in chord length and overall thickness. These results indicate that linear code mixing,
\begin{equation}
    \mathbf z(\alpha)=(1-\alpha)\,\mathbf z_0+\alpha\,\mathbf z_1,
\end{equation}
traces a reasonable path on the learned manifold even when the endpoints are far apart in the original parameter space. The method also extends naturally to convex combinations of more than two designs, $\mathbf z=\sum_m w_m \mathbf z_m$ with $w_m\!\ge\!0,\ \sum_m w_m=1$, enabling multi-anchor interpolation. Collectively, this supports linear interpolation as a simple and effective baseline for unconditional design synthesis\cite{chen2019imnet,mittal2022autosdf}.

\subsubsection{Unconditional Generation Through Gaussian Sampling}

In this section, we synthesize designs by Gaussian sampling from the training-code distribution. Let $Z=\{\mathbf z_i\}_{i=1}^{M}\subset\mathbb R^{256}$ denote the training latents. We fit a diagonal normal by estimating per-dimension mean and variance,

\begin{equation}
    \hat\mu_d=\frac{1}{M}\sum_{i=1}^{M} z_{i,d},\qquad
\hat\sigma_d^{2}=\frac{1}{M-1}\sum_{i=1}^{M}\big(z_{i,d}-\hat\mu_d\big)^2,\quad d=1,\dots,256.
\end{equation}

New codes are drawn as
\begin{equation}
    \mathbf z \sim \mathcal N\!\big(\hat\mu,\;\mathrm{diag}(\tau^{2}\hat\sigma^{2})\big),
\quad\text{equivalently}\quad
z_d=\hat\mu_d+\tau\,\hat\sigma_d\,\epsilon_d,\;\epsilon_d\sim\mathcal N(0,1),
\end{equation}
where $\tau>0$ modulates exploration (diversity). Each sample is decoded by the fixed DeepSDF decoder $f_\theta$ to a surface defined by the zero level set
\begin{equation}
    \mathcal S(\mathbf z)=\{\mathbf x\in\mathbb R^{3}: f_\theta(\mathbf z,\mathbf x)=0\}.
\end{equation}

\begin{figure}[ht!]
    \centering
    \includegraphics[width=0.99\linewidth]{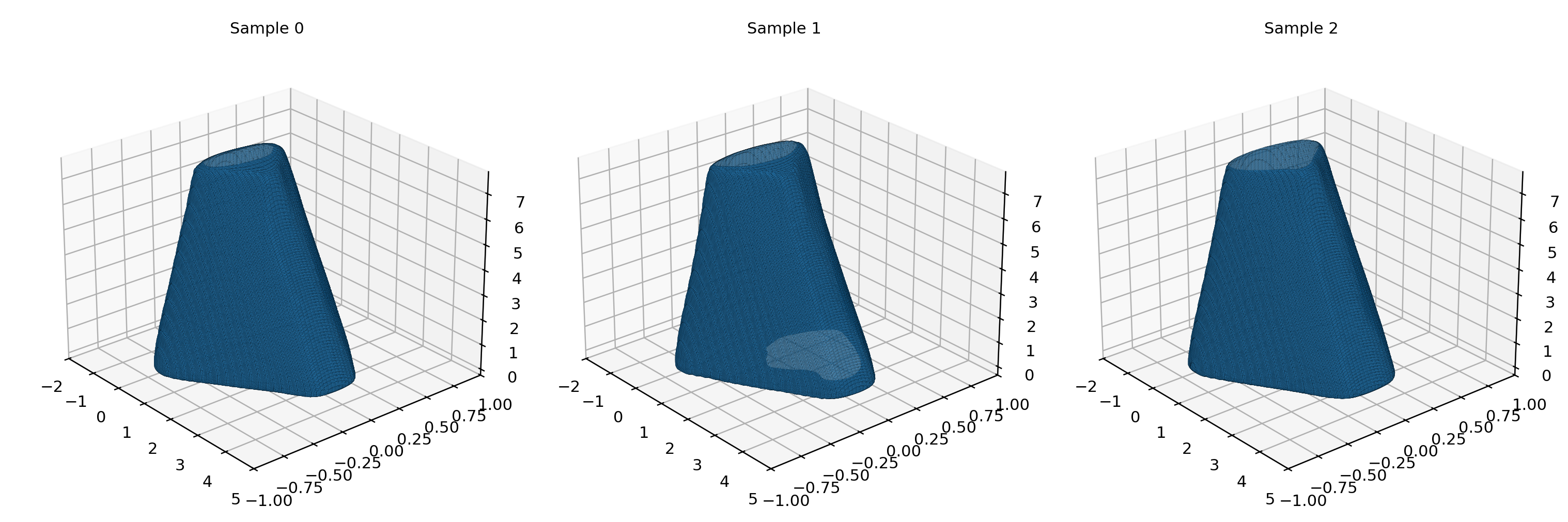}
    \caption{unconditional synthesis via diagonal-Gaussian sampling : Three decoded blades from independent draws $\mathbf z \sim \mathcal N(\hat\mu,\mathrm{diag}(\hat\sigma^{2}))$; samples exhibit smooth, plausible geometries consistent with the training manifold. }
    \label{fig:Generation_gaussian_sampling}
\end{figure}

Figure~\ref{fig:Generation_gaussian_sampling} displays decoded surfaces for three independent draws $\mathbf z\sim\mathcal N(\hat\mu,\mathrm{diag}(\hat\sigma^2))$ for $\tau=1$. The resulting geometries exhibit smooth, well-formed surfaces without obvious artifacts, illustrating that sampling near the empirical prior yields plausible designs. In practice, samples that remain within one standard deviation of $\hat\mu$ tend to preserve realism, while more extreme draws can deviate from the observed design envelope. Together with linear interpolation, these results support the viability of unconditional generation via both convex mixing and prior-consistent Gaussian sampling \cite{achlioptas2018learning}.

\subsubsection{Conditional Generation}

Having shown that unconditional synthesis via latent interpolation and diagonal-Gaussian sampling yields smooth, plausible designs, we now condition generation on target mechanical response. A neural network (NN) map $g_{\phi}:\mathbb{R}^{3}\!\to\!\mathbb{R}^{256}$ is learned to predict the latent code from a design’s maximum directional strain triplet $\varepsilon=(\varepsilon_x,\varepsilon_y,\varepsilon_z)$. The parameters $\phi$ are fit on pairs $\{(\varepsilon_i,\mathbf z_i)\}$ by minimizing,
\begin{equation}
    \min_{\phi}\;\sum_i \big\|g_{\phi}(\varepsilon_i)-\mathbf z_i\big\|_2^{2}.
\end{equation}

The NN map is a compact MLP with two hidden layers (128 units each, ReLU). At inference, for a target $\varepsilon^\ast$ we compute $\mathbf z^*=g_{\phi}(\varepsilon^\ast)$ and decode with the fixed DeepSDF decoder $f_\theta$ to obtain the surface
\begin{equation}
    \mathcal S^*=\big\{\mathbf x\in\mathbb{R}^3:\; f_\theta(\mathbf z^*,\mathbf x)=0\big\}.
\end{equation}

\begin{figure}[h!]
    \centering
    \includegraphics[width=0.5\linewidth]{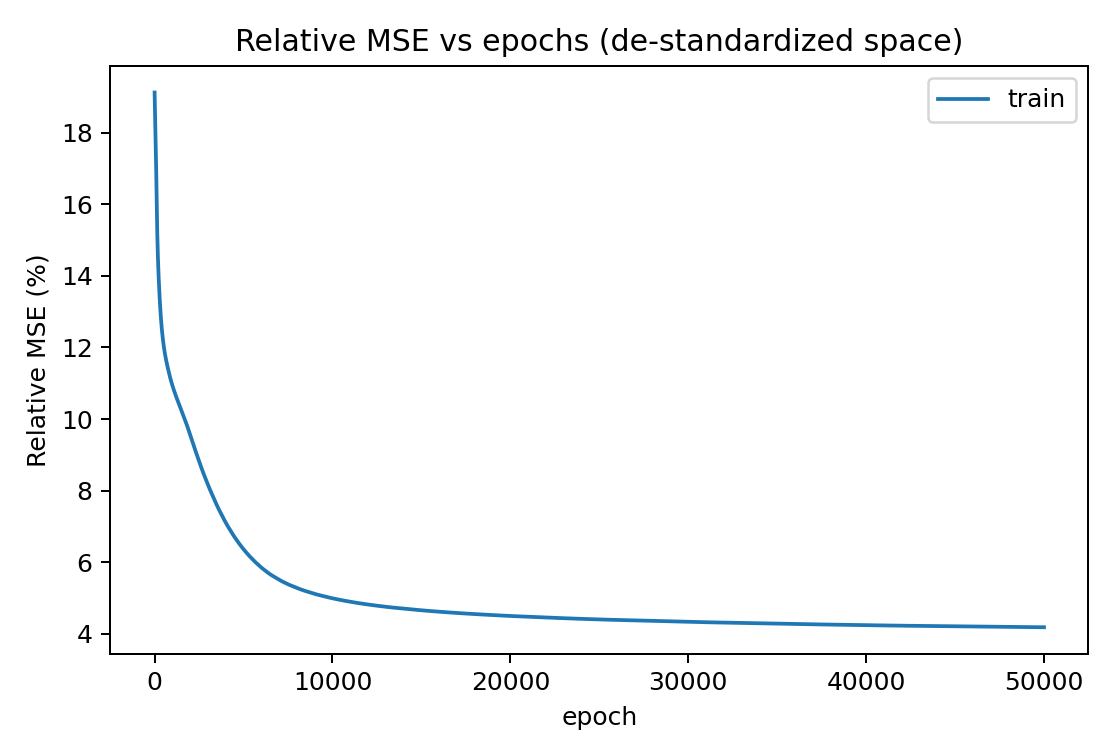}
    \caption{Training curve for the strain-to-latent NN-map $g_{\phi}$ (2×128) : NRMSE (\%) over $n_{\text{train}}=222$ blades falls to roughly 4\% by 40k epochs and then plateaus.}
    \label{fig:training_loss_nn_map}
\end{figure}

The strain-to-latent NN map was trained on the same $n_{\text{train}}=222$ blades used to train the DeepSDF model, using the optimized latent codes $\mathbf z_i$ as targets. As a scalar training metric, we report the normalized root-mean-squared error (NRMSE, \%) computed over the 256 latent dimensions. Figure~\ref{fig:training_loss_nn_map} shows that the NRMSE decreases steadily to $\sim 4\%$ by $\approx 40{,}000$ epochs and then stabilizes.

\begin{figure}[h!]
    \centering
    \begin{subfigure}[b]{0.75\textwidth}
        \centering
        \includegraphics[width=\textwidth]{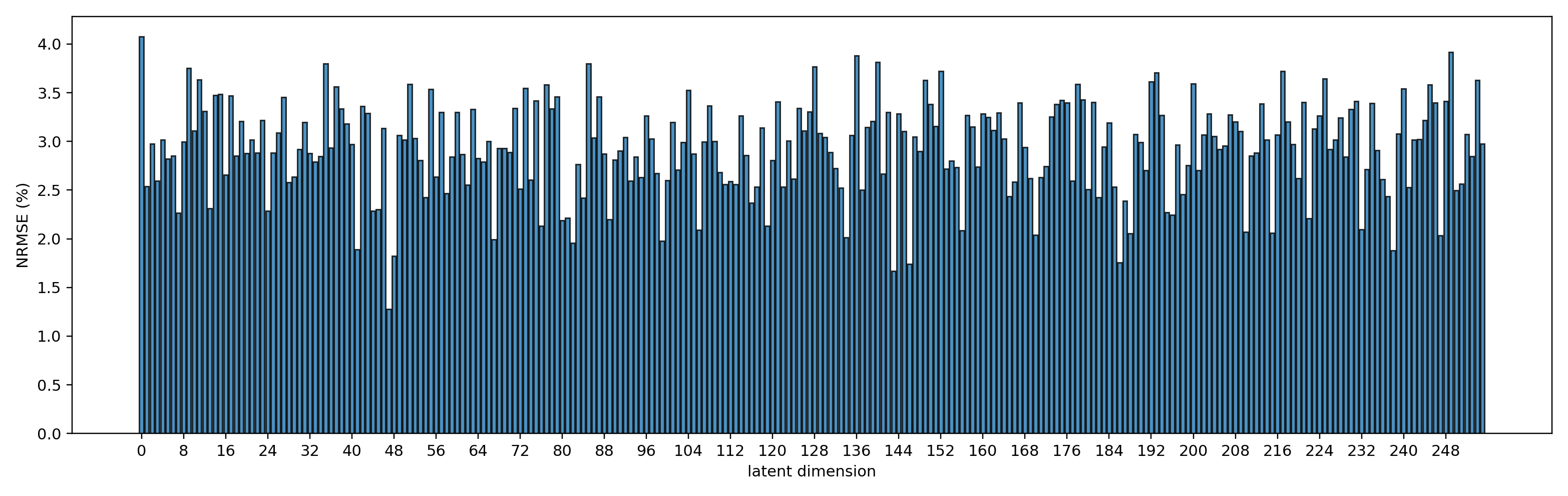}
        \caption{}
    \end{subfigure}
    \hfill
    \begin{subfigure}[b]{0.75\textwidth}
        \centering
        \includegraphics[width=\textwidth]{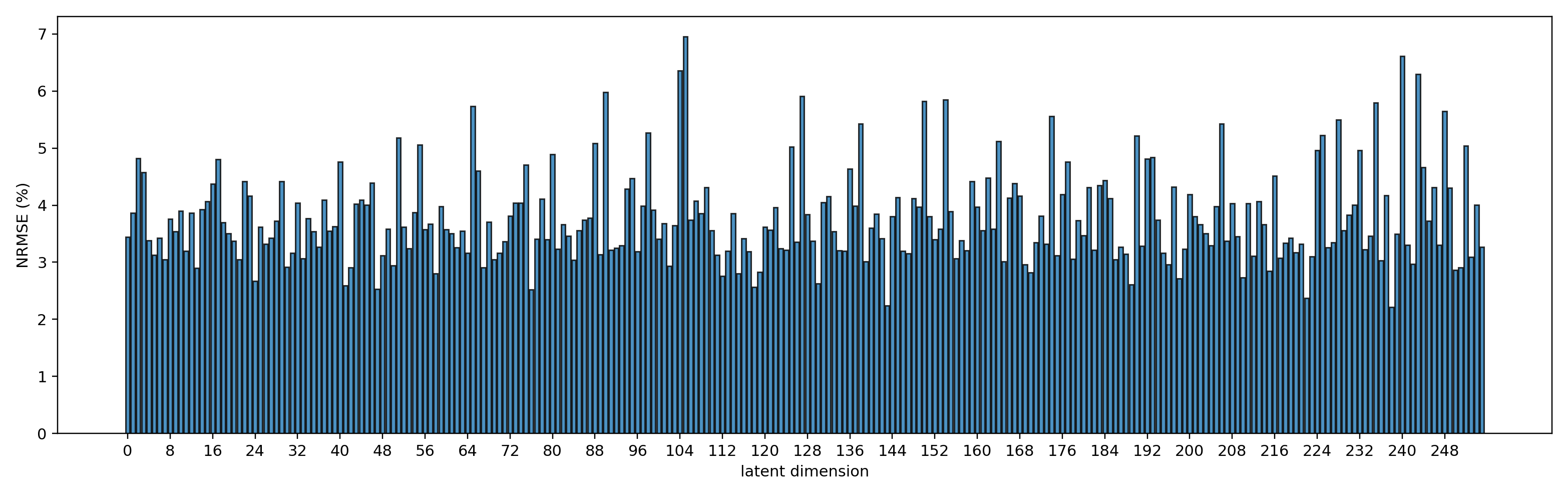}
        \caption{}
    \end{subfigure}
        
    \caption{Per-dimension NRMSE for the strain-to-latent NN-map ($g_{\phi}$) for (a) training ($n_{\text{train}}=222$), (b) test ($n_{\text{test}}=300$) sets.Distribution of NRMSE (\%) across 256 latent coordinates: training concentrates at (1\%)–(4); test shifts to (2)–(7), indicating a moderate generalization gap.}
    \label{fig:NRMSE_distribution}
\end{figure}

To assess the NN-map’s fidelity at a coordinate level, we examine the distribution of NRMSE across latent dimensions $d=1,\dots,256$ for both training and test sets (Figure~\ref{fig:NRMSE_distribution}). On training data, errors are relatively uniform across dimensions, with most coordinates in the $1\%$–$4\%$ NRMSE range and maxima aligning with the overall training minimum ($\sim\!4\%$). On the test set, the distribution remains broadly even by dimension but shifts upward to approximately $2\%$–$7\%$. This controlled shift suggests the NN-map generalizes reasonably well while exhibiting the expected out-of-sample increase in error, given the compact three dimesional strain input mapping to a 256 dimensional latent code.

\begin{figure}[h!]
    \centering
    \includegraphics[width=0.5\linewidth]{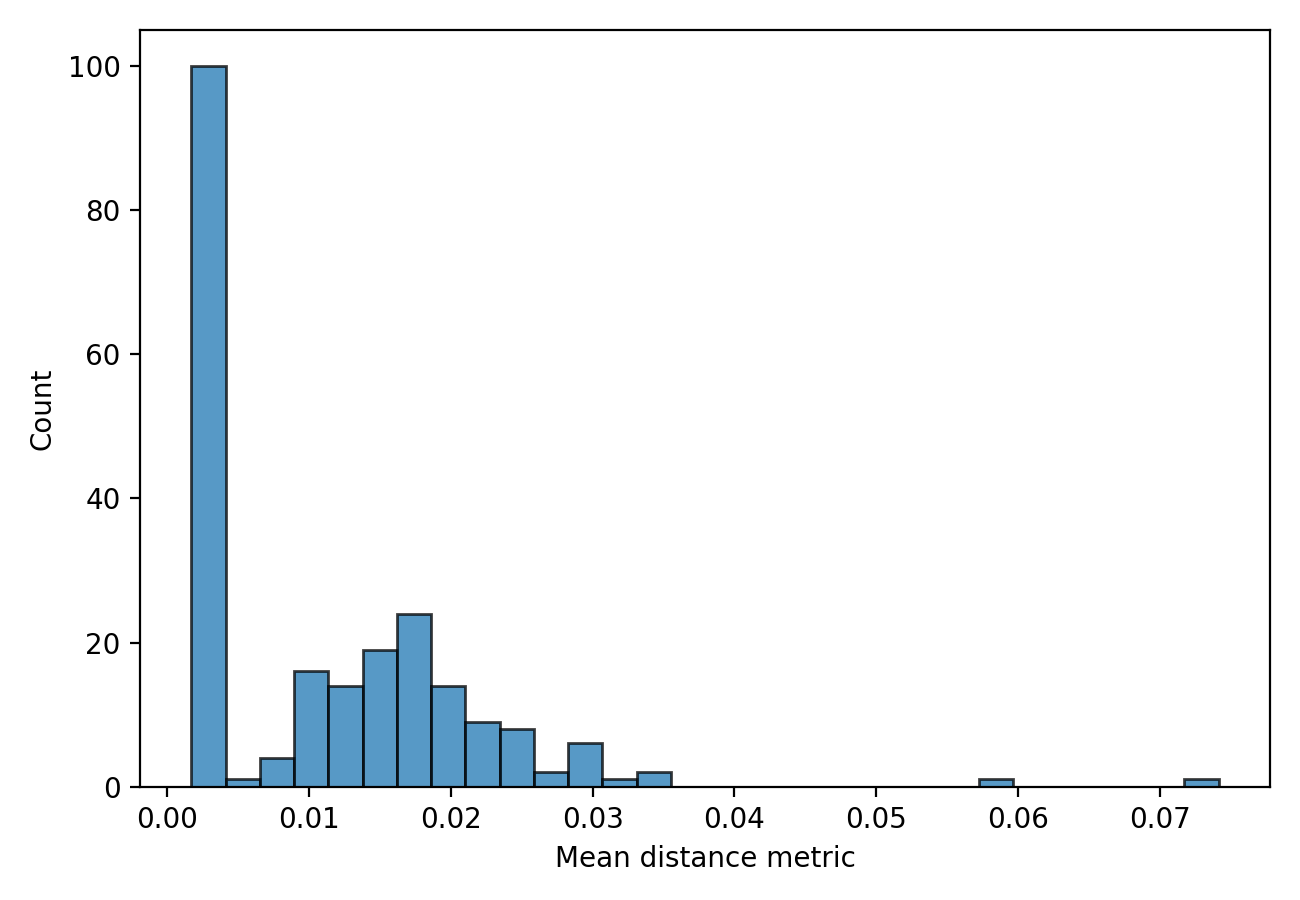}
    \caption{Distance between reconstructions from optimized latents and NN-map latents (training set) : Histogram of distance metric (Eq.~\ref{eq:Distance Metric}) between reconstructions from optimized latents and NN-map latents (training set): roughly 50\% near zero; majority in $0.01\!-\!0.04$; rare outliers $\sim\!0.07$ ($\sim\!$1\% of $D_{\max}$), indicating small additional discrepancy from the strain-to-latent mapping.}
    \label{fig:Distance_metric_nn_map}
\end{figure}

Complementing the per-dimension NRMSE analysis, we quantify the error introduced by mapping strains to latent codes by comparing, for each training design ($n_{\text{train}}=222$), the surface reconstructed from its optimized latent $\mathbf z_i$ with the surface reconstructed from the NN-map prediction $g_\phi(\varepsilon_i)$, using the directed surface distance in Eq.~\ref{eq:Distance Metric}. Figure~\ref{fig:Distance_metric_nn_map} shows the resulting distribution: roughly designs ($\sim\!$50\%) have distances effectively at 0 (numerical tolerance), indicating negligible discrepancy; most of the remainder lie in $0.01\!-\!0.04$, with a small tail near $0.07$. With $D_{\max}\!\approx\!7$, these correspond to sub-percent deviations for the bulk ($\lesssim $0.6\%) and rare near-1\% cases at the tail. Thus, the strain-to-latent mapping preserves geometry to within sub-percent error for most designs, with only a few outliers approaching the 1\% scale.

\begin{figure}[h!]
    \centering
    \begin{subfigure}[b]{0.49\textwidth}
        \centering
        \includegraphics[width=\textwidth]{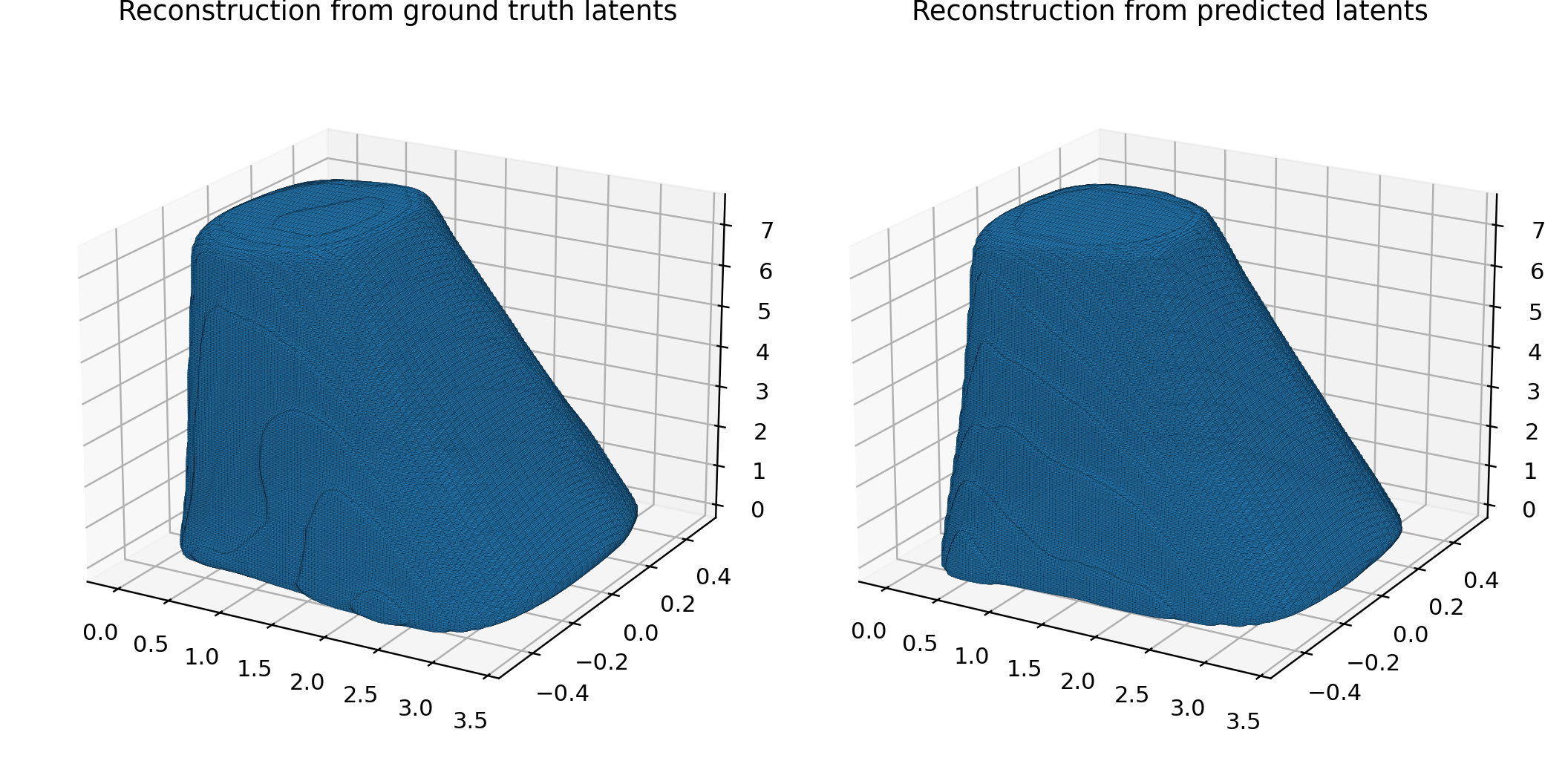}
        \caption{}
    \end{subfigure}
    \hfill
    \begin{subfigure}[b]{0.49\textwidth}
        \centering
        \includegraphics[width=\textwidth]{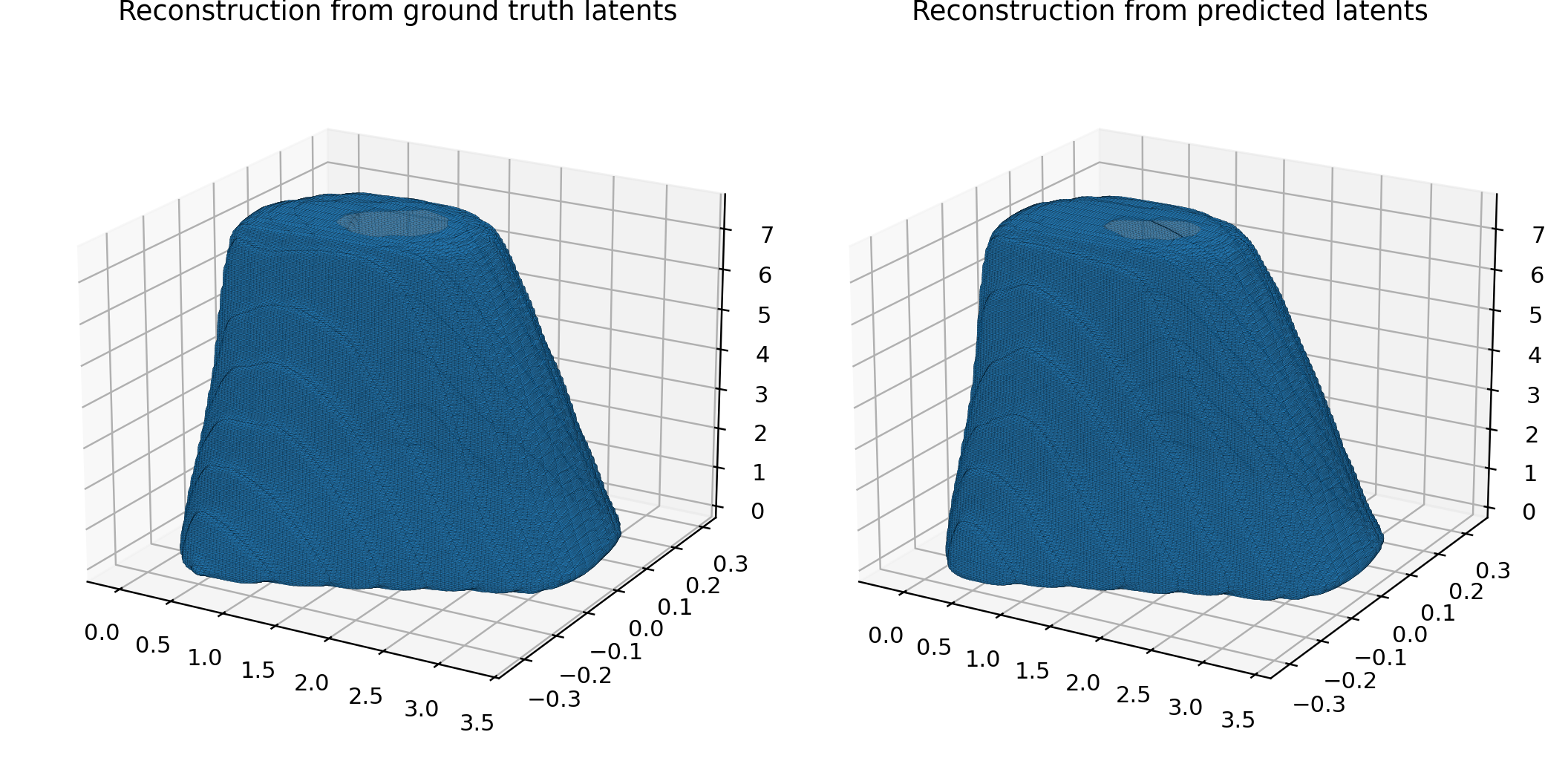}
        \caption{}
    \end{subfigure}
    
    \begin{subfigure}[b]{0.49\textwidth}
        \centering
        \includegraphics[width=\textwidth]{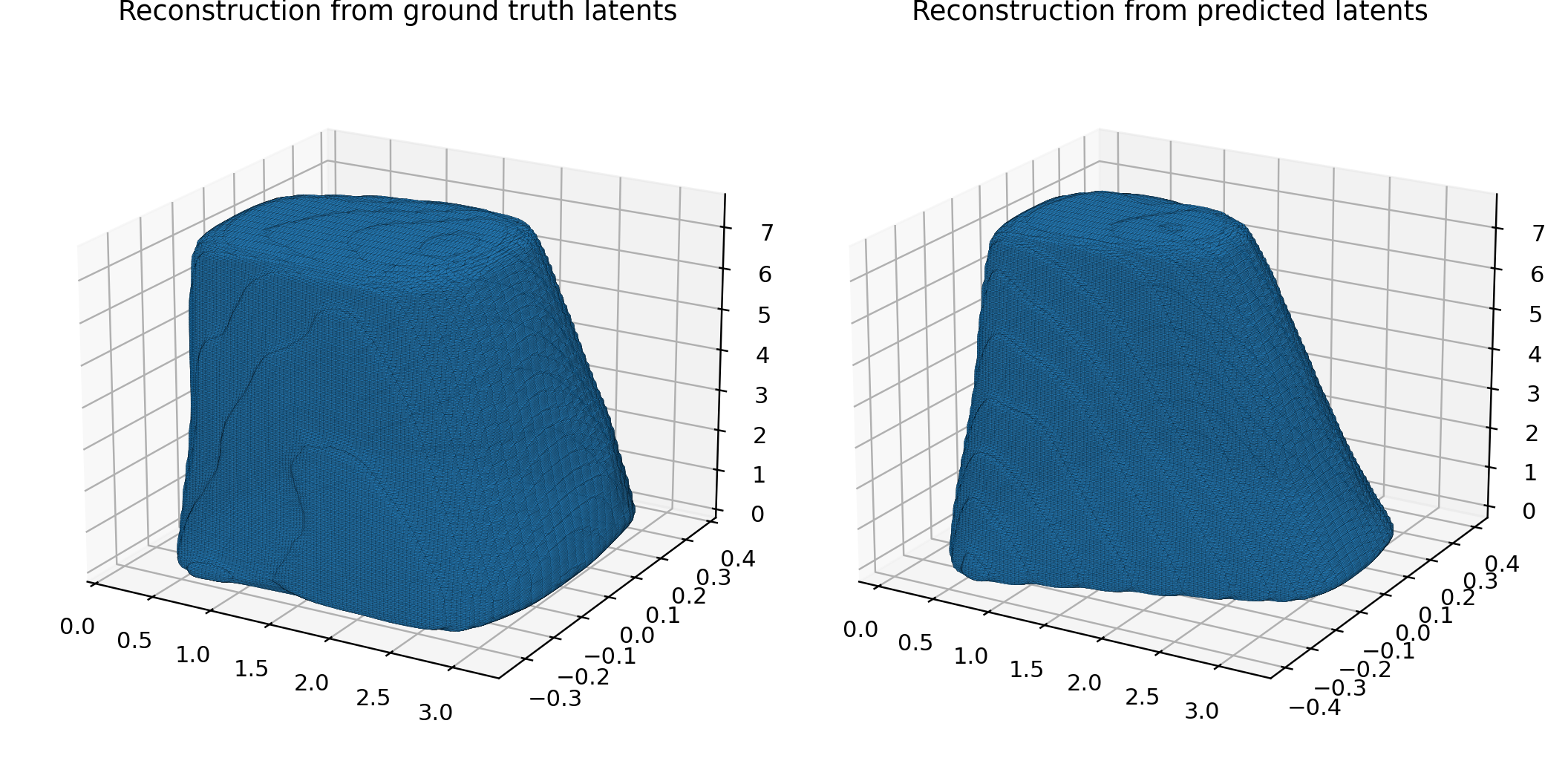}
        \caption{}
    \end{subfigure}

    \caption{Reconstructions at minimum strain extremes for (a) $\min \varepsilon_x$, (b) $\min \varepsilon_y$, (c) $\min \varepsilon_z$. NN-map decoded surfaces (left) versus reconstructions from optimized latents (right); each case uses the full strain triplet from the selected design as NN-map input. Agreement is close, with a small deviation in the ($\min \varepsilon_z$) example.}
    \label{fig:Comp_min_strains}
\end{figure}

\begin{figure}[ht!]
    \centering
    \begin{subfigure}[b]{0.49\textwidth}
        \centering
        \includegraphics[width=\textwidth]{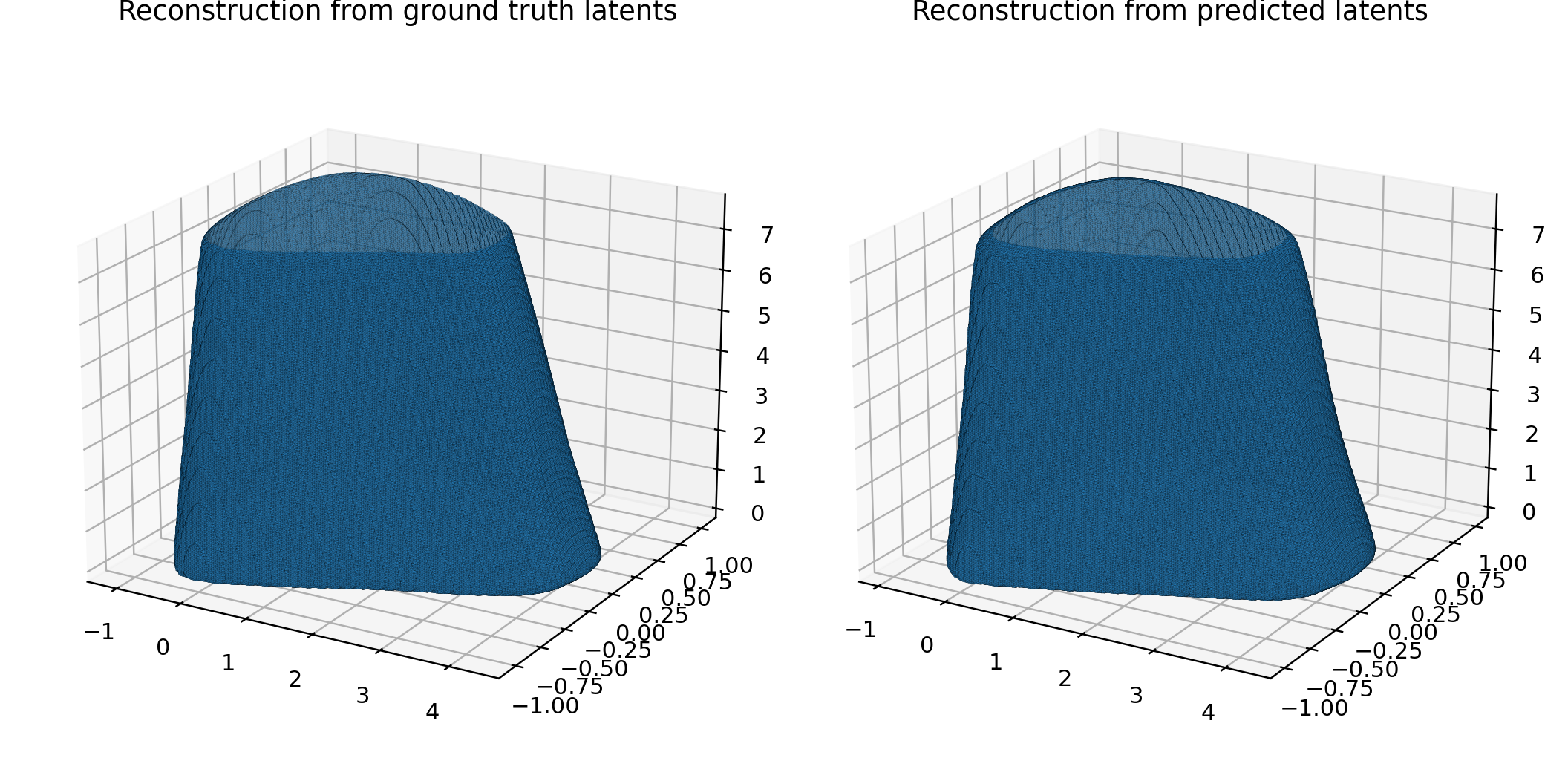}
        \caption{}
    \end{subfigure}
    \hfill
    \begin{subfigure}[b]{0.49\textwidth}
        \centering
        \includegraphics[width=\textwidth]{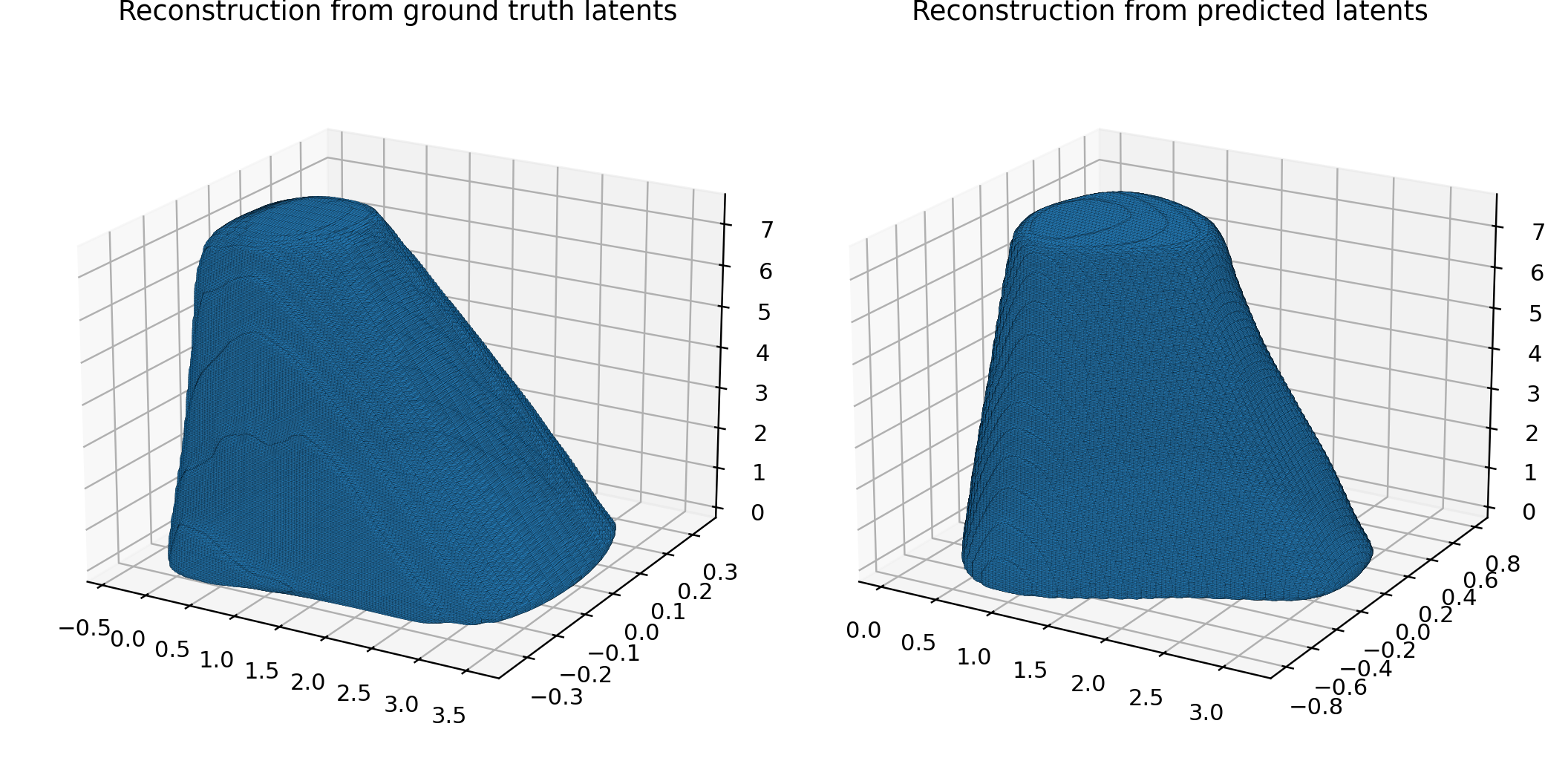}
        \caption{}
    \end{subfigure}
    
    \begin{subfigure}[b]{0.49\textwidth}
        \centering
        \includegraphics[width=\textwidth]{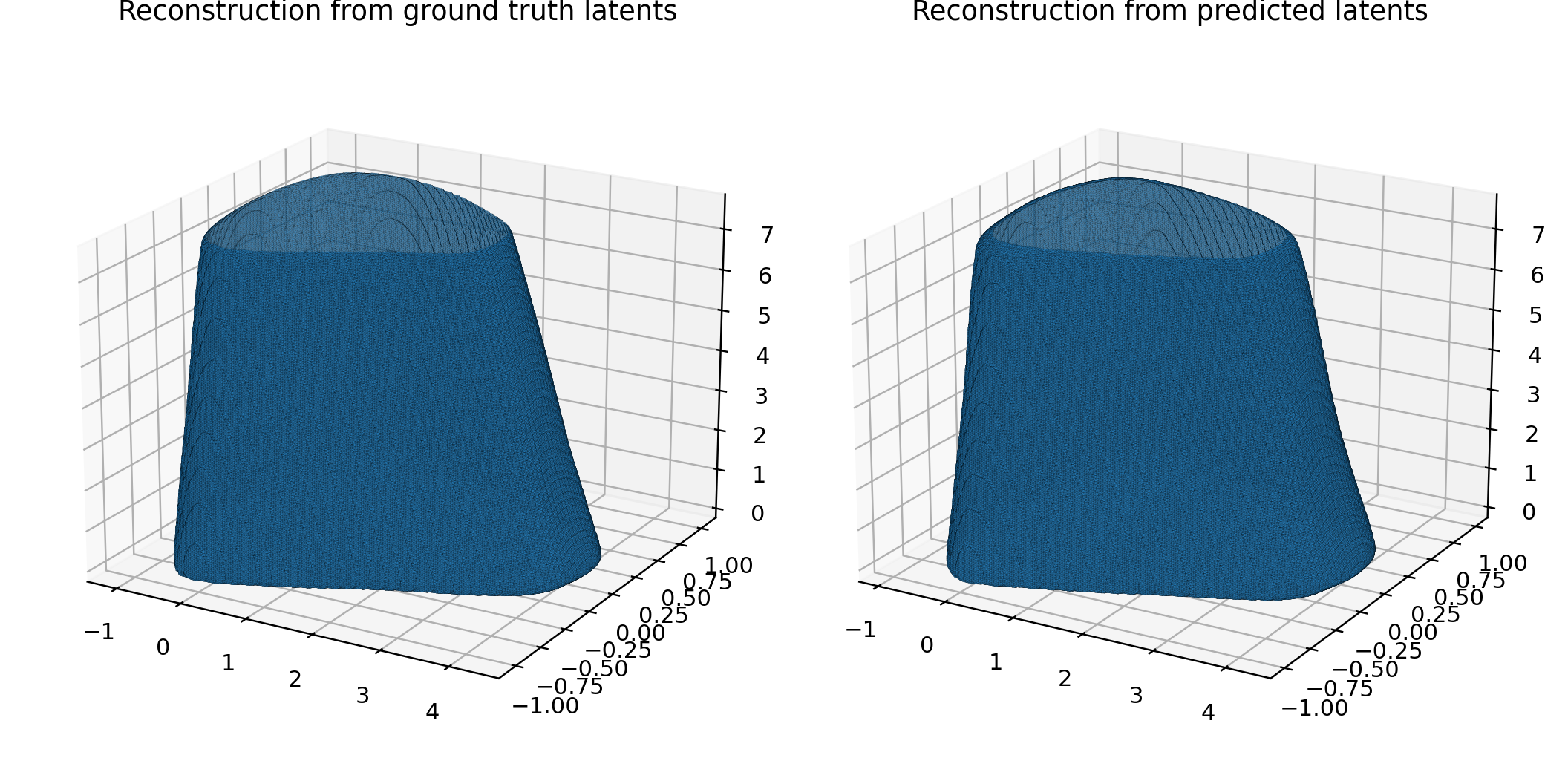}
        \caption{}
    \end{subfigure}

    \caption{Reconstructions at maximum strain extremes. NN-map decoded surfaces (left) versus reconstructions from optimized latents (right) for designs attaining $\max \varepsilon_x,\ \max \varepsilon_y,\ \max \varepsilon_z$. Visual agreement is strong; a slight discrepancy appears at $\max \varepsilon_y$, consistent with the learned map’s 4\% NRMSE.}
    \label{fig:Comp_max_strains}
\end{figure}

As a visual cross-check to the histogram-based analysis, we compare reconstructions decoded from NN-map predictions with those from the corresponding optimized latents. For each strain component, we select the design with the minimum (or maximum) $\varepsilon_x,\varepsilon_y,\varepsilon_z$ and use that design’s full strain triplet as NN-map input. Figure~\ref{fig:Comp_min_strains} (minima) shows close agreement across all three cases, with a mild deviation for the $\min \varepsilon_z$ example, consistent with the 4\% NRMSE for the strain-to-latent map and sub-percent surface distances reported earlier. Figure~\ref{fig:Comp_max_strains} (maxima) exhibits a similar good alignment, with a slight mismatch at $\max \varepsilon_y$ of comparable magnitude. Overall, the NN-map preserves geometry well at the extremes of the strain space.

\begin{figure}[ht!]
    \centering
    \begin{subfigure}[b]{0.49\textwidth}
        \centering
        \includegraphics[width=\textwidth]{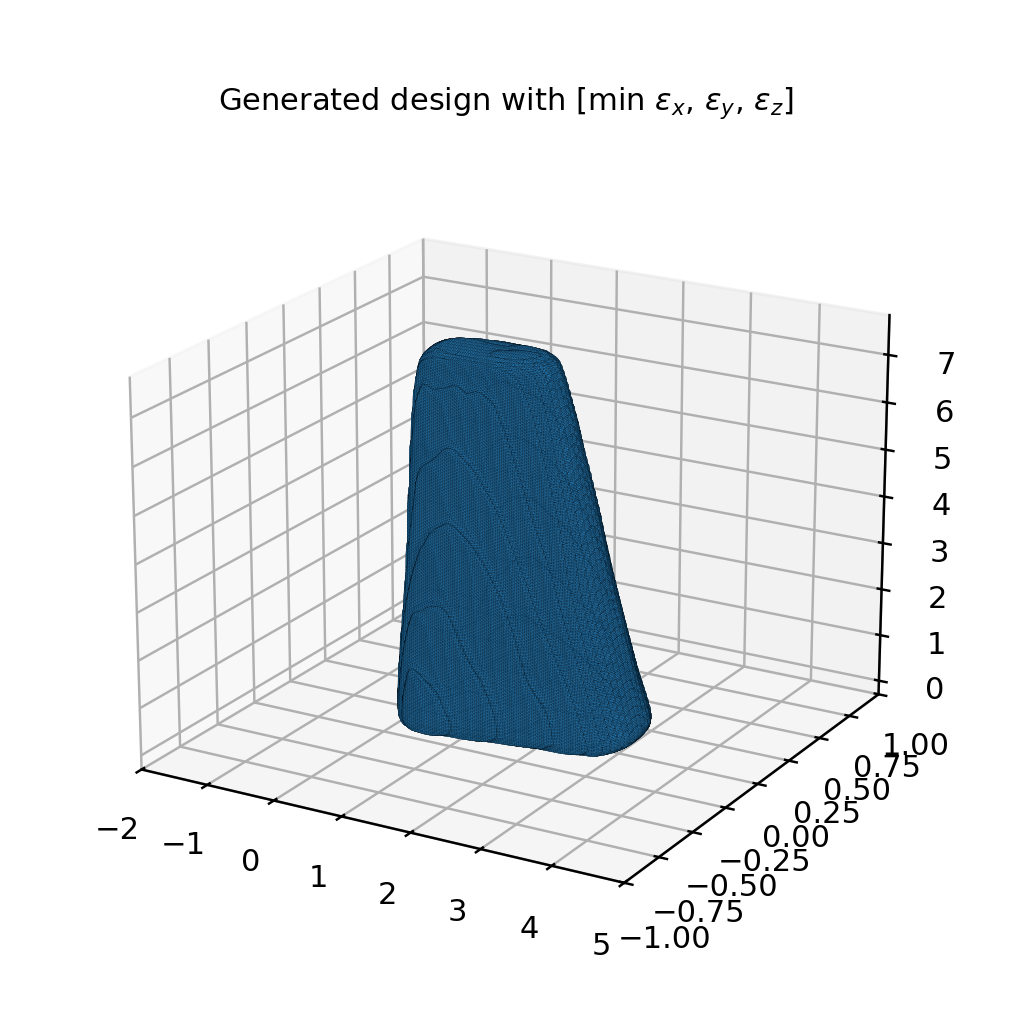}
        \caption{}
    \end{subfigure}
    \hfill
    \begin{subfigure}[b]{0.49\textwidth}
        \centering
        \includegraphics[width=\textwidth]{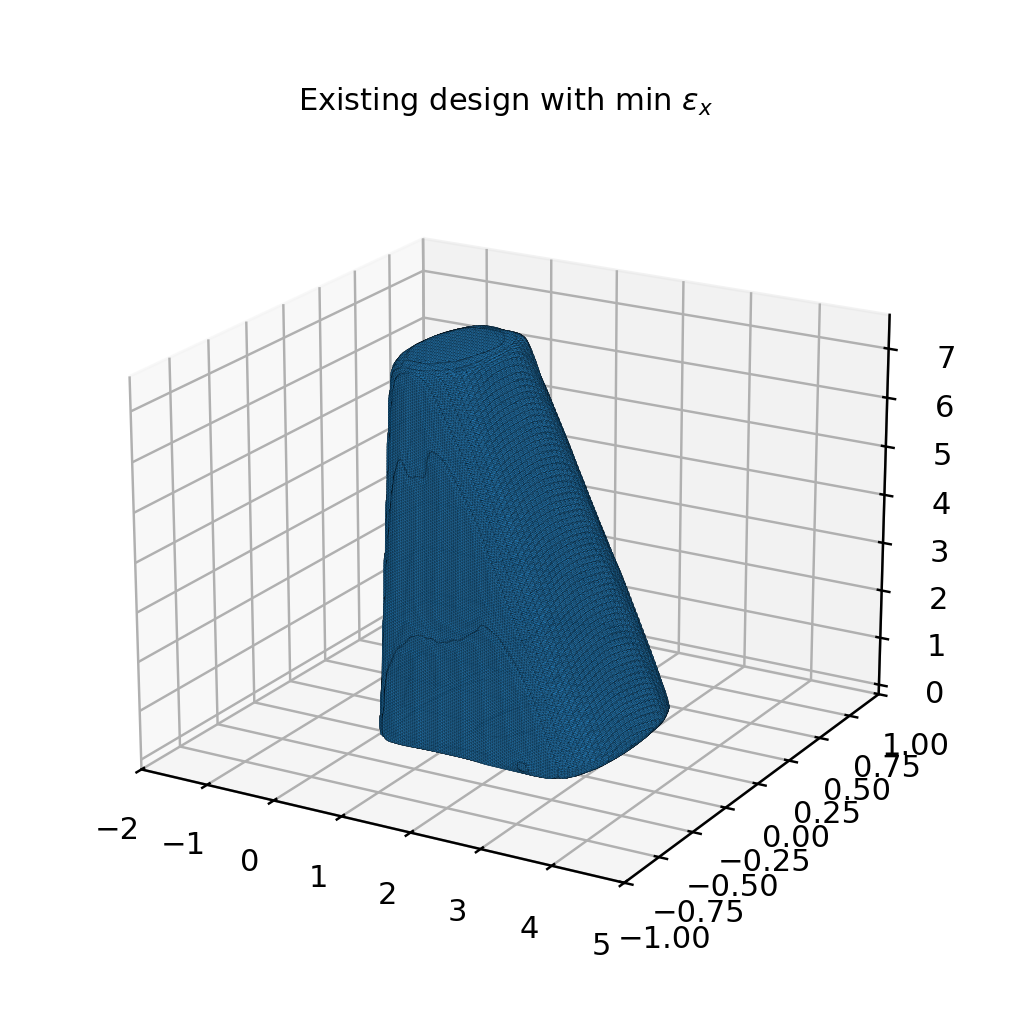}
        \caption{}
    \end{subfigure}
    
    \begin{subfigure}[b]{0.49\textwidth}
        \centering
        \includegraphics[width=\textwidth]{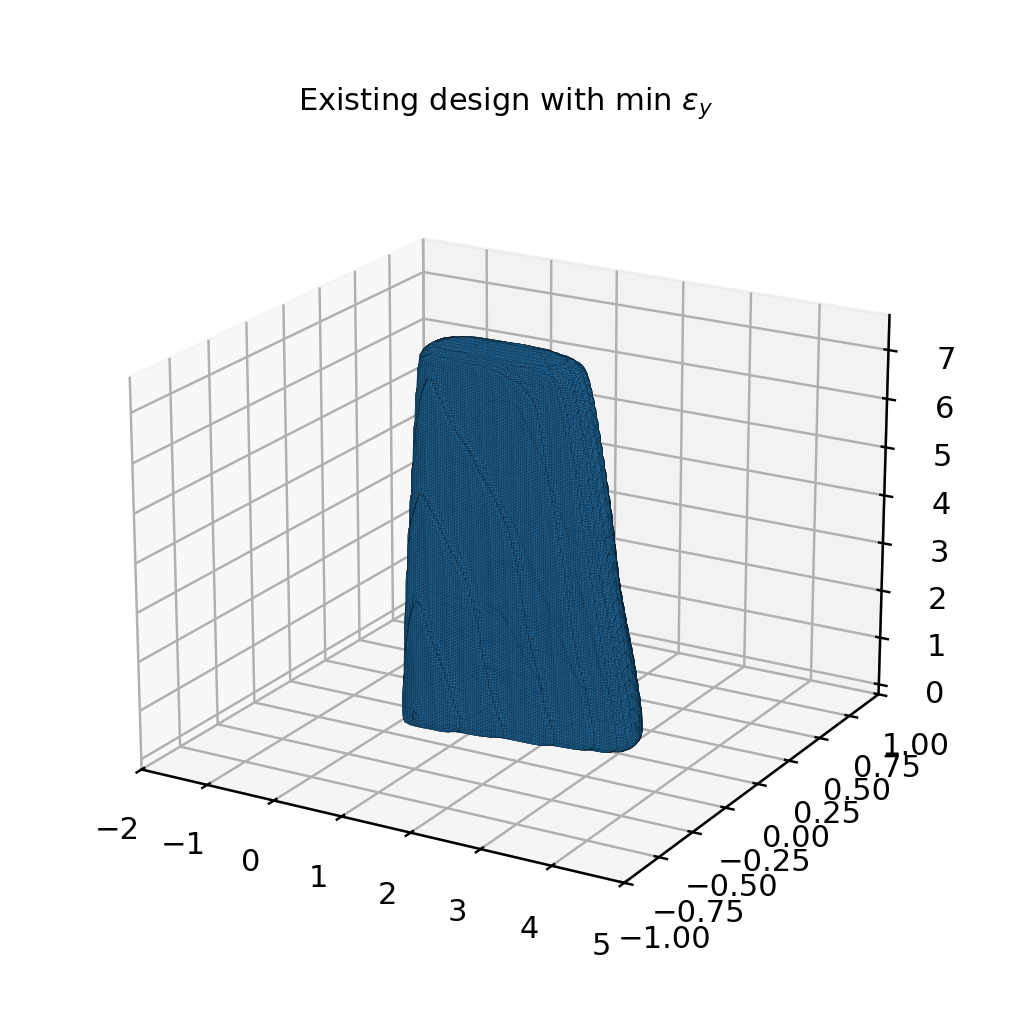}
        \caption{}
    \end{subfigure}
    \hfill
    \begin{subfigure}[b]{0.49\textwidth}
        \centering
        \includegraphics[width=\textwidth]{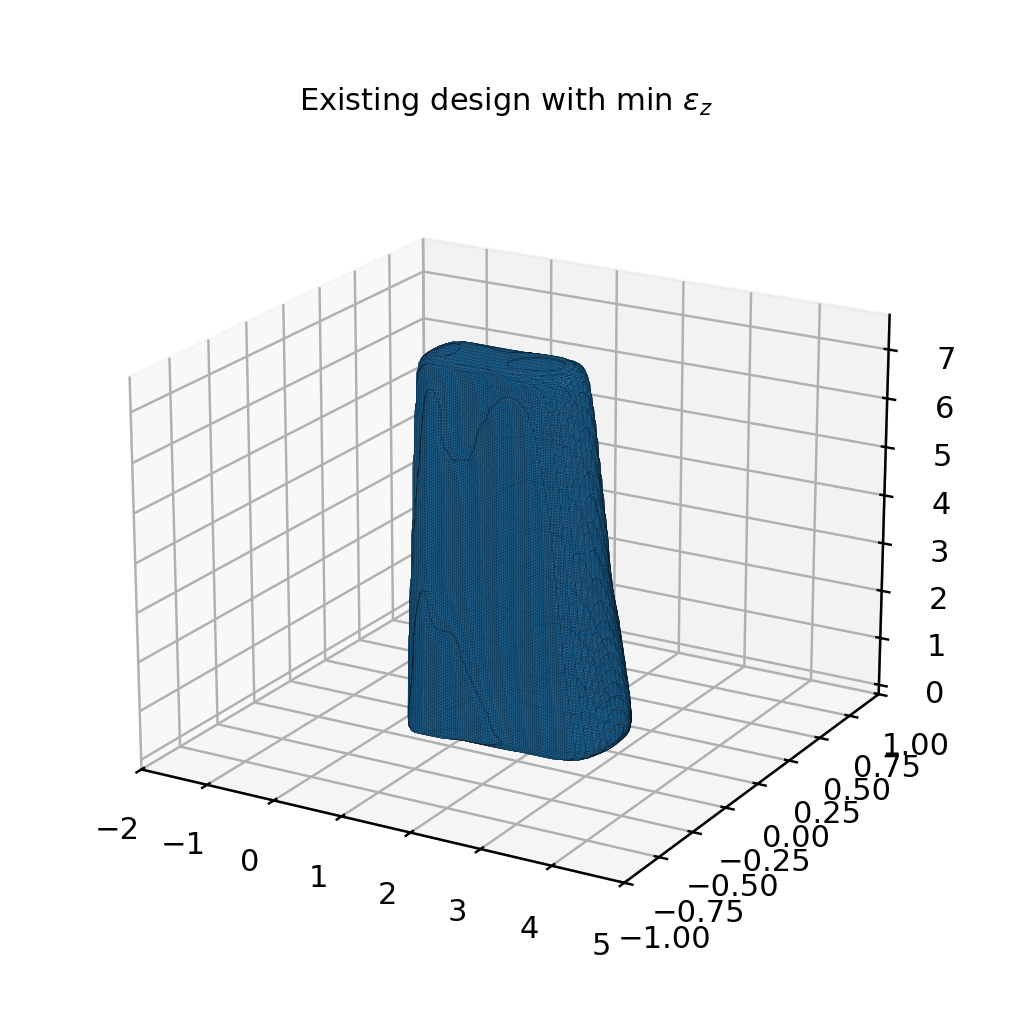}
        \caption{}
    \end{subfigure}
    
    \caption{conditional generation at strain minima : (a) Generated blade for $\varepsilon^*=(\min \varepsilon_x,\min \varepsilon_y,\min \varepsilon_z)$; (b–d) training designs with individual minima: (b) $\min \varepsilon_x$ with $(K_1,K_2,K_3)=(0.75,\,0.76,\,0.20)$, (c) $\min \varepsilon_y$ with $(0.50,\,0.66,\,0.73)$, (d) $\min \varepsilon_z$ with $(0.63,\,0.73,\,0.80)$.}
    \label{fig:min_strain_generation}
\end{figure}

 Having established a reliable strain-to-latent NN-map, we synthesize a new design by specifying a target with the lowest admissible strains in all three directions $\varepsilon^*=(\min\varepsilon_x,\min\varepsilon_y,\min\varepsilon_z)$, each chosen within the training ranges, and decoding $\mathbf z^*=g_\phi(\varepsilon^*)$. Figure~\ref{fig:min_strain_generation} compares this generated blade to the training designs that individually attain the minima of $\varepsilon_x,\varepsilon_y,\varepsilon_z$; their parameter tuples $(K_1,K_2,K_3)$ are $(0.75,0.76,0.20)$, $(0.50,0.66,0.73)$, and $(0.63,0.73,0.80)$, respectively. In the training set, the $\min \varepsilon_x$ design aligns with high $K_1$ (strong taper ratio), while the $\min \varepsilon_z$ design aligns with high $K_3$ (larger top–bottom chord ratio). The generated blade exhibits morphology consistent with high $K_1$ and high $K_3$, qualitatively matching these trends thus suggesting the NN-map can combine strain targets to navigate to plausible regions of the design manifold. A full structural analysis would be required to quantitatively verify that the synthesized geometry achieves the desired strain extrema. Within the observed strain ranges, the learned NN-map enables a target-driven, conditional synthesis that preserves geometric plausibility while offering controllable movement in the design space.

\section{Conclusions}

We showed that a DeepSDF auto-decoder can accurately reconstruct blade geometries while learning a useful, generative latent manifold. With an 8×512 decoder and 256-dimensional codes trained on 222 designs (clamped SDF loss, $\delta=0.1$), the training loss dropped rapidly and then stabilized. Reconstruction accuracy, measured by a directed surface distance, concentrated near $5.5\times10^{-2}$ with a shallow tail to $8\times10^{-2}$, corresponding to $\lesssim$1\% of the maximum blade extent ($D_{\max}\approx 7$). On 300 held-out designs, the absolute range remained comparable while the histogram flattened, consistent with frozen decoder weights and per-design code refinement at test time.

The learned latent space is structured and interpretable. PCA on test codes revealed that PC1 aligns with the taper ratio $K_1$ (transition from tapered/conical to prismatic shapes), and PC2 aligns with the top-to-bottom chord ratio $K_3$ (rectangular to trapezoidal profiles). The distribution of the training codes were close to $\mathcal N(0,0.1)$ per coordinate, with test codes exhibiting a modest increase in variance, a pattern that both explains the test-set error spread and supports simple, prior-consistent sampling schemes.

Building on this structure, two unconditional generation techniques, linear interpolation $\mathbf z(\alpha)=(1-\alpha)\mathbf z_0+\alpha\mathbf z_1$ and diagonal-Gaussian sampling $\mathbf z\sim\mathcal N(\hat\mu,\mathrm{diag}(\hat\sigma^2))$ produced smooth, artifact-free blades that traverse meaningful changes in chord and taper while staying near the training manifold. For conditional generation, a compact strain-to-latent NN-map (2 hidden layers, 128 units each) achieved ~4\% NRMSE on training codes and a controlled increase on test data; surface-distance comparisons between reconstructions from optimized codes and NN-map codes were typically within 1\% , with outliers. Specifying simultaneous minima in $(\varepsilon_x,\varepsilon_y,\varepsilon_z)$ yielded a novel blade whose morphology matched trends observed at the strain extremes (relatively high $K_1$ and $K_3$).

Overall, DeepSDF provides accurate reconstructions, an interpretable latent representation tied to original blade parameters, and practical pathways for both unconditional and strain-conditional design generation, positioning it as a viable backbone for data-driven blade modeling and targeted concept generation.

Looking ahead, integrating predictive uncertainty into reconstruction and conditional synthesis would enable risk-aware exploration and selection of candidate geometries \cite{kendall2017uncertainty}. Interactive, design-in-the-loop strategies can then prioritize informative latent regions \cite{sener2018active}, and coupling with differentiable shape programs and CAD operators can further enforce engineering constraints during generation.

\bibliography{sample}

\begin{thebibliography}{41}
\newcommand{\enquote}[1]{``#1''}
\providecommand{\natexlab}[1]{#1}
\providecommand{\url}[1]{\texttt{#1}}
\providecommand{\urlprefix}{URL }
\expandafter\ifx\csname urlstyle\endcsname\relax
  \providecommand{\doi}[1]{\discretionary{}{}{}https://doi.org/#1}\else
  \providecommand{\doi}[1]{\discretionary{}{}{}\urlstyle{rm}\url{https://doi.org/#1}}\fi

\bibitem[{Mescheder et~al.(2019)Mescheder, Oechsle, Niemeyer, Nowozin, and Geiger}]{mescheder2019occupancy}
Mescheder, L., Oechsle, M., Niemeyer, M., Nowozin, S., and Geiger, A., \enquote{Occupancy Networks: Learning 3D Reconstruction in Function Space,} , 2019.

\bibitem[{Chen and Zhang(2019{\natexlab{a}})}]{chen2019implicit}
Chen, Z., and Zhang, H., \enquote{Learning implicit fields for generative shape modeling,} \emph{Proceedings of the IEEE/CVF conference on computer vision and pattern recognition}, 2019{\natexlab{a}}, pp. 5939--5948.

\bibitem[{Gropp et~al.(2020)Gropp, Yariv, Haim, Atzmon, and Lipman}]{gropp2020igrl}
Gropp, A., Yariv, L., Haim, N., Atzmon, M., and Lipman, Y., \enquote{Implicit Geometric Regularization for Learning Shapes,} , 2020.

\bibitem[{Yariv et~al.(2020)Yariv, Kasten, Moran, Galun, Atzmon, Ronen, and Lipman}]{yariv2020multiview}
Yariv, L., Kasten, Y., Moran, D., Galun, M., Atzmon, M., Ronen, B., and Lipman, Y., \enquote{Multiview neural surface reconstruction by disentangling geometry and appearance,} \emph{Advances in Neural Information Processing Systems}, Vol.~33, 2020, pp. 2492--2502.

\bibitem[{Regenwetter et~al.(2022)Regenwetter, Nobari, and Ahmed}]{regenwetter2022deep}
Regenwetter, L., Nobari, A.~H., and Ahmed, F., \enquote{Deep generative models in engineering design: A review,} \emph{Journal of Mechanical Design}, Vol. 144, No.~7, 2022, p. 071704.

\bibitem[{Atzmon and Lipman(2020)}]{atzmon2020sal}
Atzmon, M., and Lipman, Y., \enquote{Sal: Sign agnostic learning of shapes from raw data,} \emph{Proceedings of the IEEE/CVF conference on computer vision and pattern recognition}, 2020, pp. 2565--2574.

\bibitem[{Takikawa et~al.(2021)Takikawa, Litalien, Yin, Kreis, Loop, Nowrouzezahrai, Jacobson, McGuire, and Fidler}]{takikawa2021nglod}
Takikawa, T., Litalien, J., Yin, K., Kreis, K., Loop, C., Nowrouzezahrai, D., Jacobson, A., McGuire, M., and Fidler, S., \enquote{Neural geometric level of detail: Real-time rendering with implicit 3d shapes,} \emph{Proceedings of the IEEE/CVF conference on computer vision and pattern recognition}, 2021, pp. 11358--11367.

\bibitem[{Park et~al.(2019)Park, Florence, Straub, Newcombe, and Lovegrove}]{park2019deepsdf}
Park, J.~J., Florence, P., Straub, J., Newcombe, R., and Lovegrove, S., \enquote{Deepsdf: Learning continuous signed distance functions for shape representation,} \emph{Proceedings of the IEEE/CVF conference on computer vision and pattern recognition}, 2019, pp. 165--174.

\bibitem[{Chou et~al.(2023)Chou, Bahat, and Heide}]{chou2023diffusion}
Chou, G., Bahat, Y., and Heide, F., \enquote{Diffusion-SDF: Conditional Generative Modeling of Signed Distance Functions,} , 2023.

\bibitem[{Ho et~al.(2020)Ho, Jain, and Abbeel}]{ho2020ddpm}
Ho, J., Jain, A., and Abbeel, P., \enquote{Denoising diffusion probabilistic models,} \emph{Advances in neural information processing systems}, Vol.~33, 2020, pp. 6840--6851.

\bibitem[{Nichol and Dhariwal(2021)}]{nichol2021improved}
Nichol, A.~Q., and Dhariwal, P., \enquote{Improved denoising diffusion probabilistic models,} \emph{International conference on machine learning}, PMLR, 2021, pp. 8162--8171.

\bibitem[{Rombach et~al.(2022)Rombach, Blattmann, Lorenz, Esser, and Ommer}]{rombach2022ldm}
Rombach, R., Blattmann, A., Lorenz, D., Esser, P., and Ommer, B., \enquote{High-resolution image synthesis with latent diffusion models,} \emph{Proceedings of the IEEE/CVF conference on computer vision and pattern recognition}, 2022, pp. 10684--10695.

\bibitem[{Vahdat et~al.(2021)Vahdat, Kreis, and Kautz}]{vahdat2021scorelatent}
Vahdat, A., Kreis, K., and Kautz, J., \enquote{Score-based generative modeling in latent space,} \emph{Advances in neural information processing systems}, Vol.~34, 2021, pp. 11287--11302.

\bibitem[{Chou et~al.(2022)Chou, Chugunov, and Heide}]{chou2022gensdf}
Chou, G., Chugunov, I., and Heide, F., \enquote{Gensdf: Two-stage learning of generalizable signed distance functions,} \emph{Advances in Neural Information Processing Systems}, Vol.~35, 2022, pp. 24905--24919.

\bibitem[{Yariv et~al.(2021)Yariv, Gu, Kasten, and Lipman}]{yariv2021volsdf}
Yariv, L., Gu, J., Kasten, Y., and Lipman, Y., \enquote{Volume rendering of neural implicit surfaces,} \emph{Advances in neural information processing systems}, Vol.~34, 2021, pp. 4805--4815.

\bibitem[{Wang et~al.(2021)Wang, Liu, Liu, Theobalt, Komura, and Wang}]{wang2021neus}
Wang, P., Liu, L., Liu, Y., Theobalt, C., Komura, T., and Wang, W., \enquote{Neus: Learning neural implicit surfaces by volume rendering for multi-view reconstruction,} \emph{arXiv preprint arXiv:2106.10689}, 2021.

\bibitem[{Sitzmann et~al.(2020)Sitzmann, Martel, Bergman, Lindell, and Wetzstein}]{sitzmann2020siren}
Sitzmann, V., Martel, J., Bergman, A., Lindell, D., and Wetzstein, G., \enquote{Implicit neural representations with periodic activation functions,} \emph{Advances in neural information processing systems}, Vol.~33, 2020, pp. 7462--7473.

\bibitem[{Chibane and Pons-Moll(2020)}]{chibane2020ifnets}
Chibane, J., and Pons-Moll, G., \enquote{Implicit feature networks for texture completion from partial 3d data,} \emph{European Conference on Computer Vision}, Springer, 2020, pp. 717--725.

\bibitem[{Poole et~al.(2022)Poole, Jain, Barron, and Mildenhall}]{poole2022dreamfusion}
Poole, B., Jain, A., Barron, J.~T., and Mildenhall, B., \enquote{Dreamfusion: Text-to-3d using 2d diffusion,} \emph{arXiv preprint arXiv:2209.14988}, 2022.

\bibitem[{Lin et~al.(2023)Lin, Gao, Tang, Takikawa, Zeng, Huang, Kreis, Fidler, Liu, and Lin}]{lin2023magic3d}
Lin, C.-H., Gao, J., Tang, L., Takikawa, T., Zeng, X., Huang, X., Kreis, K., Fidler, S., Liu, M.-Y., and Lin, T.-Y., \enquote{Magic3d: High-resolution text-to-3d content creation,} \emph{Proceedings of the IEEE/CVF conference on computer vision and pattern recognition}, 2023, pp. 300--309.

\bibitem[{Xu et~al.(2022)Xu, Peng, Yang, Shen, and Zhou}]{shue2022threeda}
Xu, Y., Peng, S., Yang, C., Shen, Y., and Zhou, B., \enquote{3D-Aware Image Synthesis via Learning Structural and Textural Representations,} \emph{Proceedings of the IEEE/CVF Conference on Computer Vision and Pattern Recognition (CVPR)}, 2022, pp. 18430--18439.

\bibitem[{Elton et~al.(2019)Elton, Boukouvalas, Fuge, and Chung}]{fuge2019ai4design}
Elton, D.~C., Boukouvalas, Z., Fuge, M.~D., and Chung, P.~W., \enquote{Deep learning for molecular design—a review of the state of the art,} \emph{Molecular Systems Design \& Engineering}, Vol.~4, No.~4, 2019, pp. 828--849.

\bibitem[{Aulich et~al.(2024)Aulich, Goinis, and Vo{\ss}}]{sun2021geomperf}
Aulich, M., Goinis, G., and Vo{\ss}, C., \enquote{Data-driven AI model for turbomachinery compressor aerodynamics enabling rapid approximation of 3D flow solutions,} \emph{Aerospace}, Vol.~11, No.~9, 2024, p. 723.

\bibitem[{Maulik et~al.(2020)Maulik, Fukami, Ramachandra, Fukagata, and Taira}]{maulik2020nncfd}
Maulik, R., Fukami, K., Ramachandra, N., Fukagata, K., and Taira, K., \enquote{Probabilistic neural networks for fluid flow surrogate modeling and data recovery,} \emph{Physical Review Fluids}, Vol.~5, No.~10, 2020, p. 104401.

\bibitem[{Meng et~al.(2019)Meng, Yang, Zhang, and Zhu}]{yao2020surrogate}
Meng, D., Yang, S., Zhang, Y., and Zhu, S.-P., \enquote{Structural reliability analysis and uncertainties-based collaborative design and optimization of turbine blades using surrogate model,} \emph{Fatigue \& Fracture of Engineering Materials \& Structures}, Vol.~42, No.~6, 2019, pp. 1219--1227.

\bibitem[{Viquerat et~al.(2021)Viquerat, Rabault, Kuhnle, Ghraieb, Larcher, and Hachem}]{viquerat2021turbodl}
Viquerat, J., Rabault, J., Kuhnle, A., Ghraieb, H., Larcher, A., and Hachem, E., \enquote{Direct shape optimization through deep reinforcement learning,} \emph{Journal of Computational Physics}, Vol. 428, 2021, p. 110080.

\bibitem[{Morita et~al.(2024)Morita, Shintani, Yuan, and Permenter}]{morita2024vehiclesdf}
Morita, H., Shintani, K., Yuan, C., and Permenter, F., \enquote{VehicleSDF: A 3D generative model for constrained engineering design via surrogate modeling,} , 2024.

\bibitem[{Yang et~al.(2023)Yang, Liang, Li, Zhang, Lin, Sheffer, Schaefer, Keyser, and Wang}]{wang2023neuralparamcad}
Yang, L., Liang, Y., Li, X., Zhang, C., Lin, G., Sheffer, A., Schaefer, S., Keyser, J., and Wang, W., \enquote{Neural Parametric Surfaces for Shape Modeling,} , 2023.

\bibitem[{B{\ae}rentzen(2005)}]{jones2006sdf}
B{\ae}rentzen, J.~A., \enquote{Robust generation of signed distance fields from triangle meshes,} \emph{Fourth International Workshop on Volume Graphics, 2005.}, IEEE, 2005, pp. 167--239.

\bibitem[{B{\ae}rentzen and Aanaes(2005)}]{baerentzen2005fastmarch}
B{\ae}rentzen, J.~A., and Aanaes, H., \enquote{Signed distance computation using the angle weighted pseudonormal,} \emph{IEEE Transactions on Visualization and Computer Graphics}, Vol.~11, No.~3, 2005, pp. 243--253.

\bibitem[{Barill et~al.(2018)Barill, Dickson, Schmidt, Levin, and Jacobson}]{barill2018fastwinding}
Barill, G., Dickson, N.~G., Schmidt, R., Levin, D.~I., and Jacobson, A., \enquote{Fast winding numbers for soups and clouds,} \emph{ACM Transactions on Graphics (TOG)}, Vol.~37, No.~4, 2018, pp. 1--12.

\bibitem[{Kingma and Welling(2013)}]{kingma2014vae}
Kingma, D.~P., and Welling, M., \enquote{Auto-encoding variational bayes,} , 2013.

\bibitem[{Rezende and Mohamed(2015)}]{rezende2015flows}
Rezende, D., and Mohamed, S., \enquote{Variational inference with normalizing flows,} \emph{International conference on machine learning}, PMLR, 2015, pp. 1530--1538.

\bibitem[{Lorensen and Cline(1998)}]{lorensen1987marchingcubes}
Lorensen, W.~E., and Cline, H.~E., \emph{Marching cubes: a high resolution 3D surface construction algorithm}, Association for Computing Machinery, New York, NY, USA, 1998, p. 347–353.
\newblock \urlprefix\url{https://doi.org/10.1145/280811.281026}.

\bibitem[{Kazhdan et~al.(2006)Kazhdan, Bolitho, and Hoppe}]{kazhdan2006poisson}
Kazhdan, M., Bolitho, M., and Hoppe, H., \enquote{Poisson surface reconstruction,} \emph{Proceedings of the fourth Eurographics symposium on Geometry processing}, Vol.~7, 2006.

\bibitem[{Berger et~al.(2017)Berger, Tagliasacchi, Seversky, Alliez, Guennebaud, Levine, Sharf, and Silva}]{berger2017reconSurvey}
Berger, M., Tagliasacchi, A., Seversky, L.~M., Alliez, P., Guennebaud, G., Levine, J.~A., Sharf, A., and Silva, C.~T., \enquote{A survey of surface reconstruction from point clouds,} \emph{Computer graphics forum}, Vol.~36, Wiley Online Library, 2017, pp. 301--329.

\bibitem[{Chen and Zhang(2019{\natexlab{b}})}]{chen2019imnet}
Chen, Z., and Zhang, H., \enquote{Learning Implicit Fields for Generative Shape Modeling,} , 2019{\natexlab{b}}.

\bibitem[{Mittal et~al.(2022)Mittal, Cheng, Singh, and Tulsiani}]{mittal2022autosdf}
Mittal, P., Cheng, Y.-C., Singh, M., and Tulsiani, S., \enquote{Autosdf: Shape priors for 3d completion, reconstruction and generation,} \emph{Proceedings of the IEEE/CVF conference on computer vision and pattern recognition}, 2022, pp. 306--315.

\bibitem[{Achlioptas et~al.(2018)Achlioptas, Diamanti, Mitliagkas, and Guibas}]{achlioptas2018learning}
Achlioptas, P., Diamanti, O., Mitliagkas, I., and Guibas, L., \enquote{Learning representations and generative models for 3d point clouds,} \emph{International conference on machine learning}, PMLR, 2018, pp. 40--49.

\bibitem[{Kendall and Gal(2017)}]{kendall2017uncertainty}
Kendall, A., and Gal, Y., \enquote{What uncertainties do we need in bayesian deep learning for computer vision?} \emph{Advances in neural information processing systems}, Vol.~30, 2017.

\bibitem[{Sener and Savarese(2018)}]{sener2018active}
Sener, O., and Savarese, S., \enquote{Active Learning for Convolutional Neural Networks: A Core-Set Approach,} , 2018.

\end{thebibliography}

\end{document}